\documentclass[lettersize,journal]{IEEEtran}
\usepackage{amsmath,amsfonts}
\usepackage{algorithmic}
\usepackage{algorithm}
\usepackage{array}
\usepackage[caption=false,font=normalsize,labelfont=sf,textfont=sf]{subfig}
\usepackage{textcomp}
\usepackage{stfloats}
\usepackage{url}
\usepackage{verbatim}
\usepackage{graphicx}
\usepackage{cite}
\usepackage{mathtools}
\usepackage{booktabs}
\usepackage{multirow}
\usepackage[table]{xcolor}
\usepackage[table]{xcolor}
\definecolor{darkgreen}{rgb}{0.0,0.5,0.0}

\usepackage{pifont}
\newcommand{\cmark}{\ding{51}} 
\newcommand{\xmark}{\ding{55}} 

\usepackage{tikz}
\usetikzlibrary{arrows.meta, positioning}

\usepackage{adjustbox}
\usepackage{caption}

\begin{document}

\title{PP-Net: A Hybrid Physical-Prior Neural Network for Scattered Light Removal in Biomedical Images on Embedded Devices}

\author{
	Yongfei~Guo,
	Tingjin~Chu,
	Mengzhuo~Liu,
	Hongwei~Lou,
	and~Yuanhao~Gong%
	\thanks{Yongfei Guo is with the Changchun Institute of Optics, Fine Mechanics and Physics, Chinese Academy of Sciences, Changchun~130033, China, and also with the University of Chinese Academy of Sciences, Beijing, China.}%
	\thanks{Tingjin Chu is with the School of Mathematics and Statistics, University of Melbourne, Melbourne, Australia.}%
	\thanks{Mengzhuo Liu, Hongwei Lou, and Yuanhao Gong are with the Changchun Institute of Optics, Fine Mechanics and Physics, Chinese Academy of Sciences, Changchun~130033, China, and also with the Chinese Academy of Sciences, Beijing, China.}%
	\thanks{Corresponding author: Yuanhao Gong (e-mail: gong.ai@qq.com).}%
}
\markboth{IEEE Internet of Things Journal}%
{Guo \MakeLowercase{\textit{et al.}}: PP-Net: A Hybrid Physical-Prior Neural Network for Scattered Light Removal}




\maketitle

\begin{abstract}
		
	Scattered light is common in biomedical images, yet its removal remains challenging. The difficulty arises from three aspects: first, aligned scattered-light-free biomedical ground truth is often unavailable; second, scattering is coupled with weak illumination and sensor-induced noise; and third, many learning-based restoration models are computationally expensive for embedded devices in Internet of Medical Things (IoMT) scenarios. To address these issues, this paper proposes PP-Net, a hybrid physical-prior neural network for biomedical scattered light removal. The proposed method consists of three components: DFN-Net suppresses sensor-induced noise, ASAP estimates the scattering map and recovers a physics-based prior map, and GF-Net refines the prior map by fusing it with the denoised observation.  To reduce the dependence on paired biomedical ground truth, a progressive synthetic training and cross-domain transfer strategy is developed. Experiments show that the physical-prior branch improves the peak signal-to-noise ratio (PSNR) by up to 1.26 dB on paired synthetic benchmarks. Under joint noise-and-scattering degradation, PP-Net improves PSNR by more than 10.8 dB and the structural similarity index measure (SSIM) by more than 0.62 compared with representative baseline methods. On real W2S biomedical images, the proposed method reduces the average Natural Image Quality Evaluator (NIQE) score by 43.3\%. Edge deployment with RKNN conversion and INT8 quantization achieves an average inference latency of approximately 200 ms per $512\times512$ image over 360 test images. These results demonstrate that PP-Net provides an effective and deployable solution for microscopic imaging, endoscopic inspection, and edge-assisted biomedical analysis in IoMT scenarios.
	
\end{abstract}


\section{Introduction}
\label{sec:introduction}

Scattered light is a common phenomenon in optical imaging. In natural environments, light scattering caused by atmospheric particles, aerosols, water droplets, and other turbid media often produces haze-like degradation, reducing image contrast and obscuring scene details. Similar scattering effects also exist in biomedical imaging, where photons interact with biological tissues, cellular structures, and heterogeneous media during acquisition. Unlike natural scene imaging, however, biomedical imaging is often performed under weak, localized, and safety-constrained illumination, making the captured images more vulnerable to scattering corruption, low signal-to-noise ratios, and sensor-induced noise.

In Internet of Medical Things (IoMT) scenarios, biomedical imaging devices such as endoscopes and microscopes serve as front-end sensing nodes for clinical observation, remote diagnosis, and downstream visual analysis~\cite{Survey1,Survey2,endoscopic,Dark}. Degraded images may blur tissue boundaries, reduce structural visibility, and obscure diagnostically relevant details. Meanwhile, practical IoMT systems require not only accurate restoration but also reduced dependence on paired biomedical ground truth, efficient inference, and feasible deployment on resource-constrained edge devices. These requirements motivate a restoration method that integrates physical interpretability, deep feature refinement, and edge-oriented efficiency.

\subsection{From Natural Scattering to Biomedical Imaging}

In natural image processing, scattered light is commonly studied in the form of haze or atmospheric scattering. Classical dehazing methods usually rely on an image formation model that relates the observed hazy image to scene radiance, transmission, and atmospheric light. Based on this model, a variety of physical priors have been developed to estimate the scattering component and recover the latent clear image. Among them, the dark channel prior (DCP) has become a representative approach because of its simplicity, interpretability, and strong empirical performance in natural scenes~\cite{DCP}.

Although natural image dehazing and biomedical scattered-light removal share similar physical intuition, the two problems are not identical. In natural scenes, paired hazy and haze-free images can often be synthesized or approximated using outdoor image formation models. In biomedical imaging, however, scattering is usually coupled with tissue morphology, weak illumination, sensor noise, and device-dependent acquisition conditions. More importantly, strictly aligned scattering-free biomedical ground truth is generally unavailable in real clinical settings. Therefore, methods designed for natural dehazing cannot be directly transferred to biomedical scattered-light removal without considering the specific degradation characteristics of biomedical images.

\subsection{Physical Priors for Scattered-Light Removal}

Physical-prior-based methods are attractive for scattered-light removal because they provide interpretable intermediate estimates and do not necessarily require large-scale paired training data. In natural image dehazing, DCP-based methods and their variants estimate scattering-related statistics from local image neighborhoods~\cite{DCP,IHDCP,GLP}. Similar ideas have also inspired biomedical scattered-light removal, where prior-based estimation can provide physically meaningful guidance when clean targets are unavailable~\cite{HDCP,QDCP,Dark}.

However, conventional priors remain limited in biomedical scenarios. First, fixed-window estimation may cross tissue boundaries and introduce boundary leakage or edge-shifting artifacts. Second, dark-channel statistics can be corrupted by high-frequency sensor noise and heterogeneous tissue textures. Third, biomedical images often contain fine structural details and irregular local transitions, making a single fixed neighborhood insufficient for reliable scattering estimation. These limitations suggest that the prior estimation mechanism should be adapted to local biomedical structures rather than relying on a fixed support window. This motivates the adaptive scattering-prior estimation strategy adopted in this work.

\subsection{Deep Restoration Under Biomedical Constraints}

Deep learning has significantly advanced image restoration, including denoising, deblurring, dehazing, low-light enhancement, and general image reconstruction~\cite{VITtansform,AOD-Net,MCRFS-Net}. Compared with handcrafted priors, deep models provide stronger nonlinear representation capability and can recover complex local textures from degraded observations. For scattered-light removal, learning-based refinement is particularly useful because a physical prior alone may not fully restore subtle tissue details or suppress residual artifacts.

Nevertheless, fully supervised deep restoration is difficult to apply directly to biomedical scattered-light removal. Most supervised restoration networks require paired degraded and clean images, whereas scattering-free biomedical ground truth is rarely available in real acquisition settings. In addition, models trained on synthetic data may suffer from domain gaps when transferred to real biomedical images. Recent weakly supervised, unsupervised, and cross-domain restoration strategies have attempted to reduce this dependence on paired data~\cite{FAMED-Net,gong,DS-NET}. However, without a stable physical constraint, cross-domain restoration may generate structurally inconsistent or diagnostically unreliable details. These observations motivate a hybrid design that combines physical-prior guidance with deep restoration.

\subsection{Biomedical Restoration on Embedded Devices}

Beyond restoration quality, practical deployment is an important requirement in IoMT-oriented biomedical imaging. Edge-side processing can reduce data transmission, preserve privacy, and support local visual enhancement in bandwidth-constrained or delay-sensitive scenarios. However, high-performance deep restoration networks are often computationally intensive, which limits their deployment on resource-constrained edge devices. Therefore, biomedical scattered-light removal methods for IoMT applications should be designed with both restoration accuracy and computational efficiency in mind.

Existing lightweight restoration models improve inference efficiency through compact backbones, efficient convolutions, and deployment-oriented optimization~\cite{ref_fpga_net,ref_rk3588_net}. However, many of them are designed for general natural-image restoration and do not explicitly address biomedical scattering, sensor noise, limited paired supervision, and physical interpretability at the same time. This creates a need for an edge-oriented biomedical restoration framework that is both physically meaningful and computationally practical.

\subsection{Motivations and Contributions}

The above analysis shows that biomedical scattered-light removal in IoMT scenarios requires a framework that can jointly address scattering corruption, sensor-induced noise, limited paired supervision, and edge-side deployment constraints. Physical priors provide interpretability but are sensitive to noise and local structural variations. Deep restoration models provide strong representation capability but usually require paired clean targets and may be difficult to deploy on resource-constrained devices. These challenges motivate the proposed PP-Net, a hybrid physical-prior neural network for biomedical scattered-light removal.

PP-Net follows a progressive network-prior-network design. A Denoising Front-End Network (DFN-Net), implemented using FFDNet~\cite{FFDNet}, first suppresses sensor-induced noise. An Advanced Scattering Adaptive Prior (ASAP) module then estimates the scattering map and recovers a physics-based prior map in a physically interpretable manner. Finally, a Guided Fusion Network (GF-Net) refines the prior map by fusing it with the denoised observation. To reduce the dependence on paired biomedical ground truth, we further develop a progressive synthetic training and cross-domain transfer strategy.

The contributions of this work are summarized as follows:
\begin{itemize}
	\item We formulate a noise-aware biomedical scattered-light degradation model that decomposes image degradation into tissue scattering and sensor-induced noise.
	\item We propose PP-Net, a hybrid physical-prior neural network that integrates DFN-Net, ASAP, and GF-Net for biomedical scattered-light removal.
	\item We develop a progressive synthetic training and cross-domain transfer strategy to reduce the dependence on paired biomedical ground-truth data.
	\item We implement the proposed method on a resource-constrained edge device and verify its feasibility for edge-side biomedical image enhancement in IoMT scenarios.
\end{itemize}

\section{Hybrid Physical-Prior Network Framework}
\label{sec:methodology}

\begin{figure*}[htbp]
	\centering
	\includegraphics[width=0.85\textwidth]{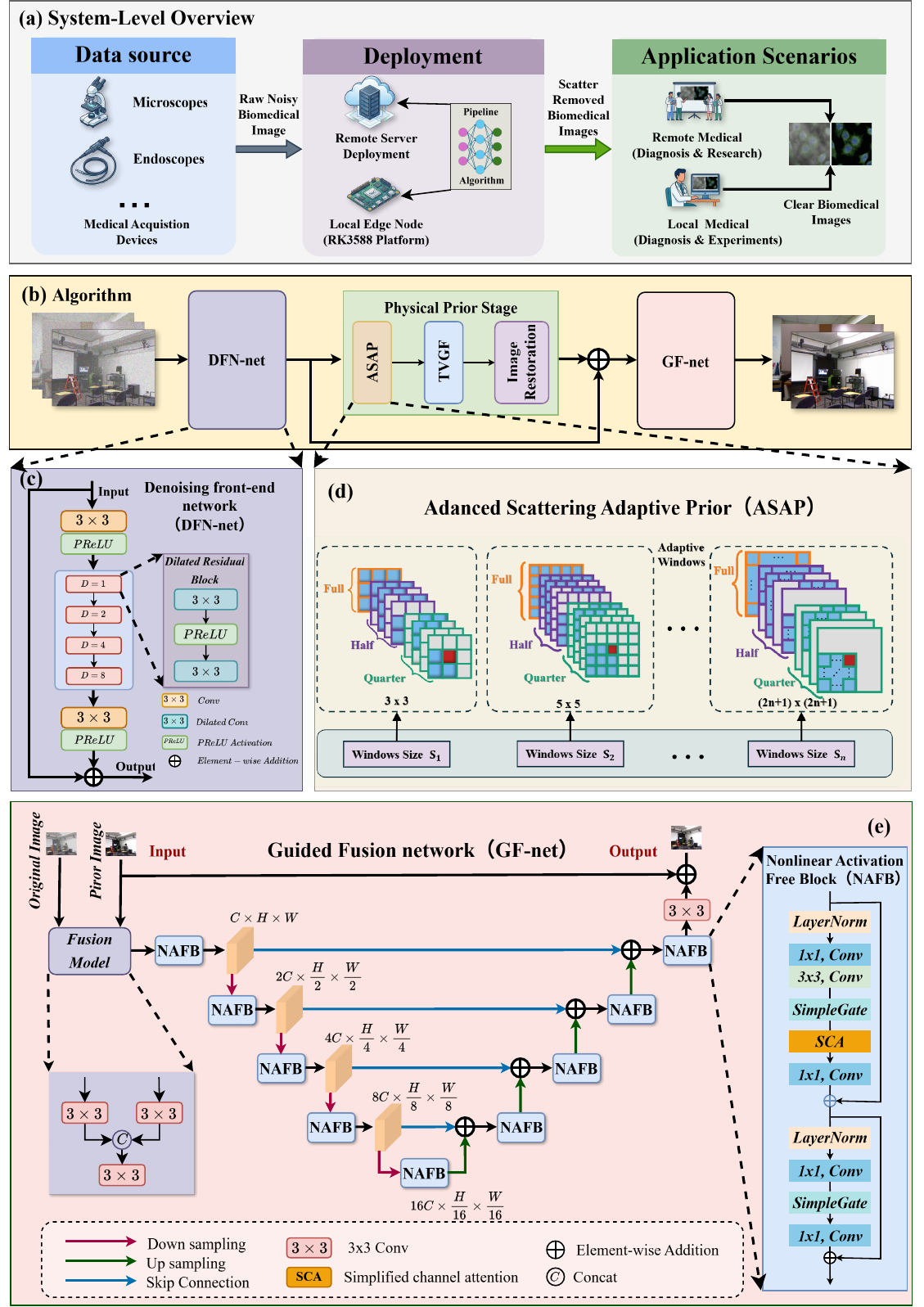}
	\caption{Overview of the proposed hybrid pipeline for biomedical scattered-light removal. \textbf{(a)} System-level architecture for IoMT edge/server deployment. \textbf{(b)} Three-stage algorithmic data flow. \textbf{(c)} DFN-Net for edge-friendly initial noise suppression. \textbf{(d)} ASAP for scattering-map estimation and structure-preserving prior-map recovery. \textbf{(e)} GF-Net for guided fusion and high-fidelity detail recovery.}
	\label{fig:overall_system}
\end{figure*}

This section presents the proposed hybrid pipeline for biomedical scattered-light removal. The framework integrates adaptive physics-based prior estimation, lightweight deep refinement, and progressive synthetic training. Given a degraded biomedical image, the pipeline sequentially estimates a denoised observation, a scattering map, a physics-based prior map, and the final restored image. As shown in Fig.~\ref{fig:overall_system}, the framework consists of three stages: 1) initial noise suppression using DFN-Net, 2) scattering-map estimation and prior-map recovery using ASAP, and 3) high-fidelity prior-map refinement using GF-Net.

For clarity, the branch composed of ASAP and GF-Net is denoted as PP-Net$_{\mathrm{P}}$, representing the physical-prior-guided restoration branch. The complete restoration method with DFN-Net, ASAP, and GF-Net is denoted as PP-Net.
\subsection{System-Level Architecture and Pipeline Overview}
\label{subsec:system_overview}

To satisfy the low-latency and privacy-preserving requirements of IoMT applications, the proposed framework is designed for flexible deployment on either local edge devices or remote servers. As illustrated in Fig.~\ref{fig:overall_system}(a), raw biomedical images acquired by devices such as endoscopes and microscopes are first fed into the restoration pipeline. Depending on the application scenario, the pipeline can run on a local RK3588 edge node for edge-side enhancement or on a remote server for collaborative medical analysis.

At the algorithmic level, the proposed framework adopts a cascaded \emph{network-prior-network} architecture, as shown in Fig.~\ref{fig:overall_system}(b). Let $F(x,y)$ denote the degraded biomedical input. The overall restoration process is summarized as follows:

\newcommand{\boxHa}{0.55cm}
\newcommand{\gapWa}{0.42cm}      
\newcommand{\arrowLWa}{0.75pt}   

\begin{center}
	\begin{tikzpicture}[
		>=Latex,
		node distance=0cm,
		every node/.style={
			font=\rmfamily\normalsize
		},
		proc/.style={
			draw,
			rounded corners=2pt,
			minimum height=\boxHa,
			align=center,
			inner xsep=2pt,
			inner ysep=1pt
		}
		]
		
		\node[proc, minimum width=1.25cm] (F) {$F(x,y)$};
		\node[proc, minimum width=1.25cm, right=\gapWa of F] (T) {$T(x,y)$};
		\node[proc, minimum width=1.25cm, right=\gapWa of T] (S) {$S(x,y)$};
		\node[proc, minimum width=1.85cm, right=\gapWa of S] (U) {$U_{\mathrm{map}}(x,y)$};
		\node[proc, minimum width=1.25cm, right=\gapWa of U] (Uh) {$\hat{U}(x,y)$};
		
		\draw[->, line width=\arrowLWa] (F) -- (T);
		\draw[->, line width=\arrowLWa] (T) -- (S);
		\draw[->, line width=\arrowLWa] (S) -- (U);
		\draw[->, line width=\arrowLWa] (U) -- (Uh);
		
	\end{tikzpicture}
\end{center}
Here, $T(x,y)$ denotes the denoised intermediate image, $S(x,y)$ denotes the scattering map estimated by ASAP, $U_{\mathrm{map}}(x,y)$ denotes the physics-based prior map recovered from $T(x,y)$ and $S(x,y)$, and $\hat{U}(x,y)$ denotes the final restored image generated by GF-Net. Under this formulation, the clean synthetic setting uses PP-Net$_{\mathrm{P}}$ for physical-prior-guided restoration, whereas the noisy synthetic and biomedical transfer settings use the complete PP-Net with DFN-Net.

The first stage suppresses sensor-induced noise, the second stage estimates the scattering map and recovers a physics-based prior map, and the third stage refines this prior map using GF-Net. This progressive design improves interpretability and reduces the learning burden imposed on the final refinement stage.
\subsection{Biomedical Scattered Light Model}
\label{subsec:biomedical_model}

To guide the restoration process, we formulate a task-oriented degradation model for biomedical light scattering. In natural image dehazing, the classical atmospheric scattering model is commonly written as
\begin{equation}
	\label{eq:atmospheric_scattering}
	I(x,y)=J(x,y)t(x,y)+A\left(1-t(x,y)\right),
\end{equation}
where $I(x,y)$ is the observed degraded image, $J(x,y)$ is the latent clear image, $t(x,y)$ is the transmission map, and $A$ denotes the global atmospheric light.

Although biomedical scattering shares certain similarities with atmospheric haze, its imaging conditions are substantially different. In many biomedical imaging systems, illumination is provided by localized active light sources rather than global ambient illumination. Under this assumption, the scattering process can be reformulated using a scattering map $S(x,y)=1-t(x,y)$ as
\begin{equation}
	\label{eq:reformulated_scattering}
	I(x,y)=J(x,y)\left(1-S(x,y)\right)+S(x,y).
\end{equation}

To model sensor noise in low-light biomedical acquisition, we adopt the following task-oriented degradation model:
\begin{equation}
	\label{eq:proposed_degradation}
	F(x,y)=\alpha U(x,y)\left(1-S(x,y)\right)+S(x,y)+N(x,y),
\end{equation}
where $F(x,y)\in[0,1]$ denotes the observed degraded biomedical image, $U(x,y)$ denotes the latent clean image, $S(x,y)$ denotes the scattering map, $\alpha>0$ is a scaling coefficient used for illumination and attenuation compensation, and $N(x,y)$ denotes sensor noise. For analytical simplicity, $N(x,y)$ is represented as a generic additive noise term. In the subsequent noisy synthetic training stage, mixed Poisson--Gaussian noise is adopted to approximate practical noise characteristics in biomedical acquisition.

This formulation explicitly decomposes biomedical degradation into additive sensor noise and multiplicative scattering corruption, thereby motivating the progressive restoration strategy adopted in this work.

\subsection{Initial Noise Suppression via DFN-Net}
\label{subsec:ffdnet}

According to \eqref{eq:proposed_degradation}, the degraded biomedical image contains both scattering corruption and sensor-induced noise. Since sensor noise may distort local statistics and reduce the reliability of subsequent prior estimation, a dedicated denoising stage is introduced before scattering estimation.

To meet edge-side efficiency requirements, we employ DFN-Net, implemented using FFDNet~\cite{FFDNet}, as a lightweight feed-forward denoising front end. As shown in Fig.~\ref{fig:overall_system}(c), DFN-Net adopts a fully convolutional structure. The network first applies a $3\times3$ convolution followed by a PReLU activation to extract shallow features. The resulting feature maps are then processed by a sequence of dilated convolutional blocks with dilation factors $D\in\{1,2,4,8\}$.

The dilated design enlarges the receptive field with limited computational overhead. In addition, a residual learning strategy is adopted to encourage the network to focus on the noise component. After feature reconstruction using a final $3\times3$ convolution and PReLU activation, the denoised image $T(x,y)$ is obtained and passed to the subsequent prior estimation stage.

\subsection{Advanced Scattering Adaptive Prior via ASAP}
\label{subsec:msadcp}

After the denoising stage, the intermediate image $T(x,y)$ can be approximated as
\begin{equation}
	\begin{aligned}
		T(x,y) &\approx F(x,y)-N(x,y) \\
		&= \alpha U(x,y)\left(1-S(x,y)\right)+S(x,y),
	\end{aligned}
	\label{eq:denoised_observation}
\end{equation}
which provides a more reliable basis for scattering estimation.

\begin{figure}[htbp]
	\centering
	\includegraphics[width=\linewidth]{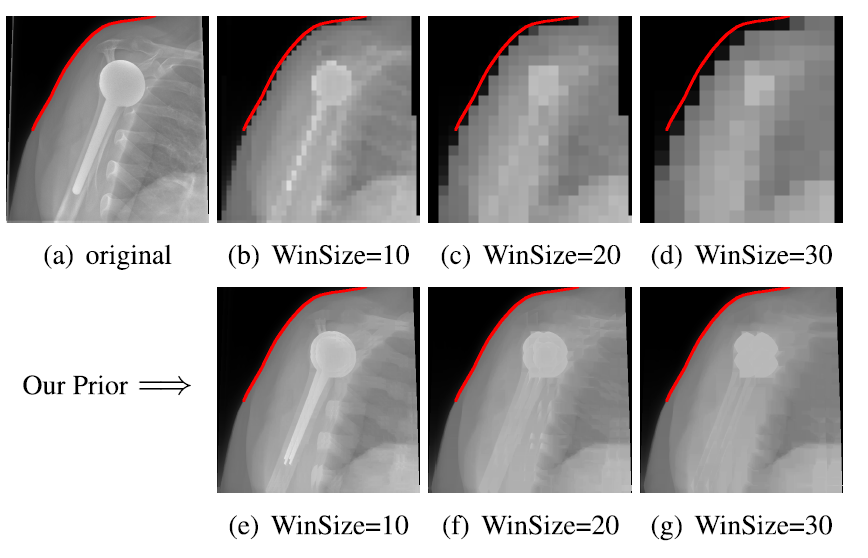}
	\caption{Visual comparison of scattering-estimation artifacts. The traditional fixed-window DCP introduces severe edge-shifting artifacts, whereas the proposed ASAP achieves more accurate edge preservation.}
	\label{fig:edge_shifting}
\end{figure}

Classical dark channel prior methods usually adopt a fixed local window to estimate transmission-related statistics~\cite{DCP}. However, biomedical images often contain complex structures, fine tissue boundaries, and heterogeneous textures. Under such conditions, fixed-window estimation may introduce boundary leakage and structural artifacts. To address this issue, we propose ASAP, which is derived from a multi-scale adaptive dark-channel estimation mechanism, as illustrated in Fig.~\ref{fig:overall_system}(d).

\begin{figure}[htbp]
	\centering
	\includegraphics[width=\linewidth]{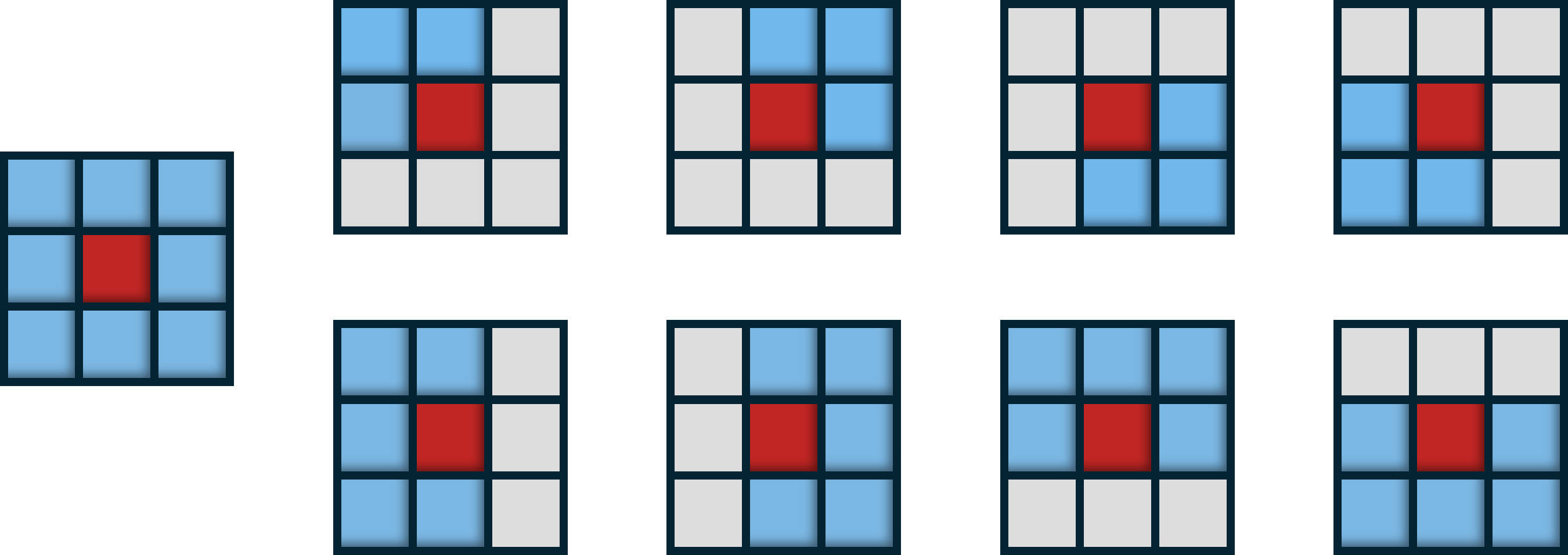}
	\caption{Illustration of the nine window morphologies evaluated in the proposed dual-adaptive mechanism. Full, half, and quarter spatial configurations are considered to reduce boundary leakage during scattering estimation.}
	\label{fig:window_shapes}
\end{figure}

Instead of using a single fixed patch, ASAP evaluates multiple candidate windows with different scales and spatial morphologies for each target pixel. Specifically, the method considers multiple window sizes and nine spatial configurations, including full, half, and quarter window patterns. By adaptively selecting the most suitable local neighborhood, ASAP improves the robustness of scattering estimation while preserving local structural boundaries.

Based on the optimal local window $W_{\mathrm{opt}}(x,y)$, the coarse scattering map $S_{0}(x,y)$ is estimated as
\begin{equation}
	\label{eq:coarse_scattering}
	S_{0}(x,y)=\min_{q\in W_{\mathrm{opt}}(x,y)}
	\left(
	\min_{c\in\{r,g,b\}} T^{c}(q)
	\right),
\end{equation}
where $T^{c}(q)$ denotes the value of the $c$-th channel of the denoised image $T(x,y)$ at pixel location $q$.

To further refine the scattering map while preserving structural discontinuities, we employ total-variation-guided refinement. The corresponding objective is formulated as
\begin{equation}
	\label{eq:tvgf_objective}
	\mathcal{E}(S)=\frac{1}{2}\left\|S-S_{0}\right\|_{2}^{2}
	+\lambda\left\|\nabla S\right\|_{1},
\end{equation}
where $\nabla$ is the gradient operator and $\lambda$ controls the regularization strength. By minimizing this objective, the refined scattering map $S(x,y)$ is obtained.

Once $T(x,y)$ and $S(x,y)$ are available, the physics-based prior map is recovered as
\begin{equation}
	\label{eq:prior_map_recovery}
	U_{\mathrm{map}}(x,y)=\frac{1}{\alpha}
	\frac{T(x,y)-S(x,y)}{1-S(x,y)}.
\end{equation}
To keep the recovered prior map within a physically meaningful intensity range and avoid numerical instability, the scaling coefficient $\alpha$ is adaptively determined as
\begin{equation}
	\label{eq:alpha_definition}
	\alpha=\max_{x,y}
	\left(
	\frac{T(x,y)-S(x,y)}{1-S(x,y)}
	\right).
\end{equation}

The resulting prior map removes the dominant scattering component and provides a physically interpretable intermediate representation. GF-Net then refines this prior map by fusing it with the denoised observation.
\subsection{Guided Fusion via GF-Net}
\label{subsec:pgnafnet}

Although the physics-based prior map attenuates the dominant scattering effects and recovers the global structure, it may still be insufficient for restoring subtle local details in complex biomedical tissues. To further improve restoration fidelity, we introduce a deep refinement stage based on GF-Net, as shown in Fig.~\ref{fig:overall_system}(e).

GF-Net jointly exploits the denoised observation $T(x,y)$ and the prior map $U_{\mathrm{map}}(x,y)$. The two inputs are concatenated along the channel dimension and projected into a high-dimensional feature space through a $3\times3$ convolution. In this manner, the network can exploit both scattering-suppressed structural guidance and retained local texture cues.

To improve computational efficiency, GF-Net is built upon the nonlinear activation-free block (NAFB). Two operations are particularly important. First, the SimpleGate operation introduces nonlinearity through element-wise interactions between split feature channels. Second, simplified channel attention (SCA) recalibrates channel-wise feature importance using global average pooling and a lightweight $1\times1$ convolution.

At the network level, GF-Net adopts a four-level encoder--decoder architecture with skip connections. The hierarchical design enlarges the receptive field and facilitates multi-scale fusion, while the skip connections help preserve spatial information. In addition, a global residual connection is introduced so that the network learns a residual correction with respect to the prior map and the denoised observation.

Accordingly, the final restored image is expressed as
\begin{equation}
	\label{eq:final_restoration}
	\hat{U}(x,y)=U_{\mathrm{map}}(x,y)
	+F_{\mathrm{GF}}\big(T(x,y),U_{\mathrm{map}}(x,y)\big),
\end{equation}
where $F_{\mathrm{GF}}(\cdot)$ represents the nonlinear restoration function learned by GF-Net.

\subsection{Progressive Synthetic Training and Cross-Domain Transfer Strategy}
\label{subsec:training_strategy}

A major challenge in biomedical image restoration is the lack of paired degraded/clean data in real clinical environments. Instead of directly performing supervised training on biomedical images, we adopt a progressive synthetic training strategy and then transfer the trained model to real biomedical images for inference. This design is motivated by two considerations: paired scattering-free biomedical ground truth is generally unavailable, and the denoising front end is required only when scattering and sensor noise coexist.

To match the structure of the proposed framework, the training process is divided into two stages. In the first stage, PP-Net$_{\mathrm{P}}$ is trained for physical-prior-guided restoration under scattering-dominated degradation without explicit sensor noise. In the second stage, mixed synthetic noise is introduced to train the complete PP-Net for joint scattered-light removal and noise suppression. After this progressive two-stage training, the trained PP-Net is directly applied to real biomedical images for cross-domain inference and evaluation.
\subsubsection{Stage I: Synthetic Training of PP-Net$_{\mathrm{P}}$}
\label{subsubsec:stage1_training}

In the first stage, the model is trained on the ITS subset of RESIDE~\cite{RESIDE} to learn prior-guided restoration from paired synthetic data. Since this stage considers scattering-dominated degradation without explicit sensor noise, only ASAP and GF-Net are used. ASAP estimates the scattering map and recovers a physics-based prior map from the degraded input, and GF-Net refines the prior map to recover the restored image. The resulting physical-prior branch is denoted as PP-Net$_{\mathrm{P}}$.

Let $F_{\mathrm{syn}}$ denote the synthetic hazy input and $U_{\mathrm{syn}}$ denote the corresponding clean target image. The network is optimized using the Charbonnier loss
\begin{equation}
	\label{eq:lsup_stage1}
	\mathcal{L}_{\mathrm{sup}}=
	\sqrt{\left\|\hat{U}_{\mathrm{syn}}-U_{\mathrm{syn}}\right\|_{2}^{2}+\epsilon^{2}},
\end{equation}
where $\hat{U}_{\mathrm{syn}}$ is the restored output and $\epsilon$ is a small constant for numerical stability. This stage establishes physical-prior-guided restoration capability from paired synthetic data.
\subsubsection{Stage II: Synthetic Training of PP-Net Under Joint Noise-and-Scattering Degradation}
\label{subsubsec:stage2_training}

Although the first stage provides a useful initialization, practical biomedical image acquisition often involves both scattering degradation and sensor-induced noise. To simulate this condition, we add mixed Poisson--Gaussian noise to the ITS images and train the complete PP-Net pipeline. In this stage, DFN-Net suppresses sensor-induced noise, ASAP estimates the scattering map and recovers a physics-based prior map, and GF-Net refines the prior map by fusing it with the denoised observation. The mixed noise model approximates photon fluctuations and sensor noise commonly observed in biomedical imaging.

The noisy-stage training is optimized using the following supervised objective:
\begin{equation}
	\label{eq:lsup_stage2}
	\mathcal{L}_{\mathrm{sup}}^{\mathrm{noise}}=
	\sqrt{\left\|\hat{U}_{\mathrm{syn}}^{\mathrm{noise}}-U_{\mathrm{syn}}\right\|_{2}^{2}+\epsilon^{2}},
\end{equation}
where $\hat{U}_{\mathrm{syn}}^{\mathrm{noise}}$ denotes the restored output under noisy degradation. This stage enables PP-Net to jointly address scattering removal and noise suppression.

\subsubsection{Cross-Domain Transfer to Real Biomedical Images}
\label{subsubsec:cross_domain_transfer}

After the two-stage training process, the resulting PP-Net is directly transferred to real biomedical images for cross-domain inference and evaluation. Since paired biomedical ground truth is unavailable during training, this step is treated as cross-domain transfer rather than supervised adaptation.

In summary, the proposed training strategy is consistent with the progressive structure of the restoration framework. The first stage establishes prior-guided restoration capability through PP-Net$_{\mathrm{P}}$, while the second stage introduces DFN-Net to form the complete PP-Net and improve robustness under joint noise-and-scattering degradation. The final trained PP-Net is then transferred to real biomedical images for inference and evaluation.

\section{Experiments}
\label{sec:experiments}

This section evaluates the proposed framework from four aspects. First, we validate the effectiveness of the proposed ASAP module on multiple dehazing benchmarks. Second, we compare the proposed hybrid restoration framework with representative learning-based dehazing methods on paired synthetic datasets. Third, we investigate the robustness of the complete PP-Net under joint noise-and-scattering degradation. Finally, after progressive training on synthetic hazy and noisy-hazy data, we directly transfer the trained model to real biomedical images for qualitative analysis and no-reference image quality assessment.

\subsection{Experimental Setup}
\label{subsec:exp_setup}

Experiments are conducted on both synthetic dehazing benchmarks and real biomedical images. The evaluation is designed to examine the effectiveness of the proposed adaptive prior, the restoration capability of the hybrid framework, its robustness under noisy degradation, and its transferability to real biomedical imaging scenarios.

\paragraph{Prior evaluation on multi-domain benchmarks.}

To validate the effectiveness of ASAP, we compare it with representative physics-based dehazing methods on five datasets: SOTS, HSTS, I-HAZE, O-HAZE, and D-HAZY. These datasets cover synthetic and real haze conditions, as well as indoor and outdoor scenes, thereby providing a comprehensive evaluation of prior-based scattering estimation.

\paragraph{Hybrid pipeline evaluation on paired synthetic datasets.}
To evaluate the effectiveness of the proposed physical-prior-guided restoration framework, we conduct experiments on the RESIDE-ITS and RESIDE-6K datasets. These two paired datasets are used to assess whether PP-Net$_{\mathrm{P}}$ can outperform representative learning-based dehazing models under paired synthetic supervision.

\paragraph{Robustness evaluation under noisy degradation.}

To simulate the noisy biomedical imaging scenario considered in this work, we construct noisy synthetic data by injecting mixed Poisson--Gaussian noise into hazy images. The Poisson scaling factor and Gaussian noise level are set to $\lambda_p=30$ and $\sigma_g=0.05$, respectively. This setting is designed to assess the robustness of PP-Net when scattering degradation and sensor noise coexist.
\paragraph{Cross-domain evaluation on real biomedical images.}

To examine cross-domain transferability, we directly apply the trained model to multiple categories of real biomedical images. Since paired scattering-free biomedical ground truth is unavailable and no target-domain fine-tuning is performed, the biomedical experiments are used mainly for qualitative analysis and no-reference image quality assessment.

For synthetic datasets with paired references, PSNR and SSIM are adopted as full-reference evaluation metrics. For real biomedical images, NIQE and BRISQUE are used as no-reference quality metrics, where lower values indicate better perceptual quality.

\subsubsection{Implementation Details}
\label{subsubsec:implementation}

All experiments are conducted on a workstation equipped with four NVIDIA RTX 6000 Ada GPUs and an Intel Core i9-14900K CPU. The proposed model is implemented in PyTorch. Unless otherwise specified, the input patch size is set to $256\times256$, the batch size is set to 32, and the Adam optimizer is used for training. The initial learning rate is set to $1\times10^{-4}$ and is updated using the CosineAnnealingLR schedule. The total number of training epochs is set to 500.

\subsubsection{Baseline Methods}
\label{subsubsec:baselines}

To provide comprehensive comparisons, we consider two categories of baseline methods. For evaluating ASAP, we compare it with representative physics-based dehazing methods, including FVR~\cite{FVR}, DCP~\cite{DCP}, CAP~\cite{CAP}, MR~\cite{MR}, CEP~\cite{CEP}, CC~\cite{CC}, NLBF~\cite{NLBF}, SLP~\cite{SLP}, ROP+~\cite{ROP}, ALSP~\cite{ALSP}, GLP~\cite{GLP}, and IHDCP~\cite{IHDCP}. For evaluating the proposed hybrid framework, we further compare it with representative learning-based dehazing methods, including MSCNN~\cite{MSCNN}, AOD-Net~\cite{AOD-Net}, GFN~\cite{GFN}, MSBDN~\cite{MSBDN}, PFDN~\cite{PFDN}, FFA-Net~\cite{FFA-Net}, TBN~\cite{TBN}, CDVA~\cite{CDVA}, IDB~\cite{IDB}, and MPMF-Net~\cite{MPMF}. For the noisy degradation setting, the publicly available models of the compared learning-based methods are directly tested on the same noisy synthetic inputs without additional retraining, unless otherwise specified. For the W2S biomedical evaluation, we additionally compare with DCP~\cite{DCP}, HDCP~\cite{HDCP}, QDCP~\cite{QDCP}, and Dark~\cite{Dark} as representative prior-based or biomedical image enhancement baselines.

\subsubsection{Training and Transfer Setting}
\label{subsubsec:training_protocol}

The experimental protocol is consistent with the progressive synthetic training strategy described in Section~\ref{subsec:training_strategy}. Specifically, PP-Net$_{\mathrm{P}}$ is first trained on paired synthetic hazy data to learn physical-prior-guided restoration without explicit sensor noise. Then, the complete PP-Net is trained on noisy synthetic data to handle joint noise-and-scattering degradation. After two-stage training, PP-Net is directly applied to real biomedical images for cross-domain inference and evaluation.

\subsection{Evaluation of the Proposed ASAP}
\label{subsec:msadcp_eval}

We first evaluate ASAP on five benchmark datasets, including SOTS, HSTS, I-HAZE, O-HAZE, and D-HAZY, by comparing it with representative physics-based dehazing methods.
\begin{table*}[htbp]
	\centering
	\caption{Quantitative comparison between the proposed ASAP and representative physics-based dehazing methods on five benchmark datasets. Higher PSNR and SSIM indicate better restoration quality. {\color{red}\textbf{Red}}, {\color{darkgreen}\textbf{green}}, and {\color{blue}\textbf{blue}} denote the {\color{red}\textbf{best}}, {\color{darkgreen}\textbf{second-best}}, and {\color{blue}\textbf{third-best}} results, respectively.}
	\label{tab:msadcp_sots_dhazy}
	\scriptsize
	\setlength{\tabcolsep}{3.5pt}
	\renewcommand{\arraystretch}{1.10}
	\resizebox{\textwidth}{!}{%
		\begin{tabular}{c|cc|cc|cc|cc|cc}
			\hline
			\hline
			Method
			& \multicolumn{2}{c|}{SOTS \cite{RESIDE}}
			& \multicolumn{2}{c|}{HSTS \cite{RESIDE}}
			& \multicolumn{2}{c|}{I-HAZE \cite{I-HAZE}}
			& \multicolumn{2}{c|}{O-HAZE \cite{O-HAZE}}
			& \multicolumn{2}{c}{D-HAZY \cite{D-HAZE}} \\
			\cline{2-11}
			& SSIM$\uparrow$ & PSNR$\uparrow$
			& SSIM$\uparrow$ & PSNR$\uparrow$
			& SSIM$\uparrow$ & PSNR$\uparrow$
			& SSIM$\uparrow$ & PSNR$\uparrow$
			& SSIM$\uparrow$ & PSNR$\uparrow$ \\
			\hline
			(ICCV'09) FVR \cite{FVR}      & 0.7388 & 13.2622 & 0.7002 & 10.7931 & 0.4823 & 10.0210 & 0.2543 & 13.6041 & 0.8051 & 14.3512 \\
			(TPAMI'11) DCP \cite{DCP}      & 0.8028 & 14.6035 & 0.7942 & 13.7737 & 0.5361 & 10.2095 & {\color{darkgreen}\textbf{0.6250}} & 15.0083 & {\color{darkgreen}\textbf{0.8312}} & {\color{blue}\textbf{15.0923}} \\
			(TIP'15) CAP \cite{CAP}        & 0.8129 & 18.5728 & 0.8391 & {\color{blue}\textbf{19.7211}} & 0.6602 & 13.6274 & 0.3247 & 15.3071 & 0.7264 & 13.6844 \\
			(TIP'18) MR \cite{MR}        & 0.8421 & 17.4497 & 0.8267 & 15.8208 & 0.6160 & 10.9294 & 0.3374 & 15.7883 & 0.7951 & 14.3503 \\
			(TIP'18) CEP \cite{CEP}        & 0.7200 & 13.9373 & 0.7209 & 13.7334 & 0.5275 & 10.5288 & 0.4791 & 13.3644 & 0.7590 & 14.3906 \\
			(TCSVT'20) CC \cite{CC}      & 0.8703 & 18.3954 & 0.8523 & 17.2227 & 0.5116 & 9.3274 & 0.3636 & 15.3048 & 0.7870 & {\color{darkgreen}\textbf{15.3652}} \\
			(TIP'20) NLBF \cite{NLBF}      & 0.5933 & 13.9673 & 0.6024 & 13.1347 & 0.4181 & 10.5647 & 0.2572 & 13.7174 & 0.7053 & 13.9485 \\
			(TIP'23) SLP \cite{SLP}        & {\color{blue}\textbf{0.8758}} & {\color{blue}\textbf{19.8575}} & {\color{blue}\textbf{0.8533}} & 19.2735 & {\color{blue}\textbf{0.6763}} & 13.2467 & {\color{red}\textbf{0.6369}} & {\color{blue}\textbf{16.1066}} & {\color{blue}\textbf{0.8293}} & 14.3466 \\
			(TPAMI'23) ROP+ \cite{ROP}     & 0.5924 & 11.2562 & 0.6016 & 13.0116 & 0.5086 & {\color{blue}\textbf{15.1343}} & 0.4225 & 13.6476 & 0.5160 & 11.9830 \\
			(TIP'25) ALSP \cite{ALSP}      & 0.7932 & 16.8118 & 0.8075 & 17.2046 & 0.5700 & 12.2069 & 0.3464 & 12.9536 & 0.4871 & 10.4283 \\
			(TMM'26) GLP \cite{GLP}       & 0.8408 & 18.5458 & 0.7864 & 18.7565 & 0.5705 & 15.3048 & 0.3934 & {\color{darkgreen}\textbf{17.1895}} & 0.7869 & 14.4929 \\
			(TIP'26) IHDCP \cite{IHDCP}         & {\color{darkgreen}\textbf{0.8941}} & {\color{darkgreen}\textbf{20.8982}} & {\color{darkgreen}\textbf{0.9122}} & {\color{darkgreen}\textbf{21.5930}} & {\color{red}\textbf{0.7619}} & {\color{darkgreen}\textbf{16.6717}} & 0.3945 & 15.4133 & 0.7292 & 13.4291 \\
			\textbf{(Ours) ASAP}     & {\color{red}\textbf{0.9188}} & {\color{red}\textbf{23.2113}} & {\color{red}\textbf{0.9165}} & {\color{red}\textbf{22.0553}} & {\color{darkgreen}\textbf{0.6996}} & {\color{red}\textbf{16.7522}} & {\color{blue}\textbf{0.5882}} & {\color{red}\textbf{17.5845}} & {\color{red}\textbf{0.8505}} & {\color{red}\textbf{16.1583}} \\
			\hline
			\hline
		\end{tabular}%
	}
\end{table*}

Table~\ref{tab:msadcp_sots_dhazy} reports the quantitative comparison. The proposed ASAP achieves the best overall performance on most datasets, particularly on SOTS, HSTS, O-HAZE, and D-HAZY. On I-HAZE, ASAP remains competitive and achieves the highest PSNR. These results indicate that the proposed adaptive prior provides more reliable scattering estimation than conventional handcrafted priors across diverse haze conditions.

\begin{figure*}[t]
	\centering
	\setlength{\tabcolsep}{2pt}
	\renewcommand{\arraystretch}{1.0}
	
	\newcommand{\imgw}{0.13\textwidth}
	\newcommand{\rlabel}[1]{\adjustbox{valign=m}{\rotatebox[origin=c]{90}{\scriptsize\textbf{#1}}}}
	\newcommand{\chead}[1]{\makebox[\imgw][c]{\footnotesize\textbf{#1}}}
	\newcommand{\cheadtwo}[2]{\makebox[\imgw][c]{\shortstack{\footnotesize\textbf{#1}\\[-0.4ex]\footnotesize\textbf{#2}}}}
	
	\begin{adjustbox}{max totalsize={\textwidth}{0.95\textheight},center}
		\begin{tabular}{c ccccccc}
			& \chead{Hazy} &
			\chead{DCP\cite{DCP}} &
			\chead{IHDCP\cite{IHDCP}} &
			\chead{ROP+\cite{ROP}} &
			\chead{SLP\cite{SLP}} &
			\chead{ASAP (Ours)} &
			\chead{GT} \\
						\rlabel{SOTS} &
			\adjustbox{valign=m}{\includegraphics[width=\imgw]{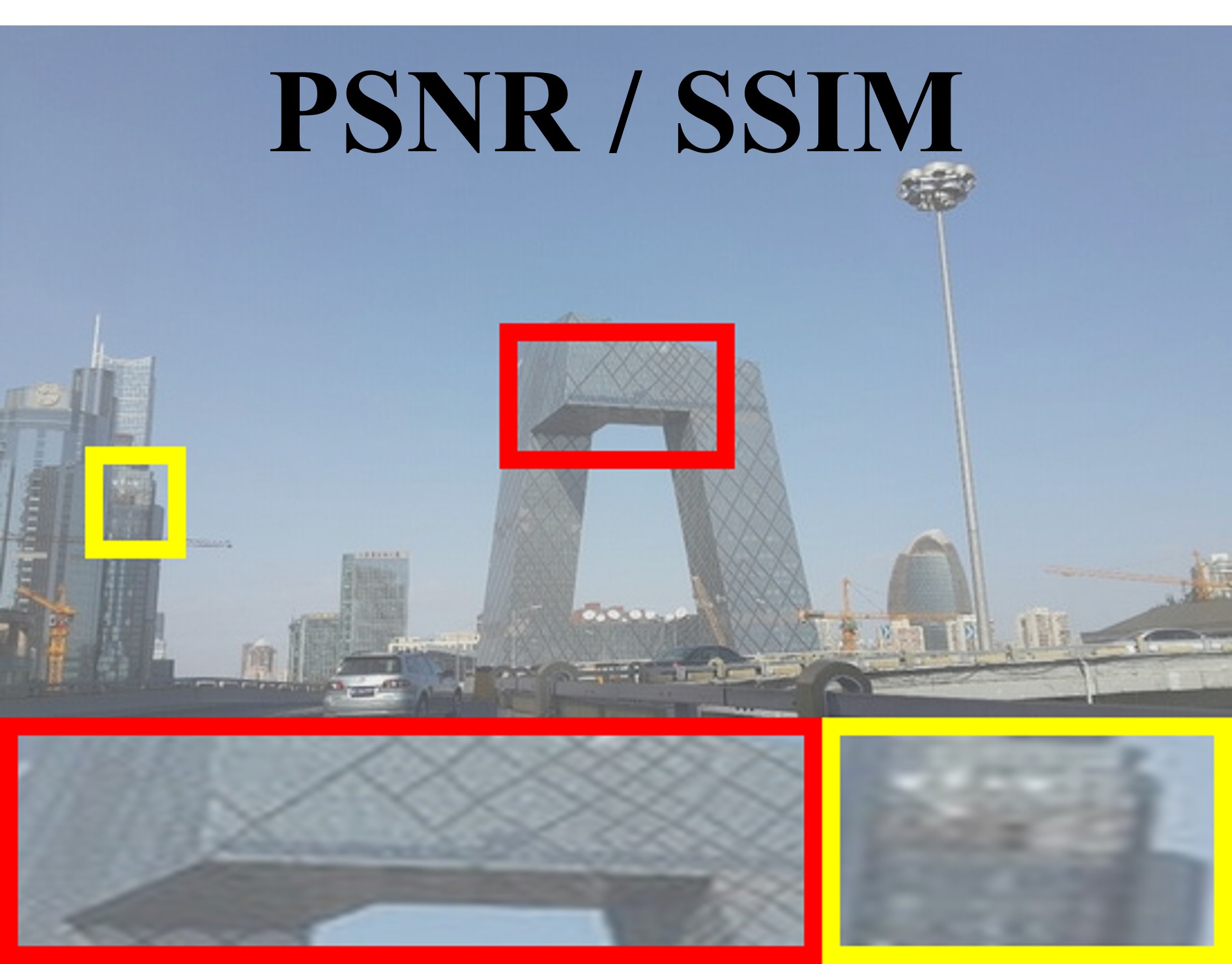}} &
			\adjustbox{valign=m}{\includegraphics[width=\imgw]{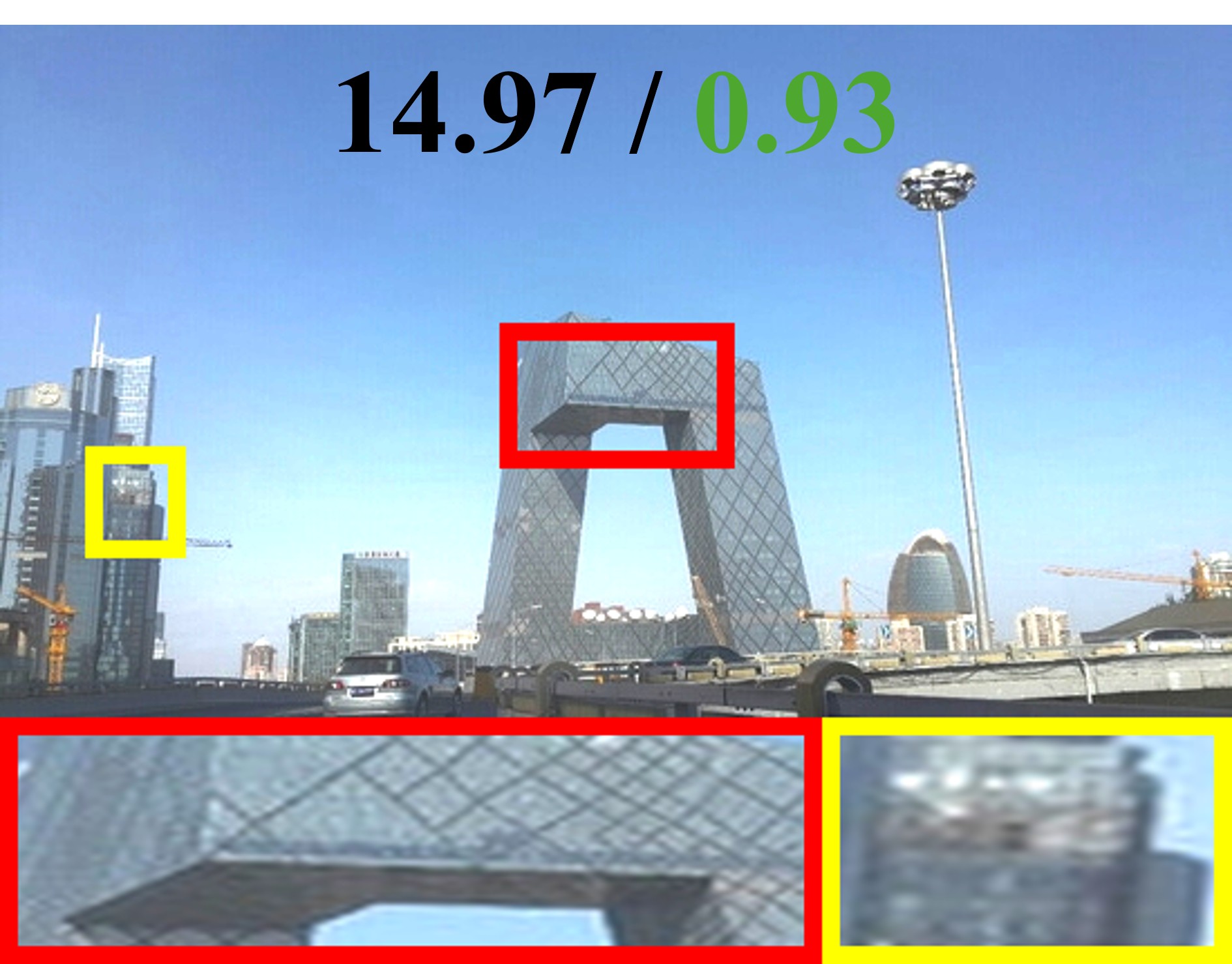}} &
			\adjustbox{valign=m}{\includegraphics[width=\imgw]{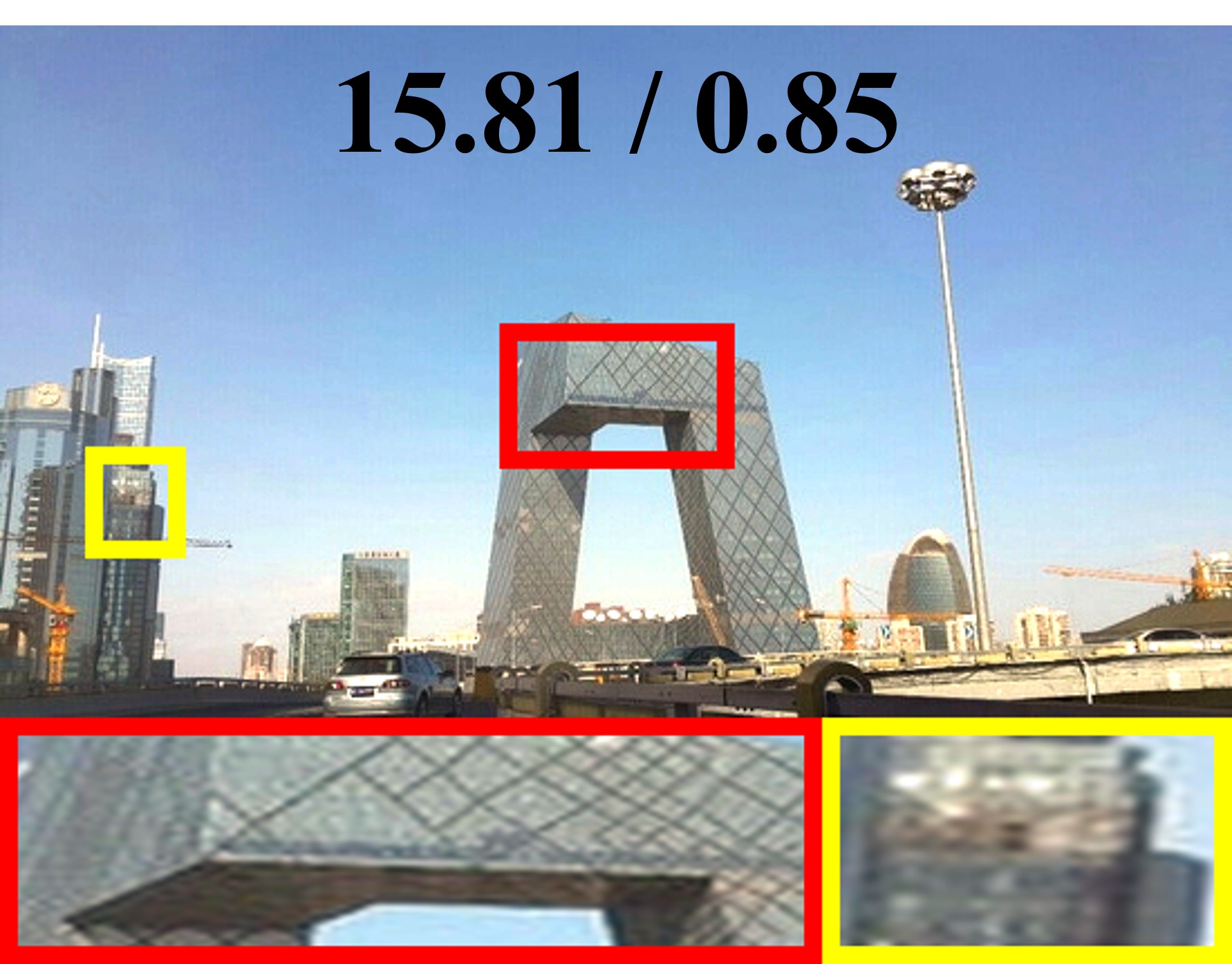}} &
			\adjustbox{valign=m}{\includegraphics[width=\imgw]{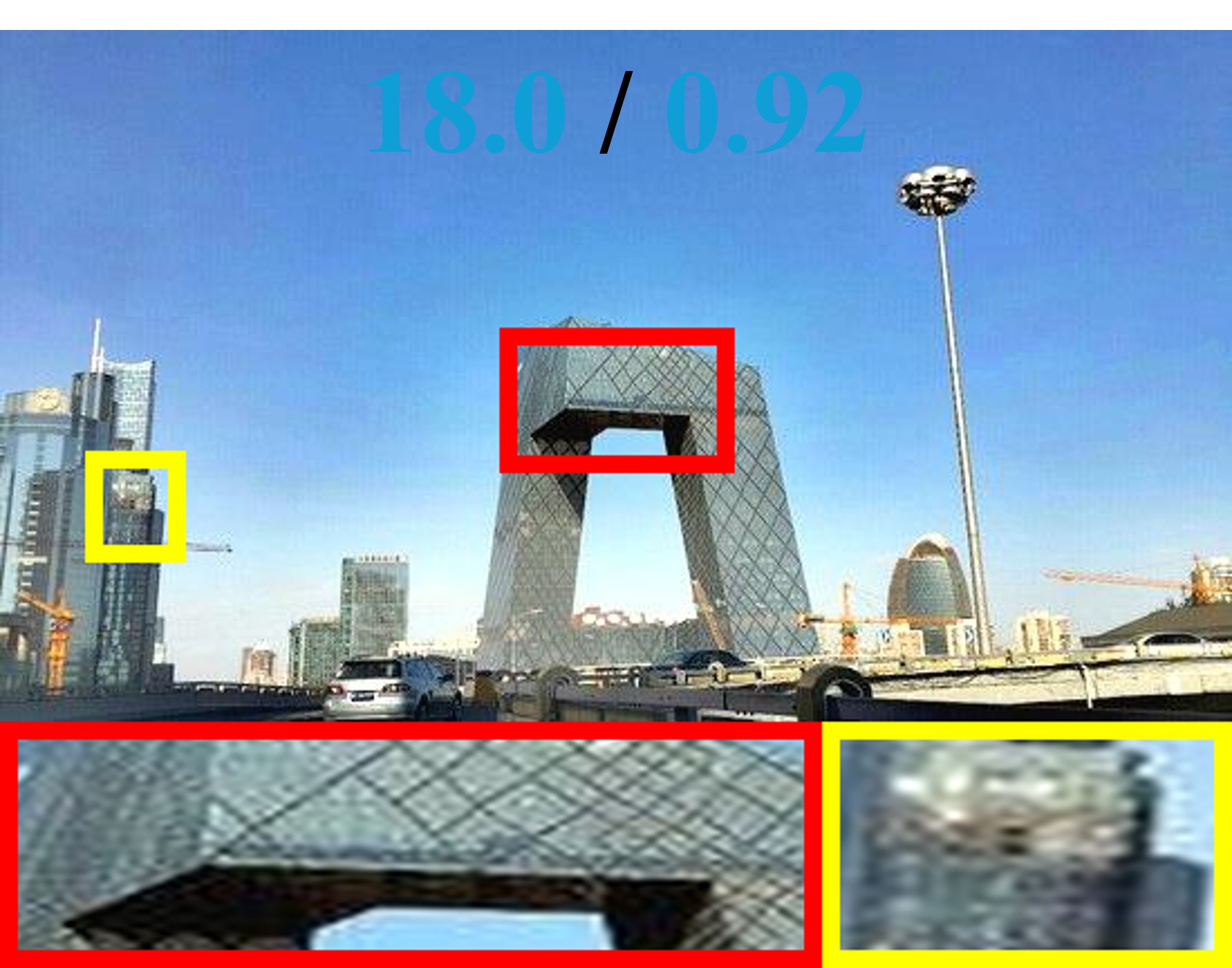}} &
			\adjustbox{valign=m}{\includegraphics[width=\imgw]{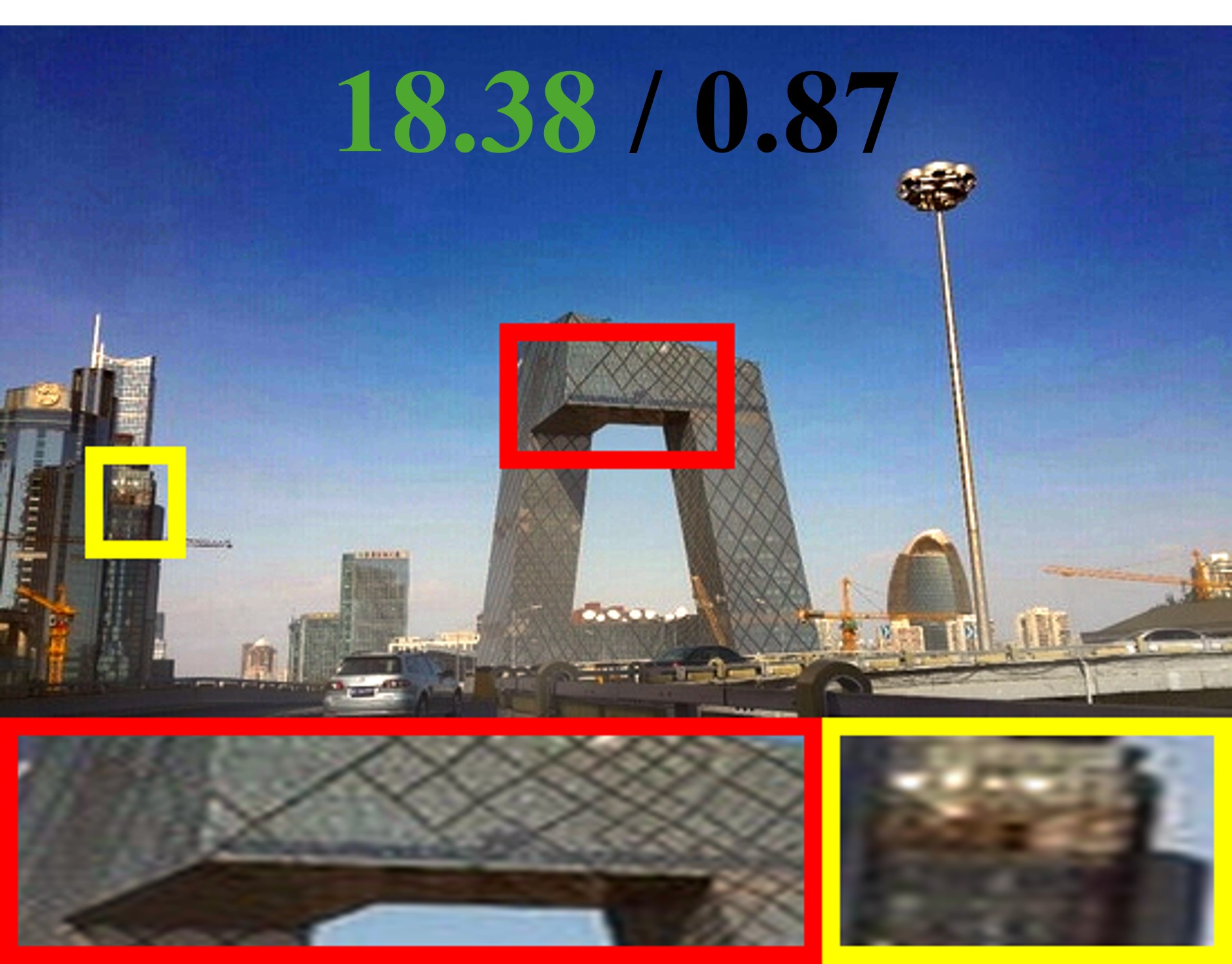}} &
			\adjustbox{valign=m}{\includegraphics[width=\imgw]{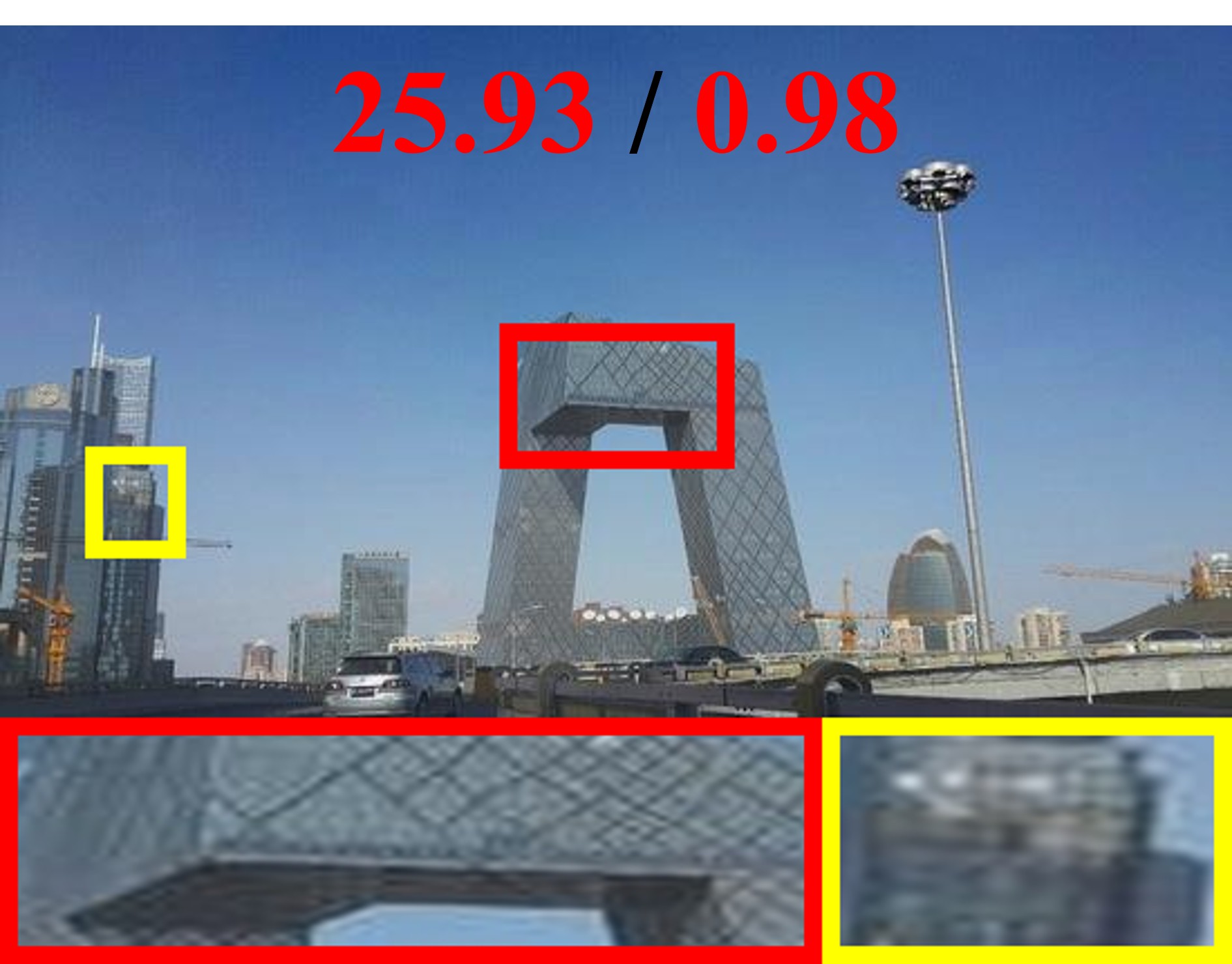}} &
			\adjustbox{valign=m}{\includegraphics[width=\imgw]{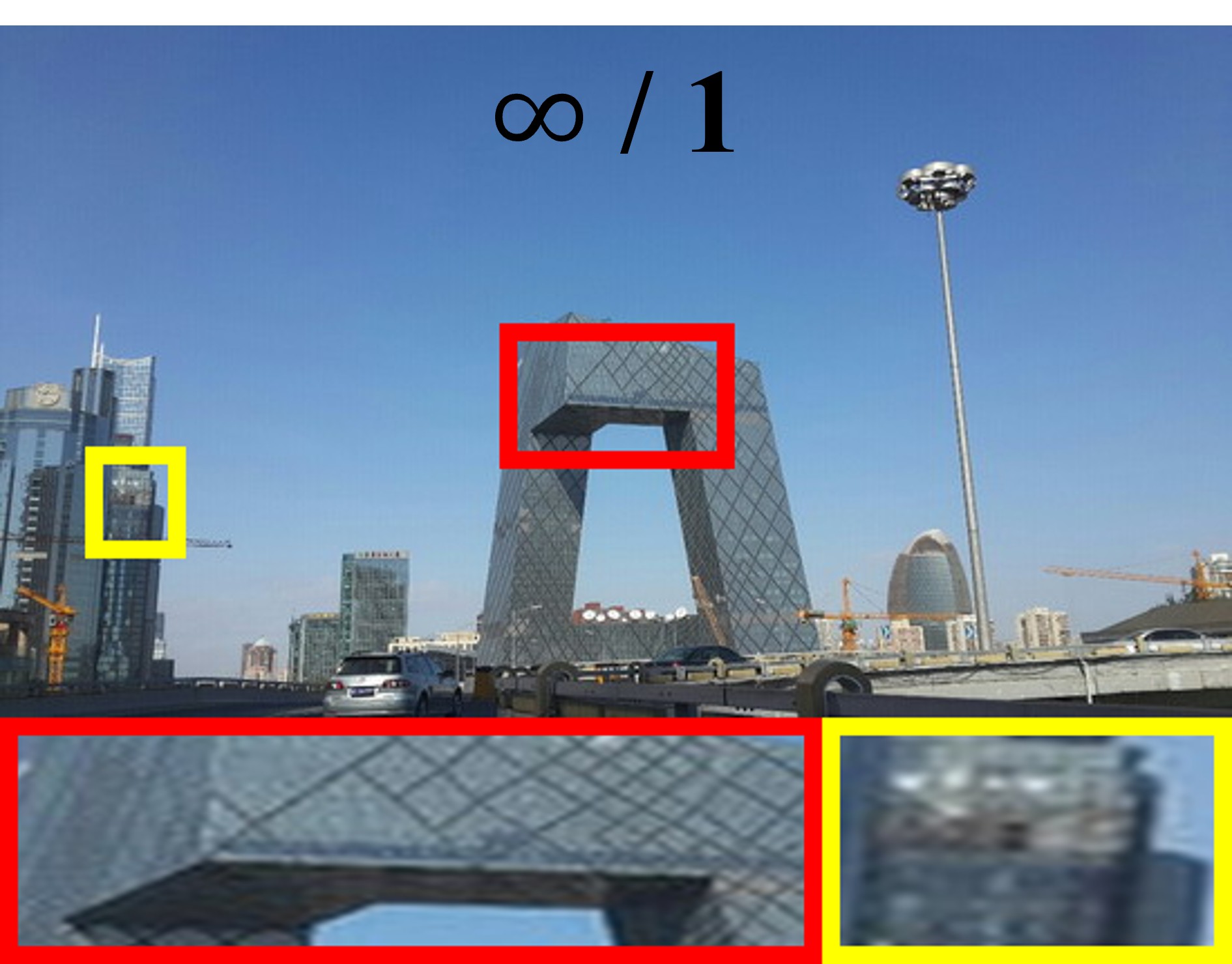}} \\
			
			\rlabel{HSTS} &
			\adjustbox{valign=m}{\includegraphics[width=\imgw]{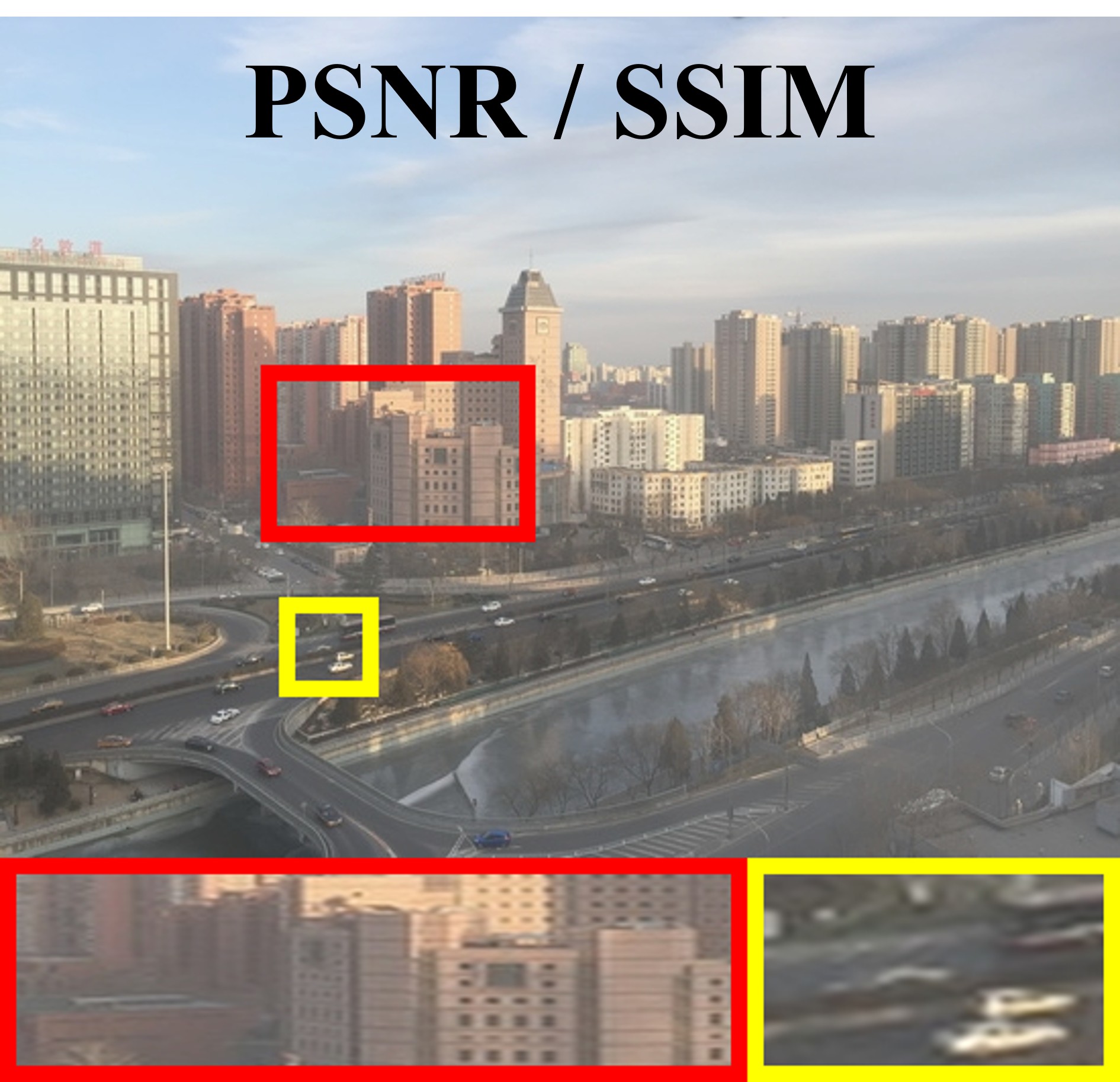}} &
			\adjustbox{valign=m}{\includegraphics[width=\imgw]{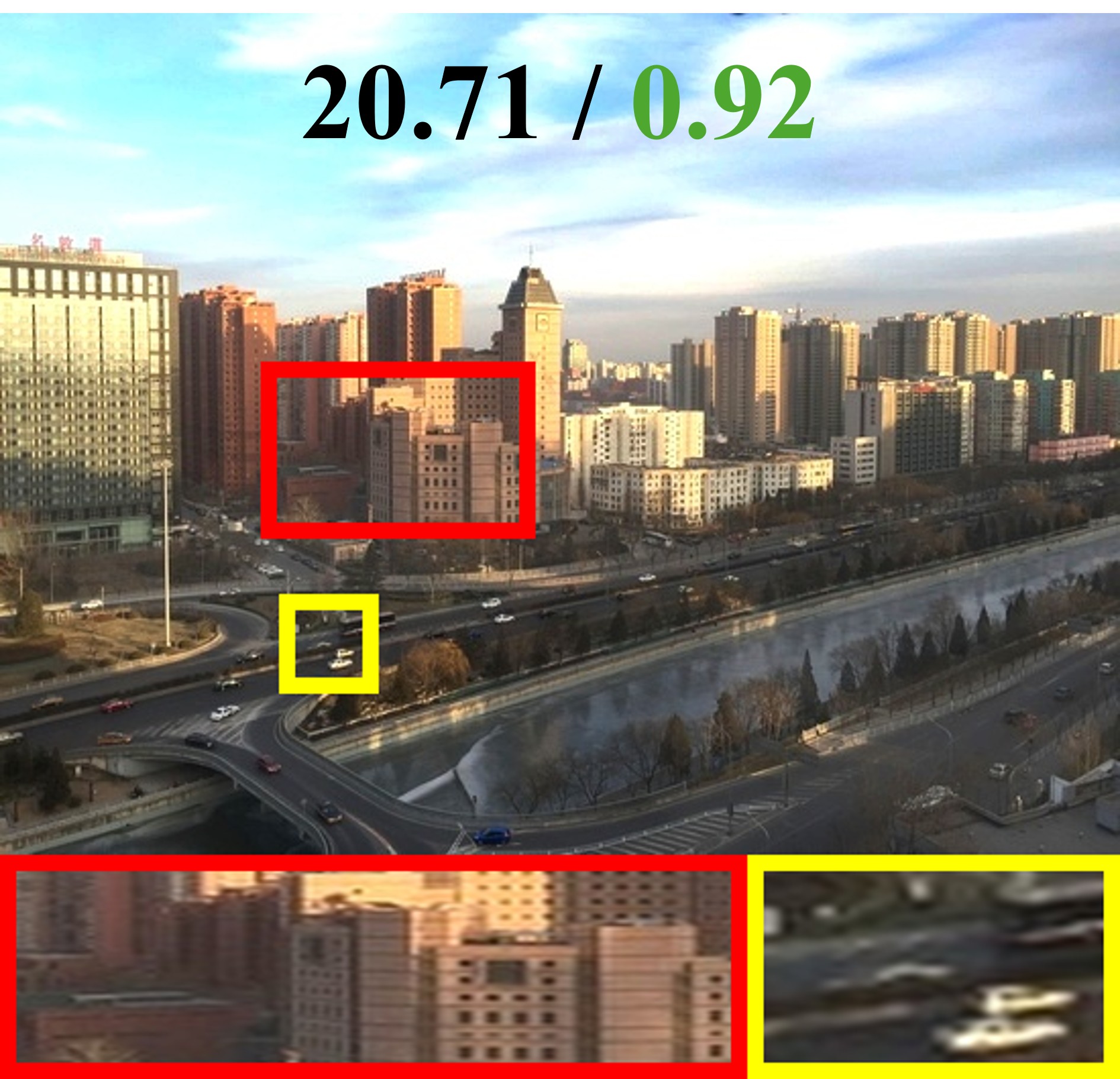}} &
			\adjustbox{valign=m}{\includegraphics[width=\imgw]{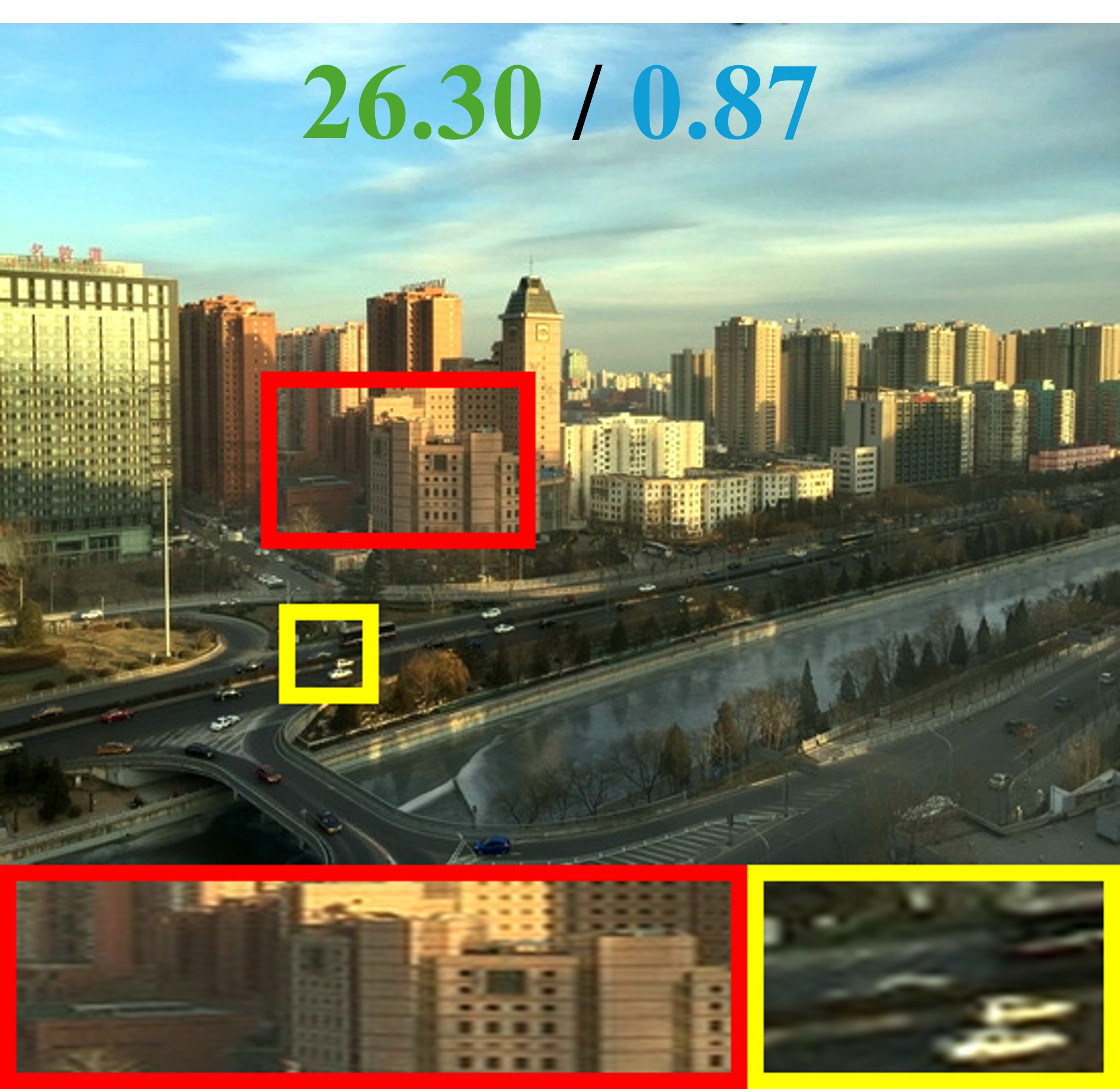}} &
			\adjustbox{valign=m}{\includegraphics[width=\imgw]{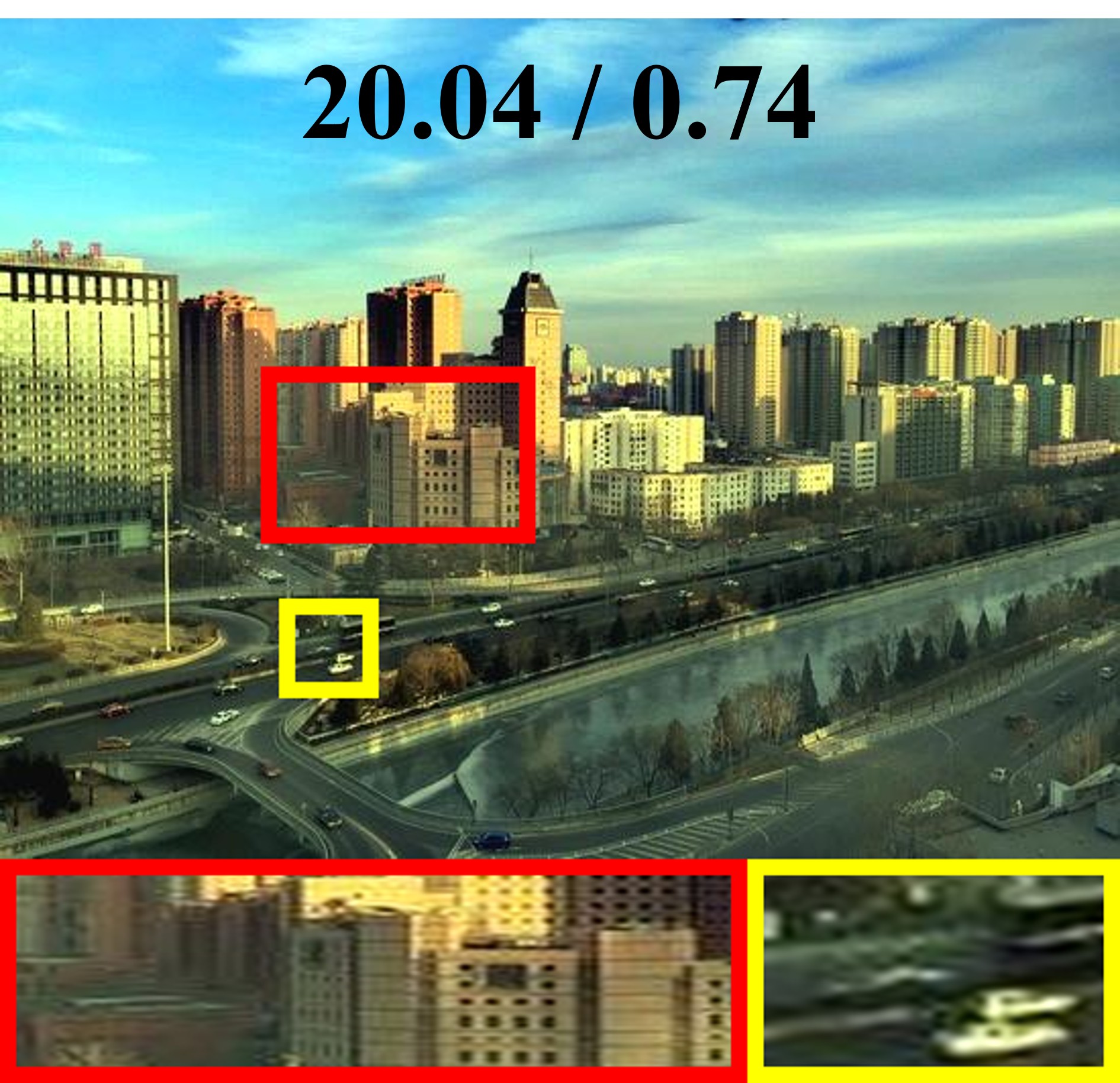}} &
			\adjustbox{valign=m}{\includegraphics[width=\imgw]{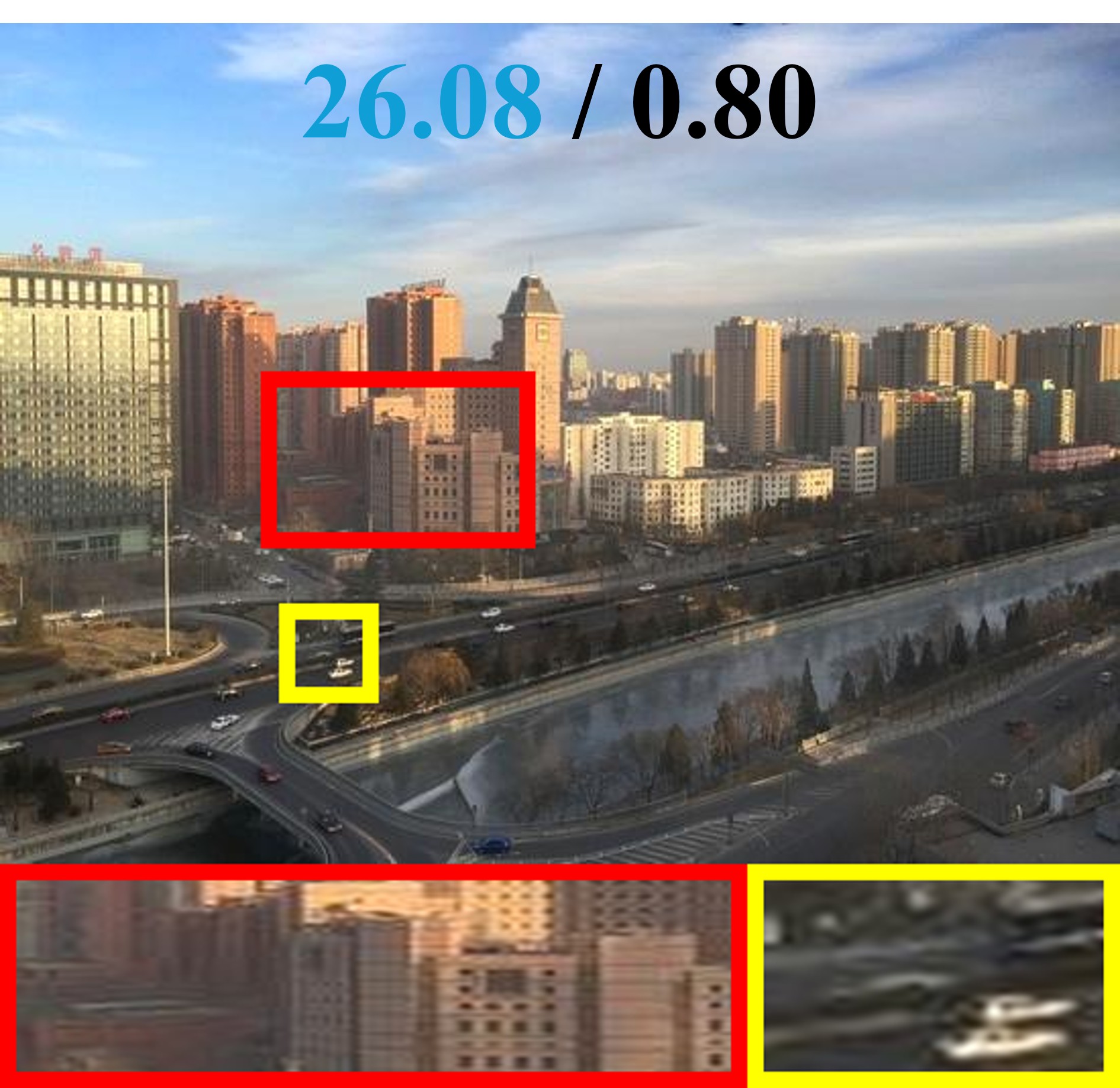}} &
			\adjustbox{valign=m}{\includegraphics[width=\imgw]{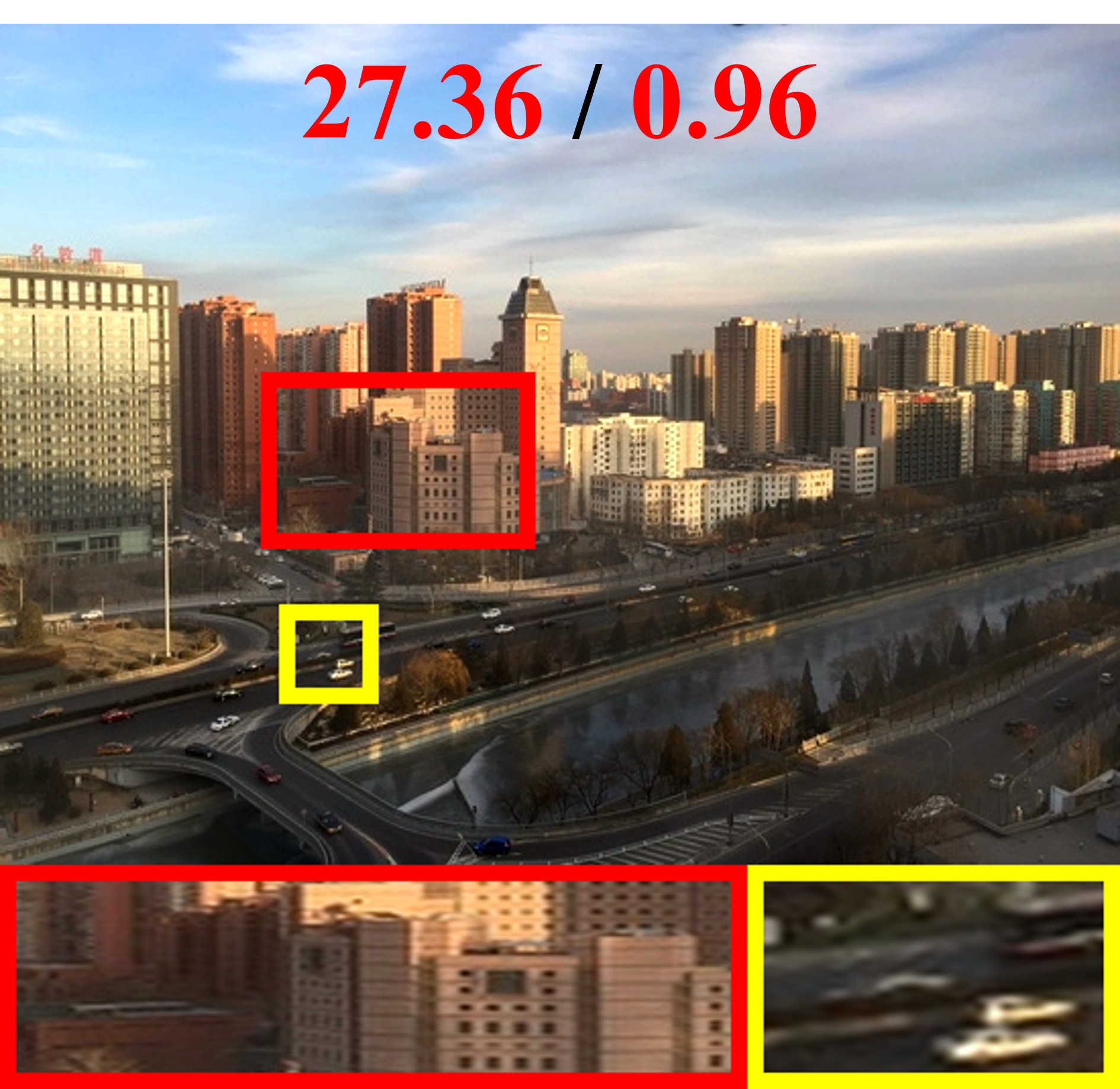}} &
			\adjustbox{valign=m}{\includegraphics[width=\imgw]{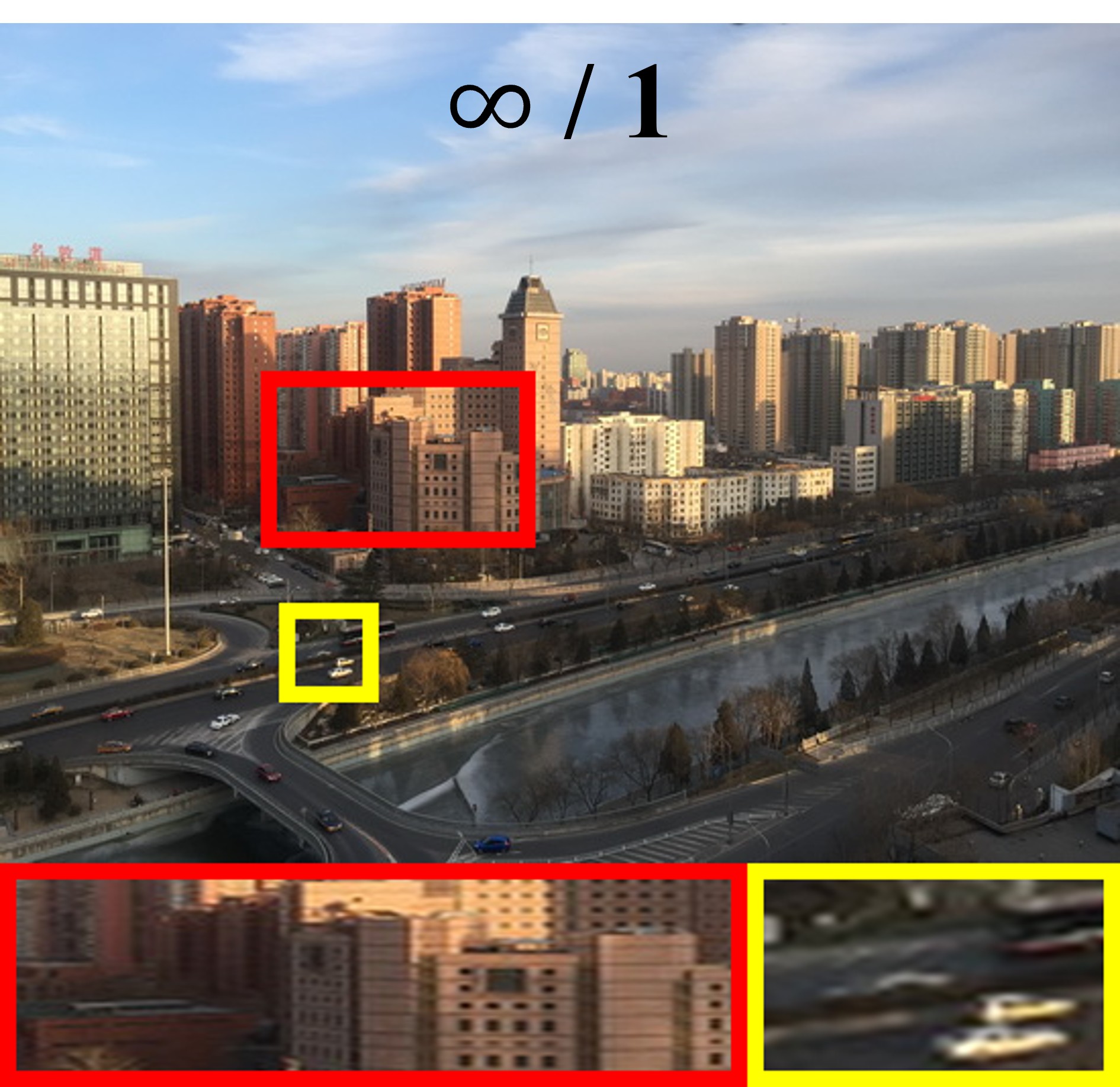}} \\
			
			\rlabel{I-HAZE} &
			\adjustbox{valign=m}{\includegraphics[width=\imgw]{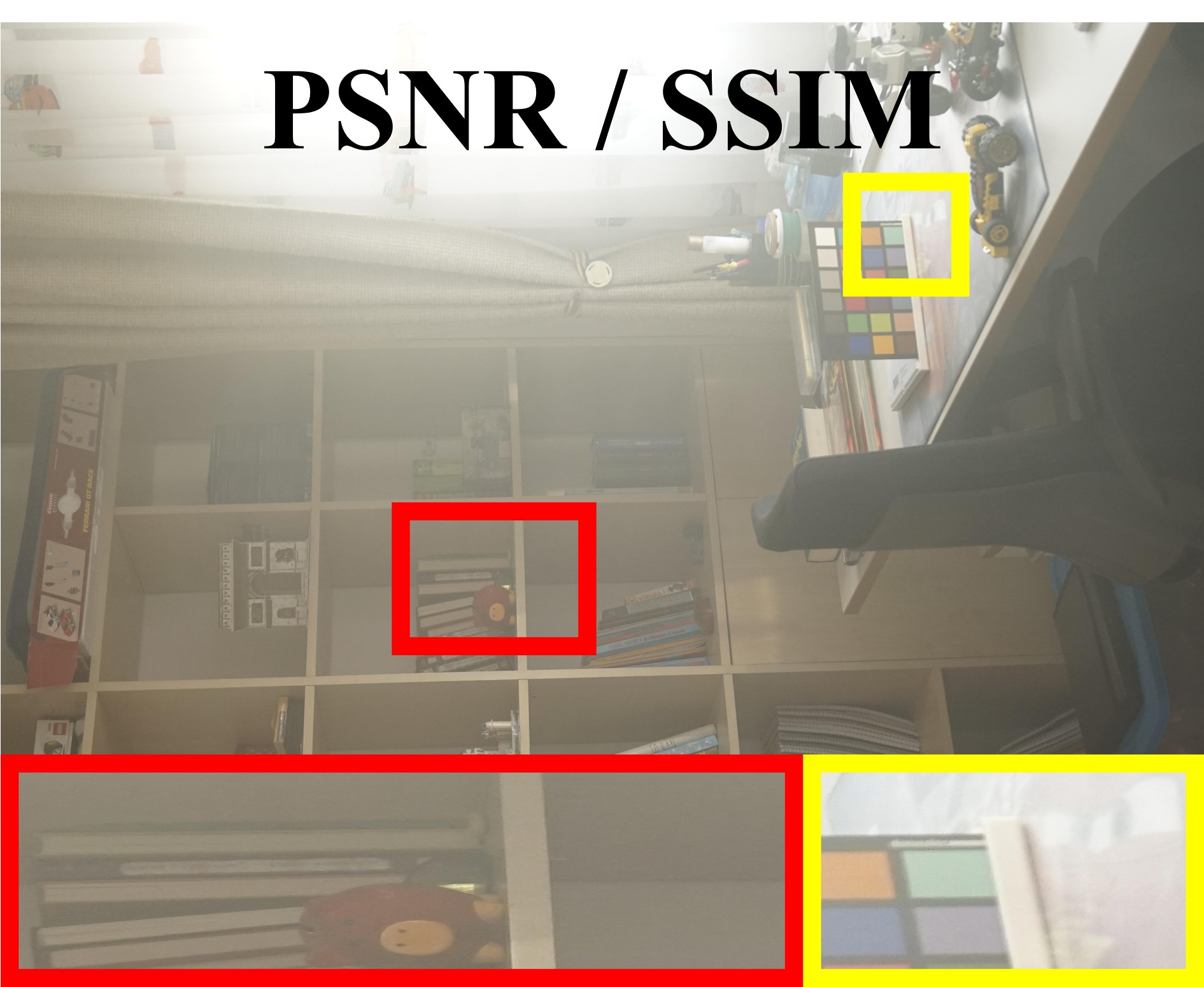}} &
			\adjustbox{valign=m}{\includegraphics[width=\imgw]{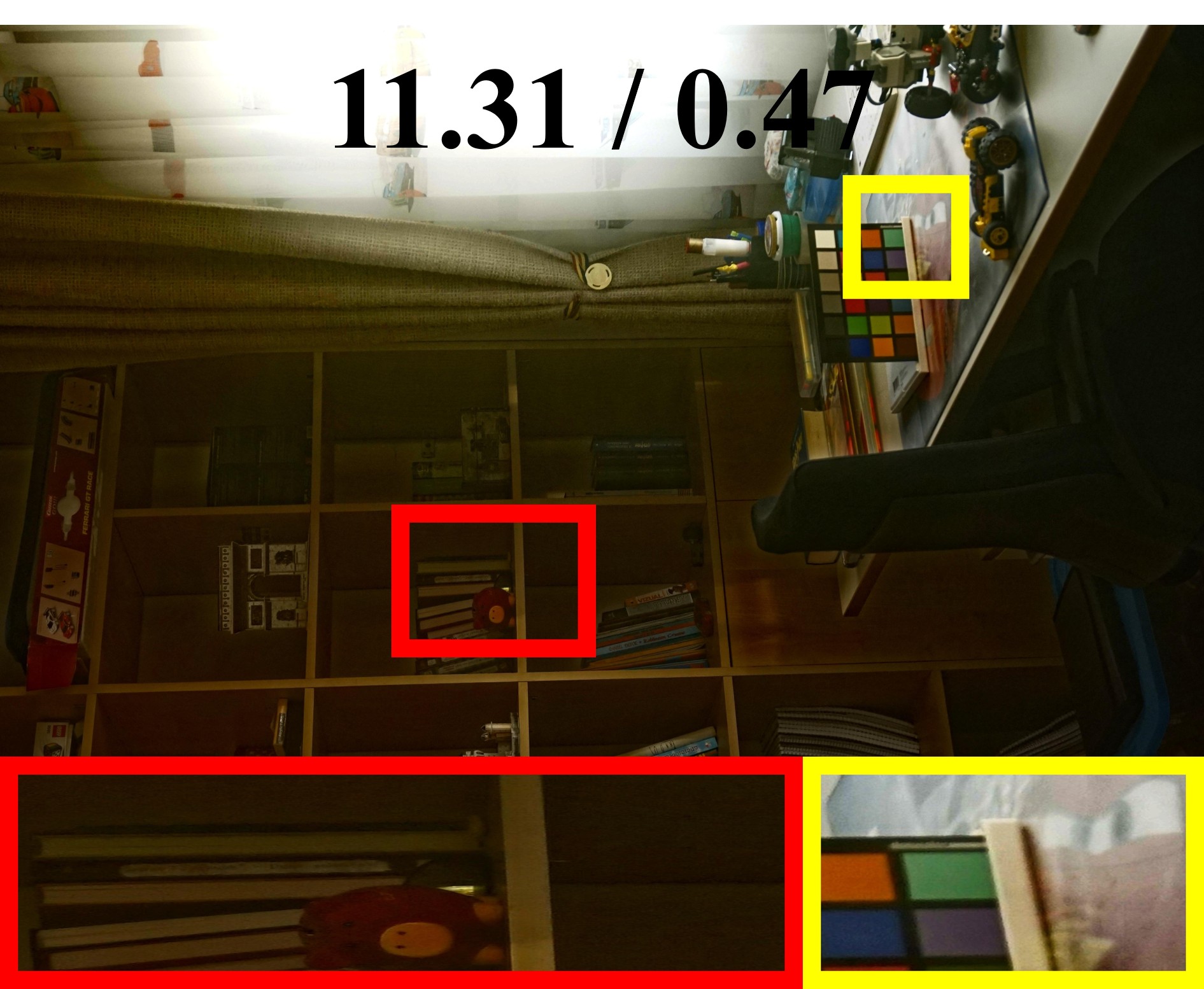}} &
			\adjustbox{valign=m}{\includegraphics[width=\imgw]{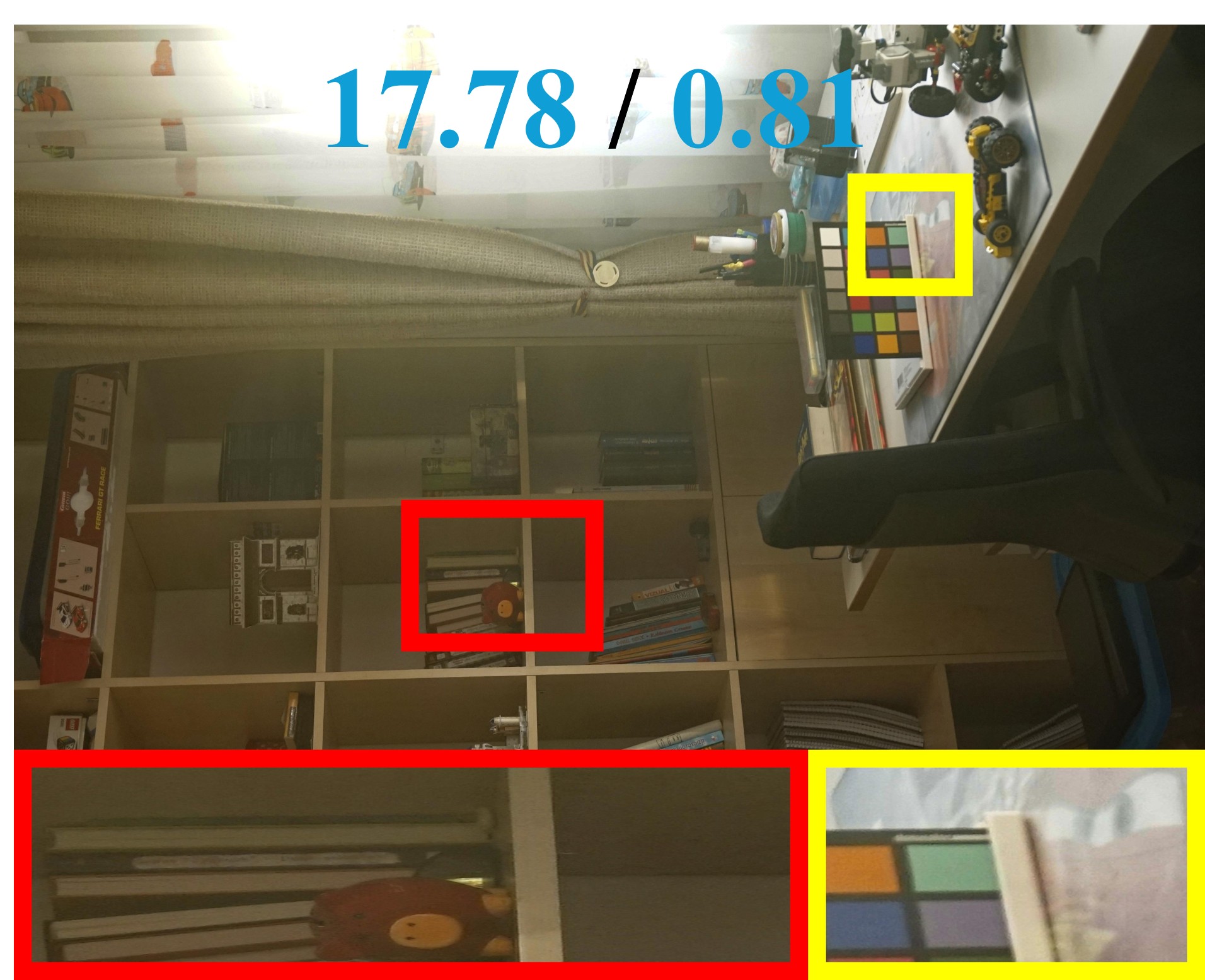}} &
			\adjustbox{valign=m}{\includegraphics[width=\imgw]{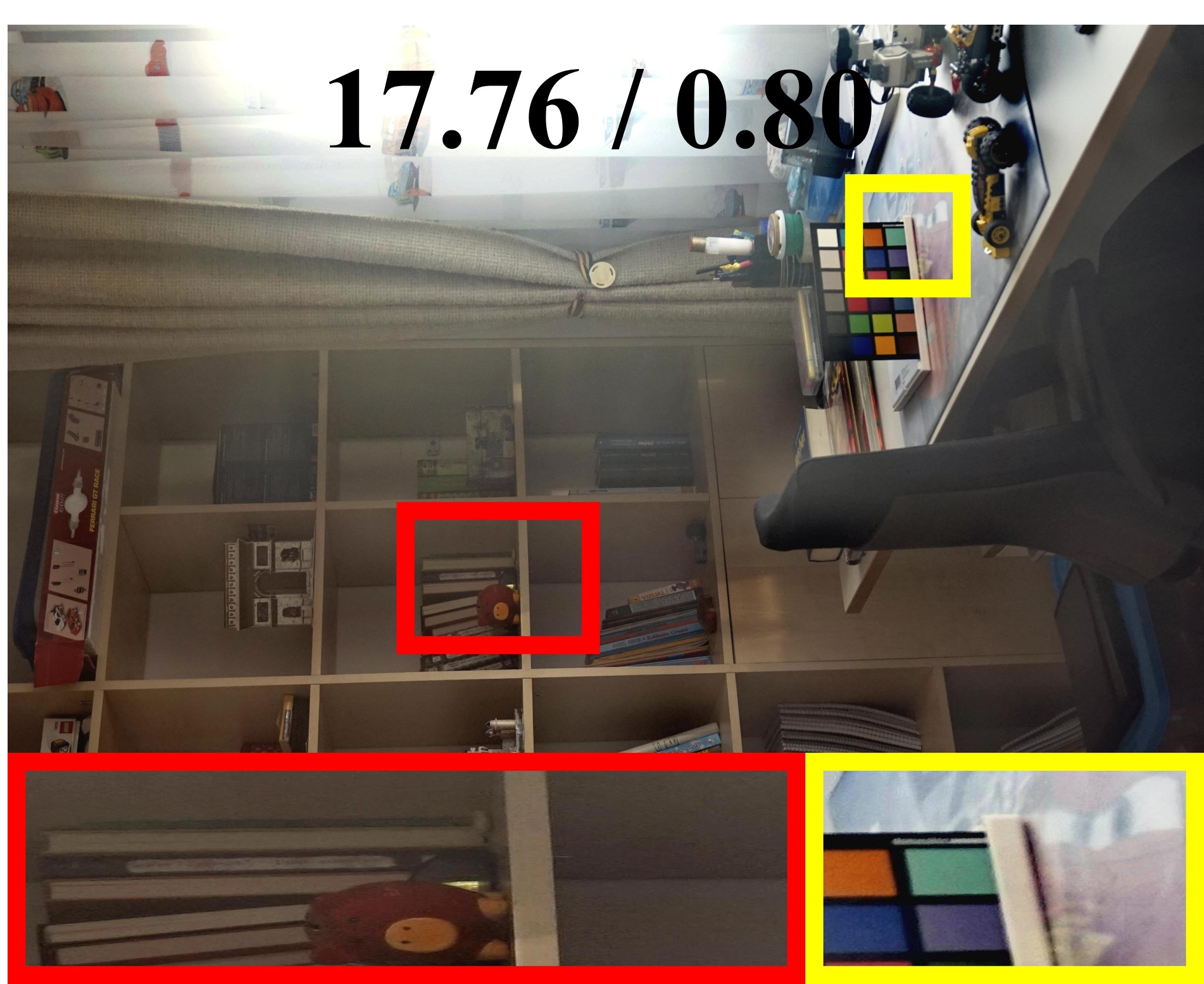}} &
			\adjustbox{valign=m}{\includegraphics[width=\imgw]{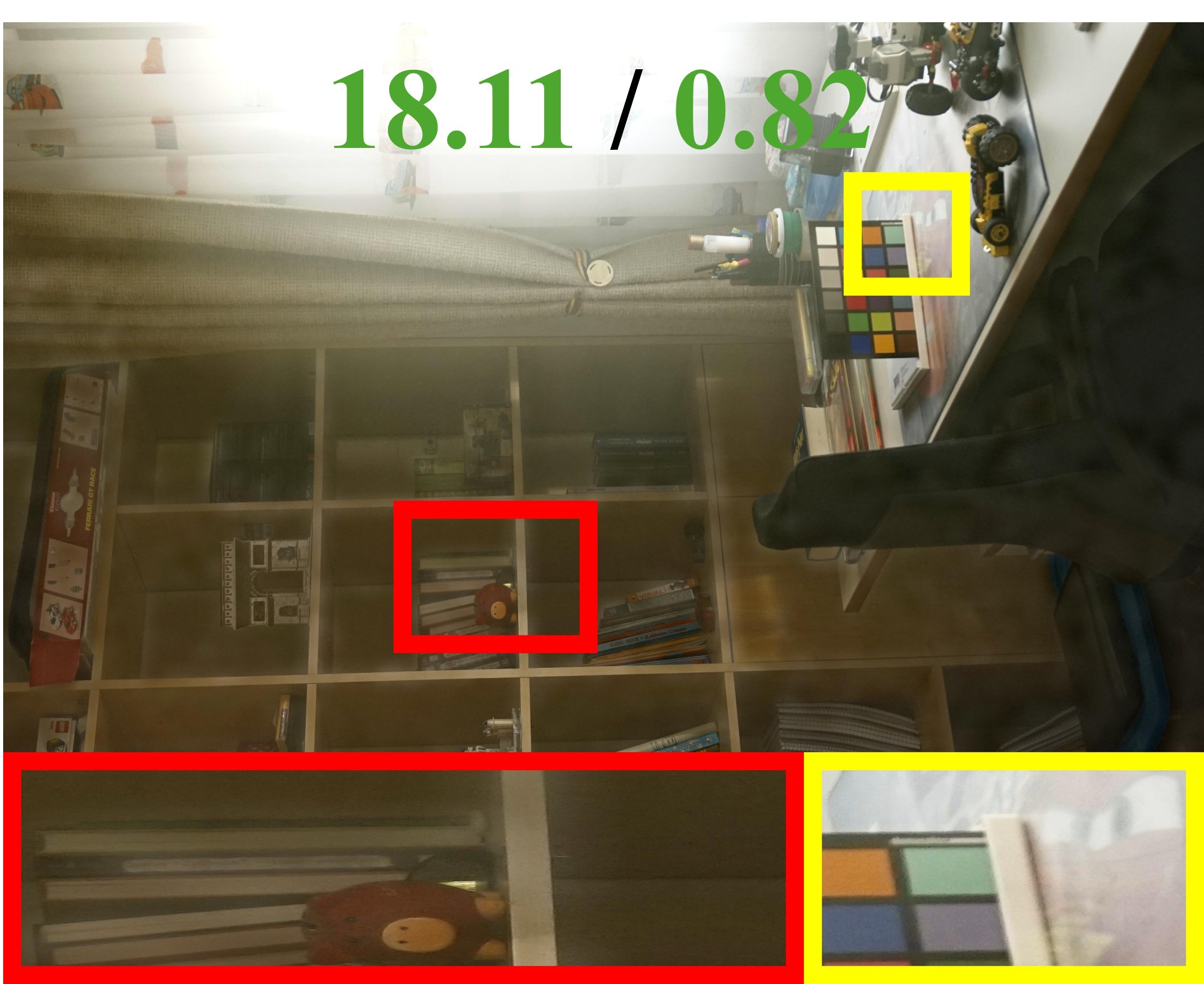}} &
			\adjustbox{valign=m}{\includegraphics[width=\imgw]{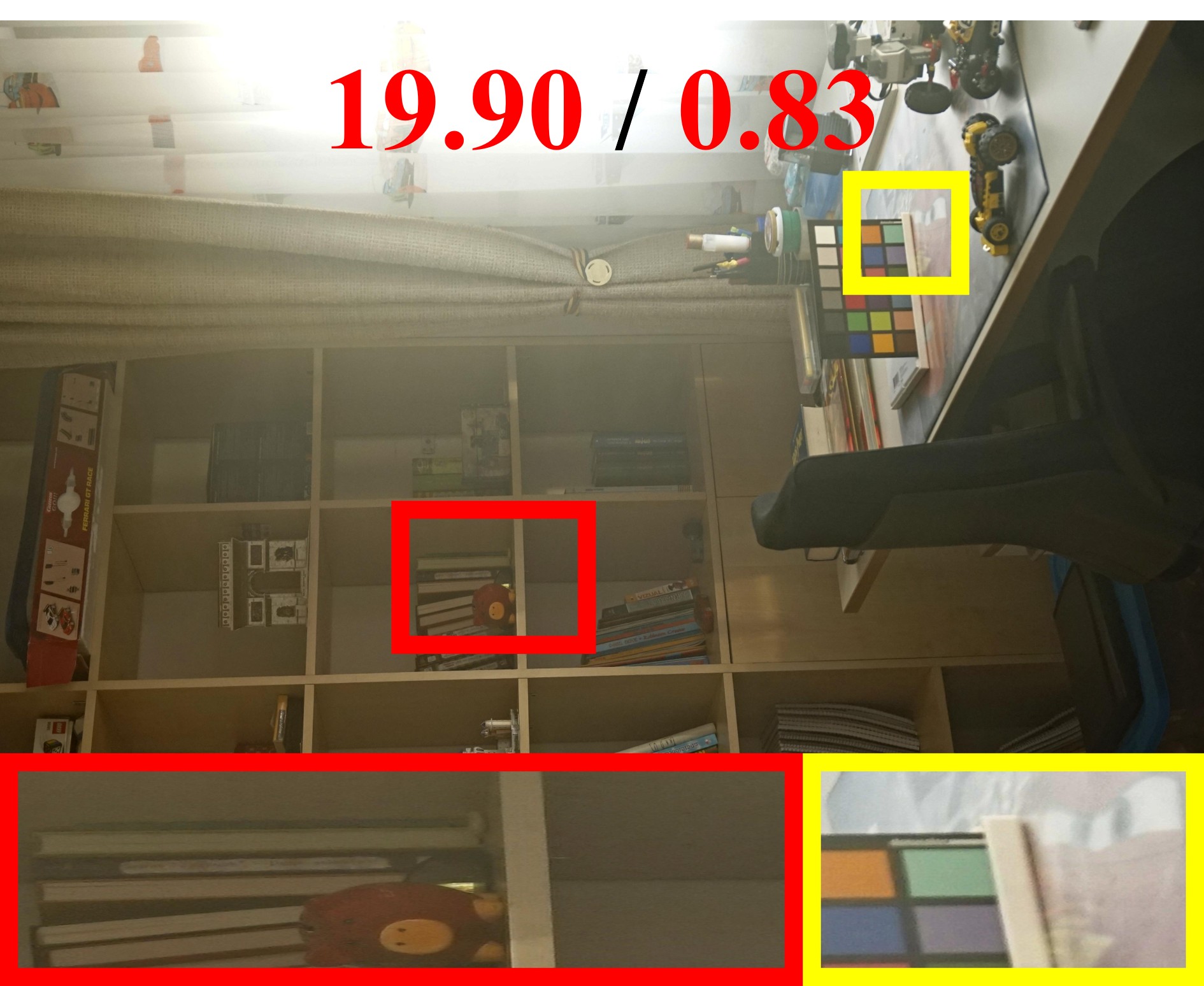}} &
			\adjustbox{valign=m}{\includegraphics[width=\imgw]{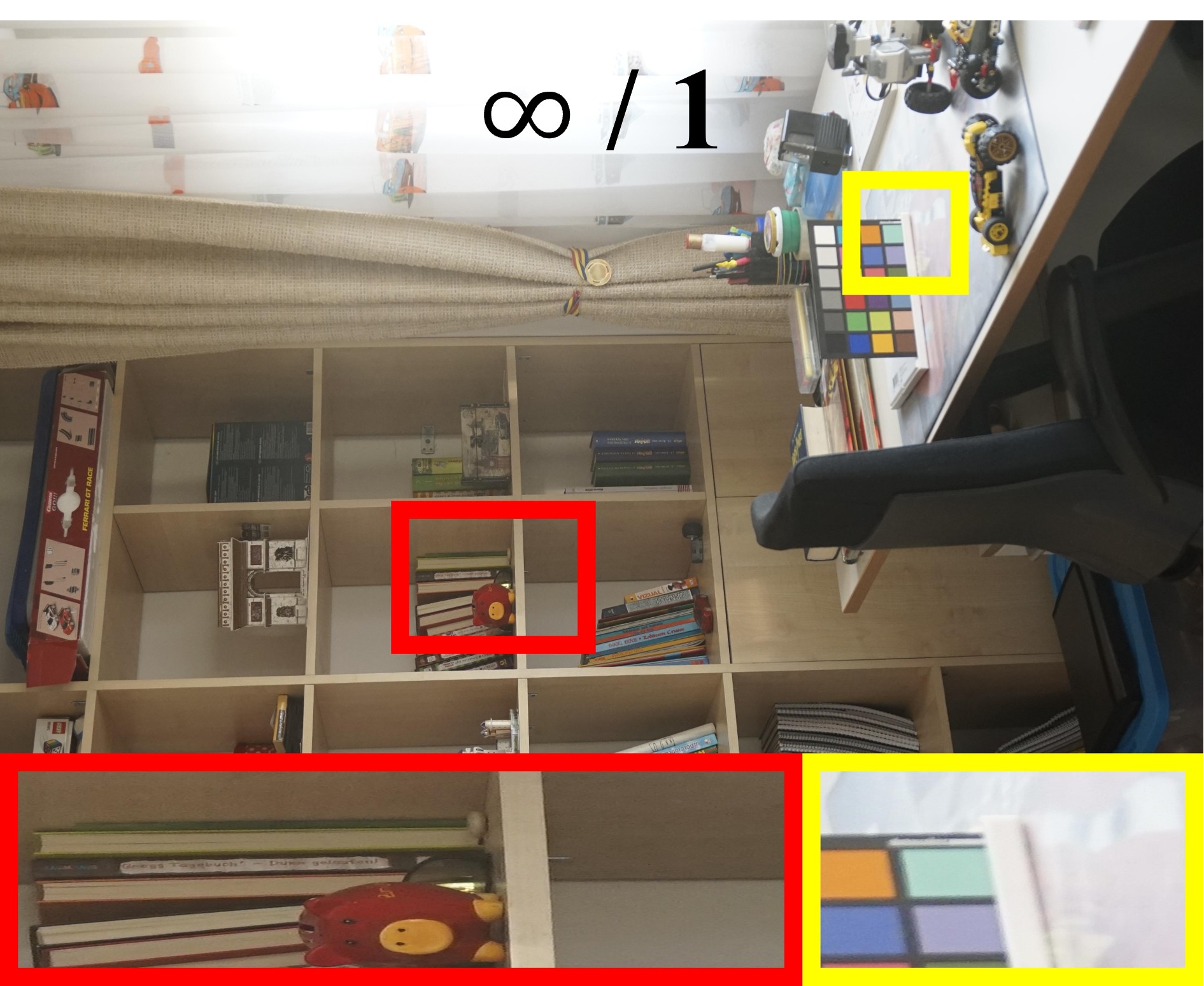}} \\
			
			\rlabel{O-HAZE} &
			\adjustbox{valign=m}{\includegraphics[width=\imgw]{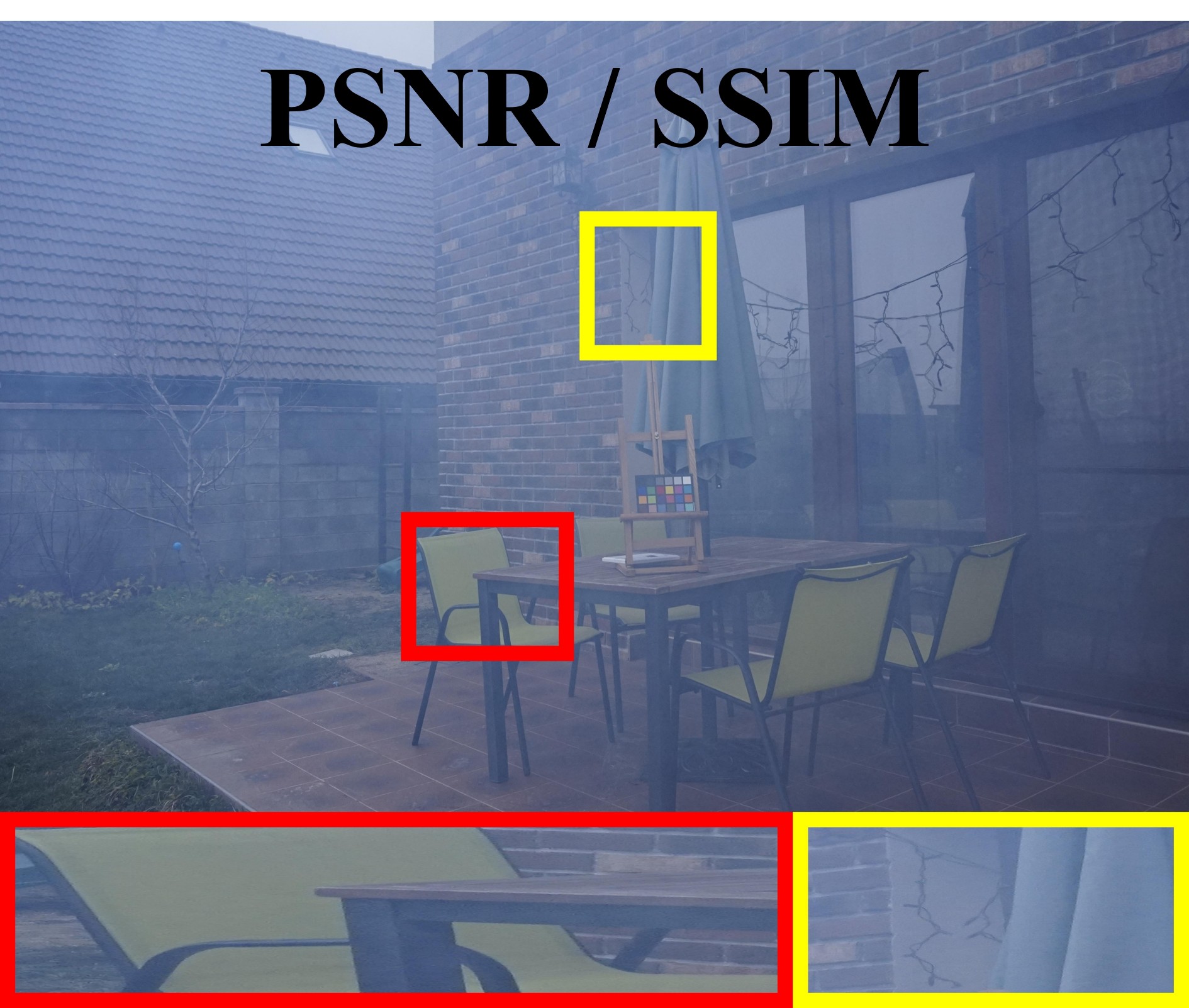}} &
			\adjustbox{valign=m}{\includegraphics[width=\imgw]{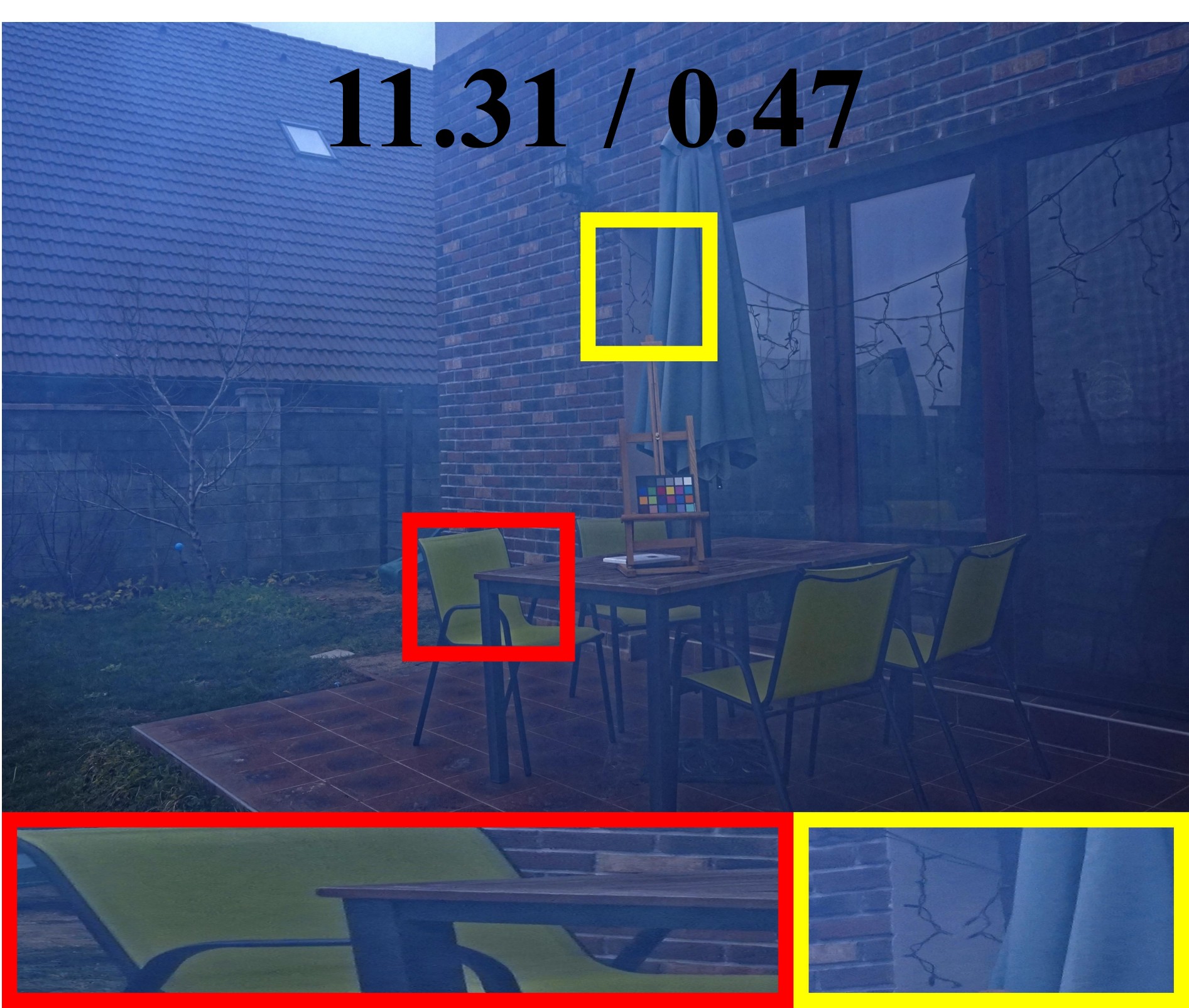}} &
			\adjustbox{valign=m}{\includegraphics[width=\imgw]{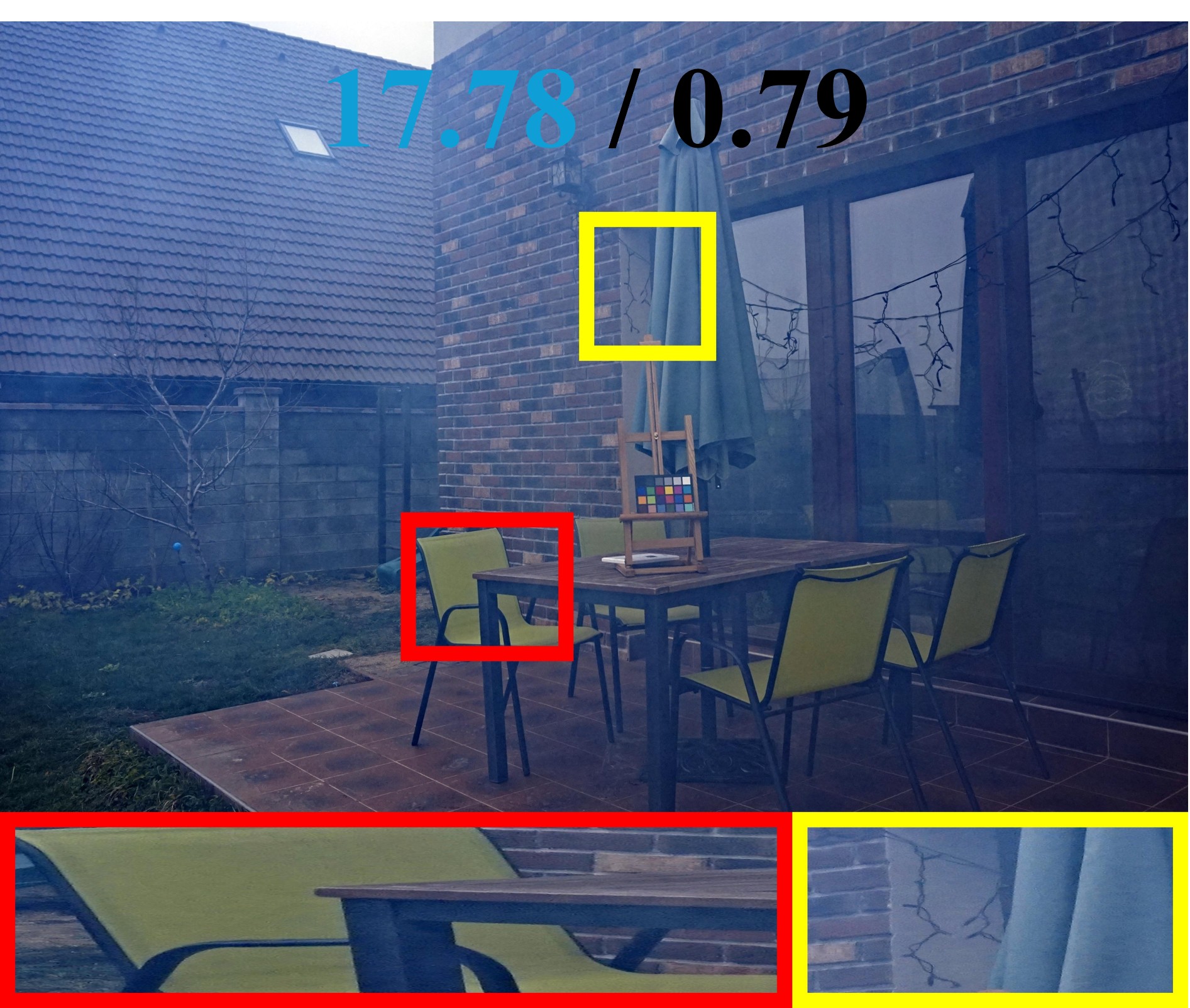}} &
			\adjustbox{valign=m}{\includegraphics[width=\imgw]{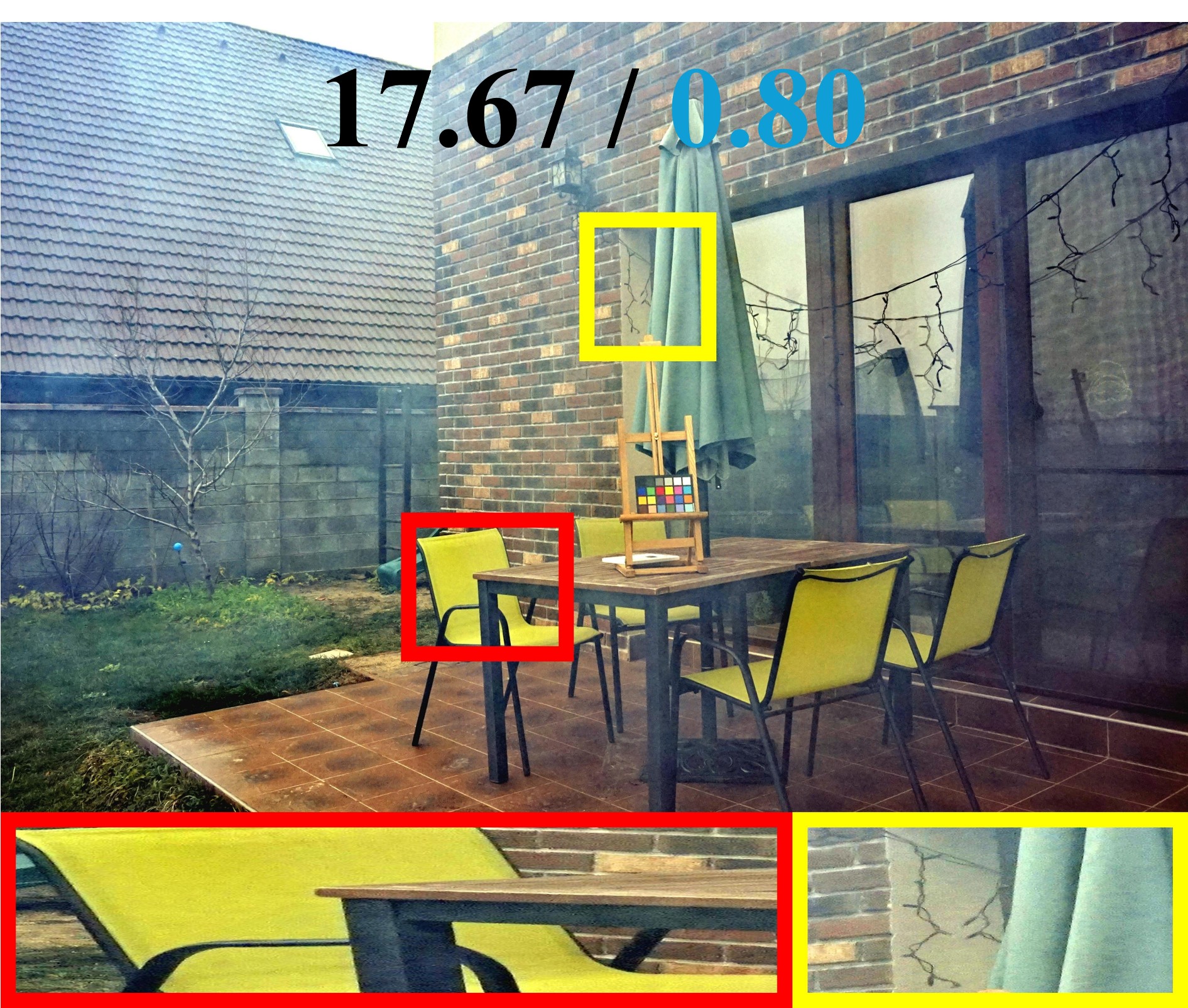}} &
			\adjustbox{valign=m}{\includegraphics[width=\imgw]{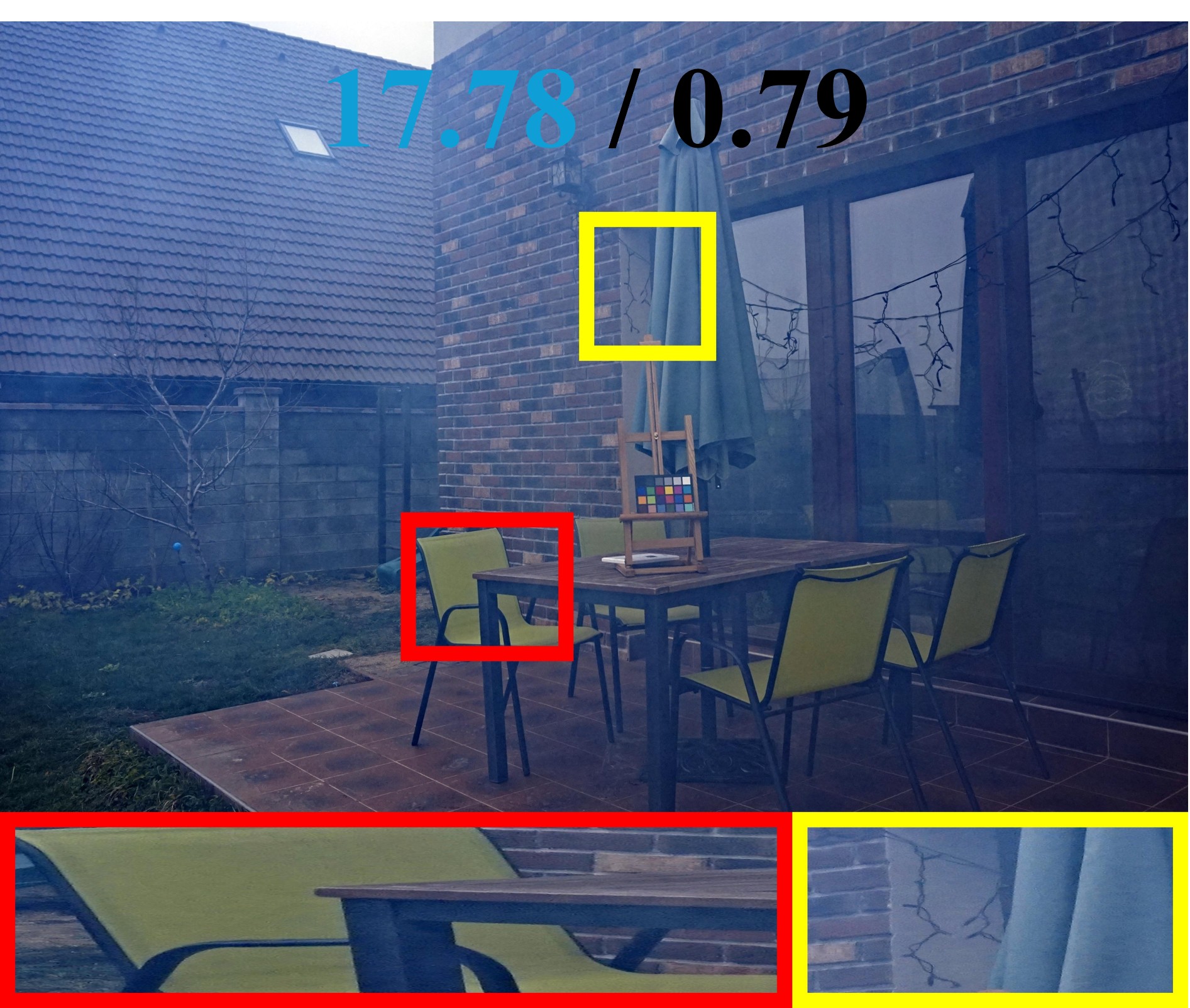}} &
			\adjustbox{valign=m}{\includegraphics[width=\imgw]{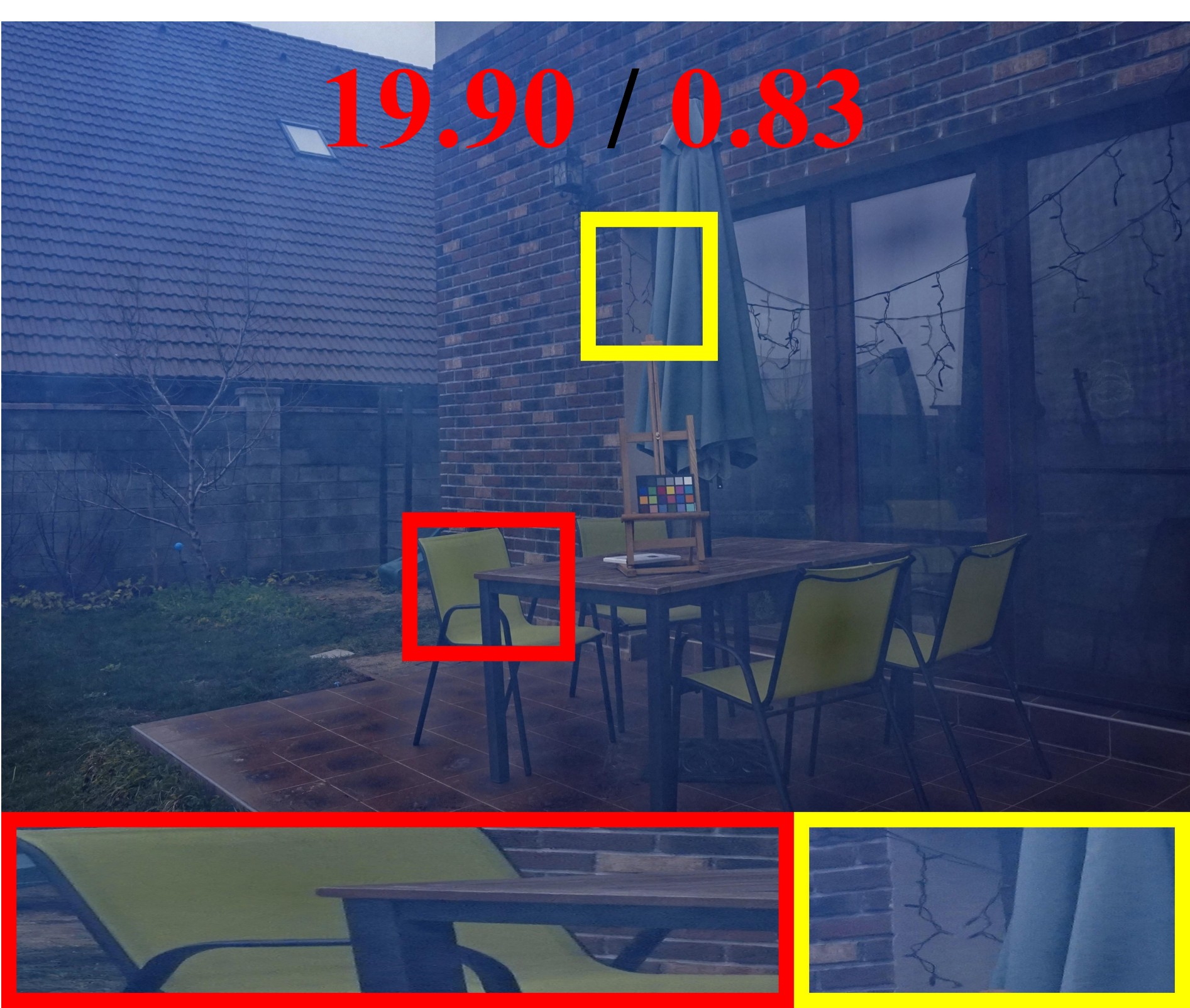}} &
			\adjustbox{valign=m}{\includegraphics[width=\imgw]{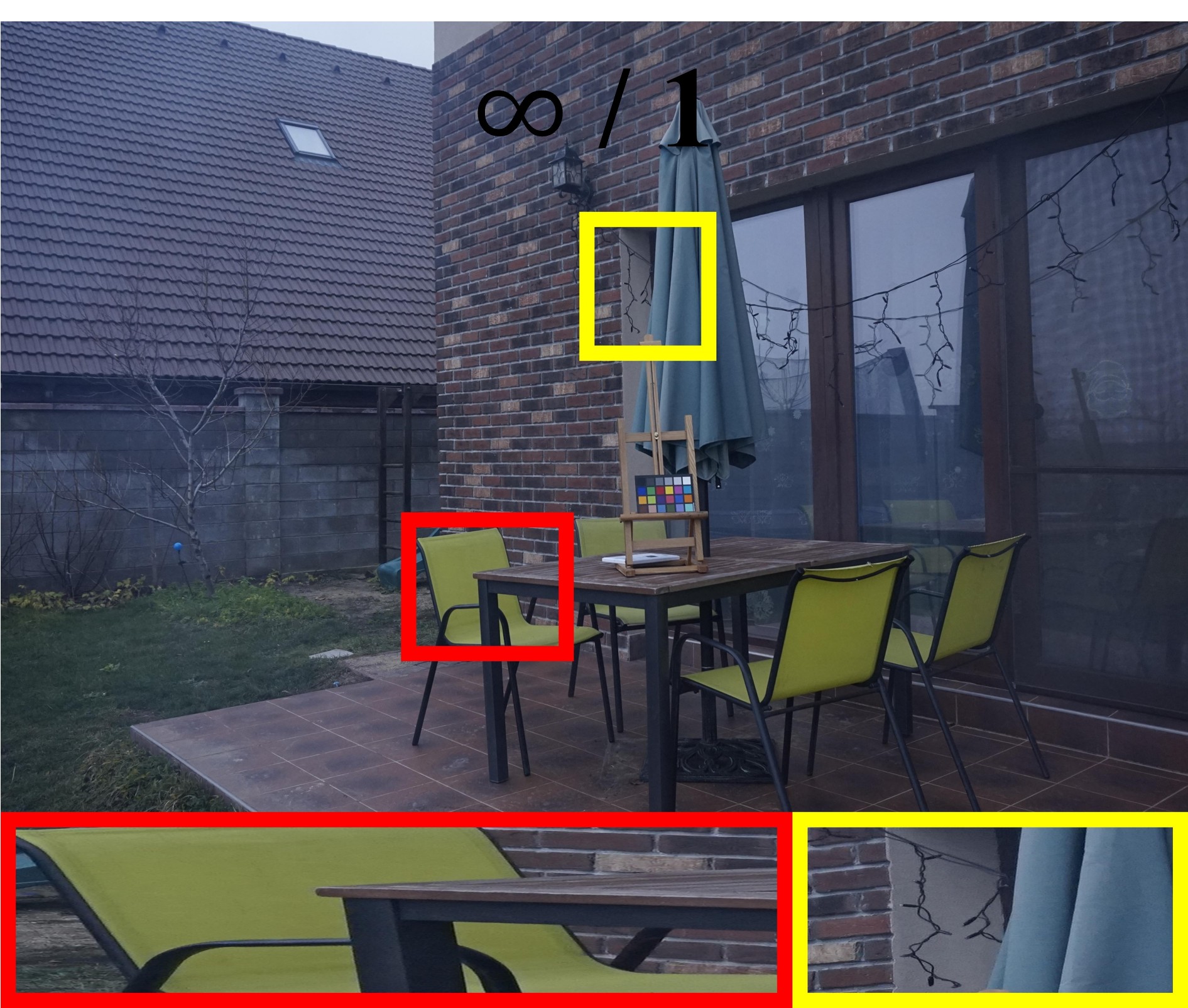}} \\
			
			\rlabel{D-HAZY} &
			\adjustbox{valign=m}{\includegraphics[width=\imgw]{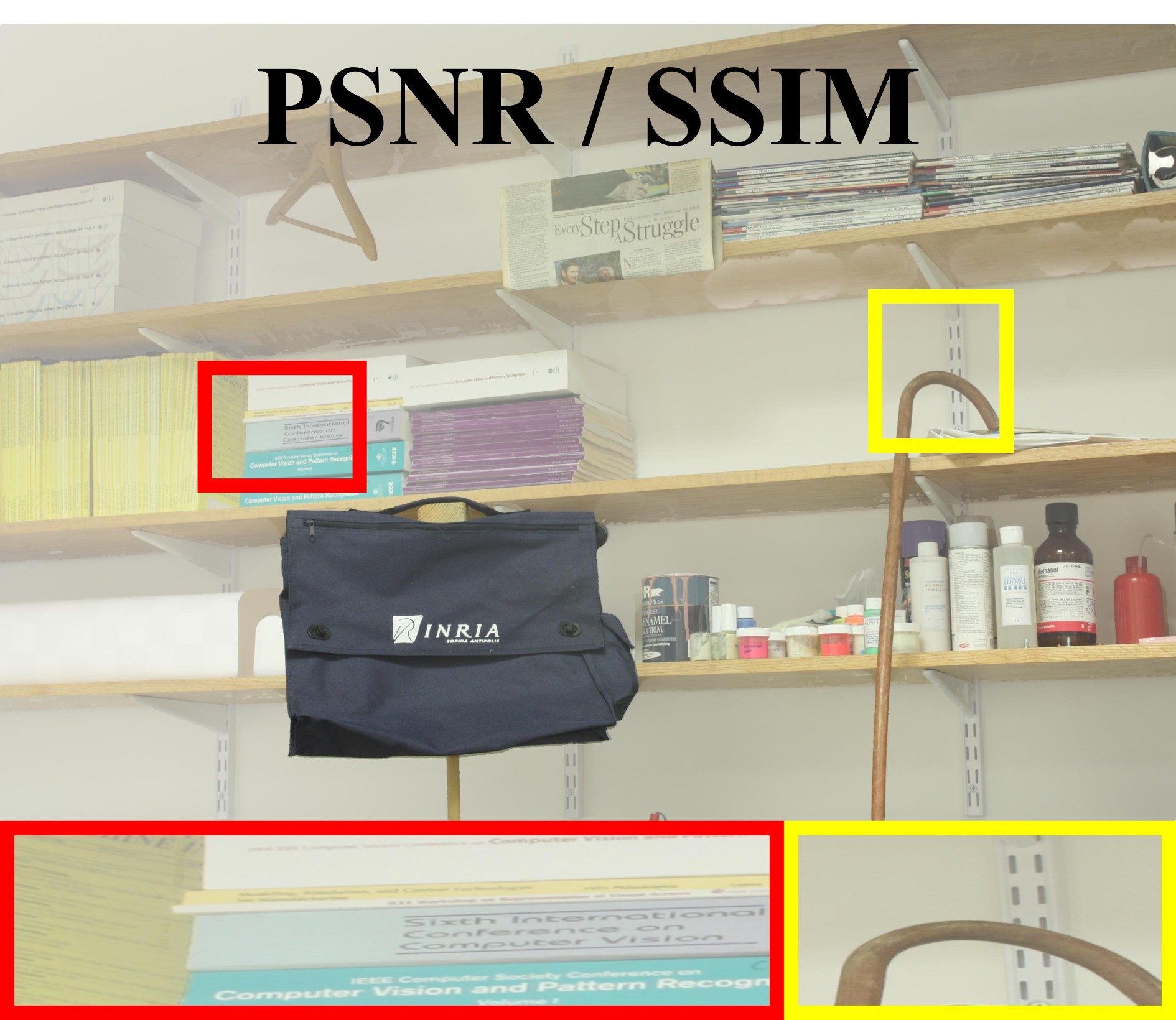}} &
			\adjustbox{valign=m}{\includegraphics[width=\imgw]{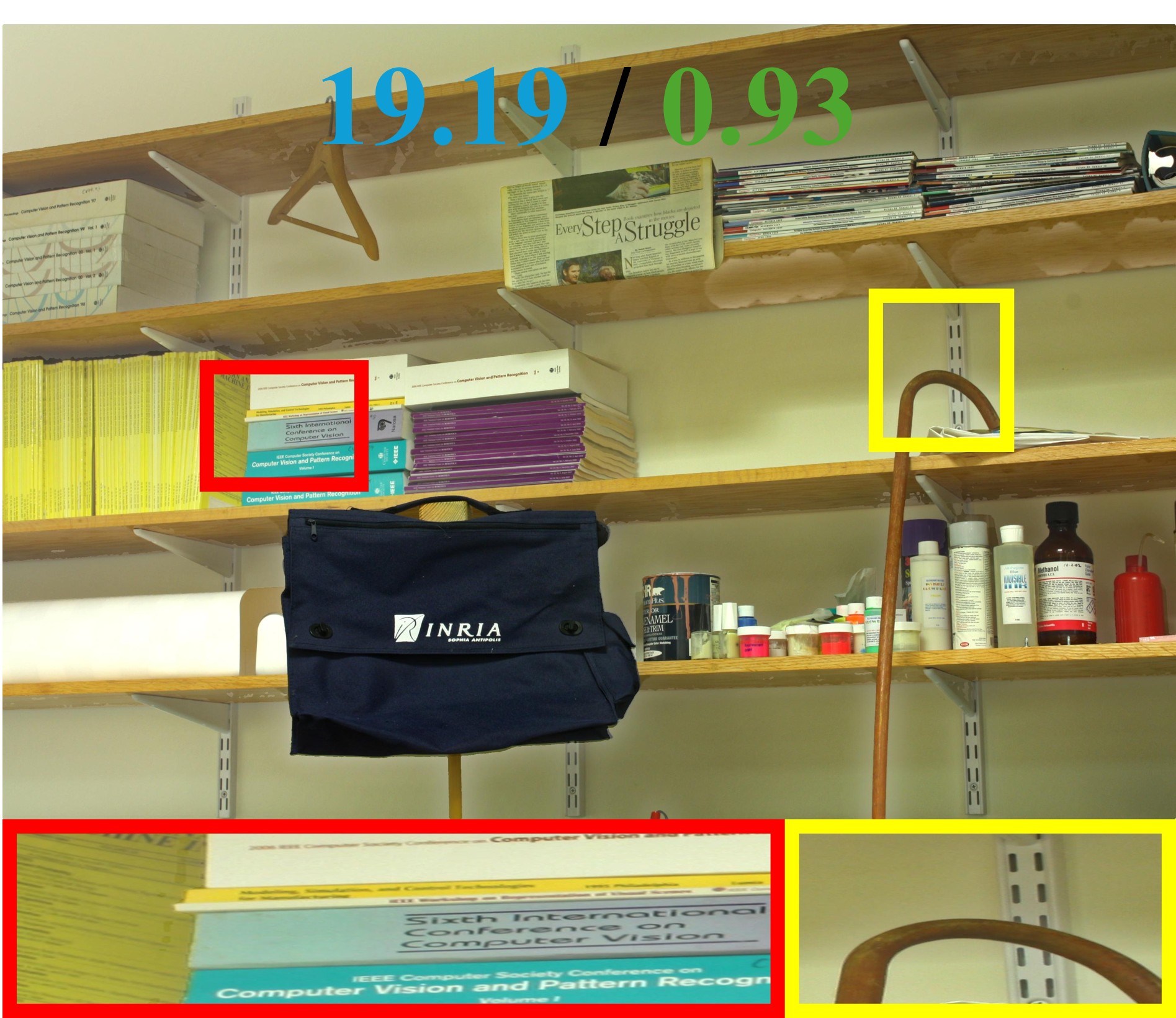}} &
			\adjustbox{valign=m}{\includegraphics[width=\imgw]{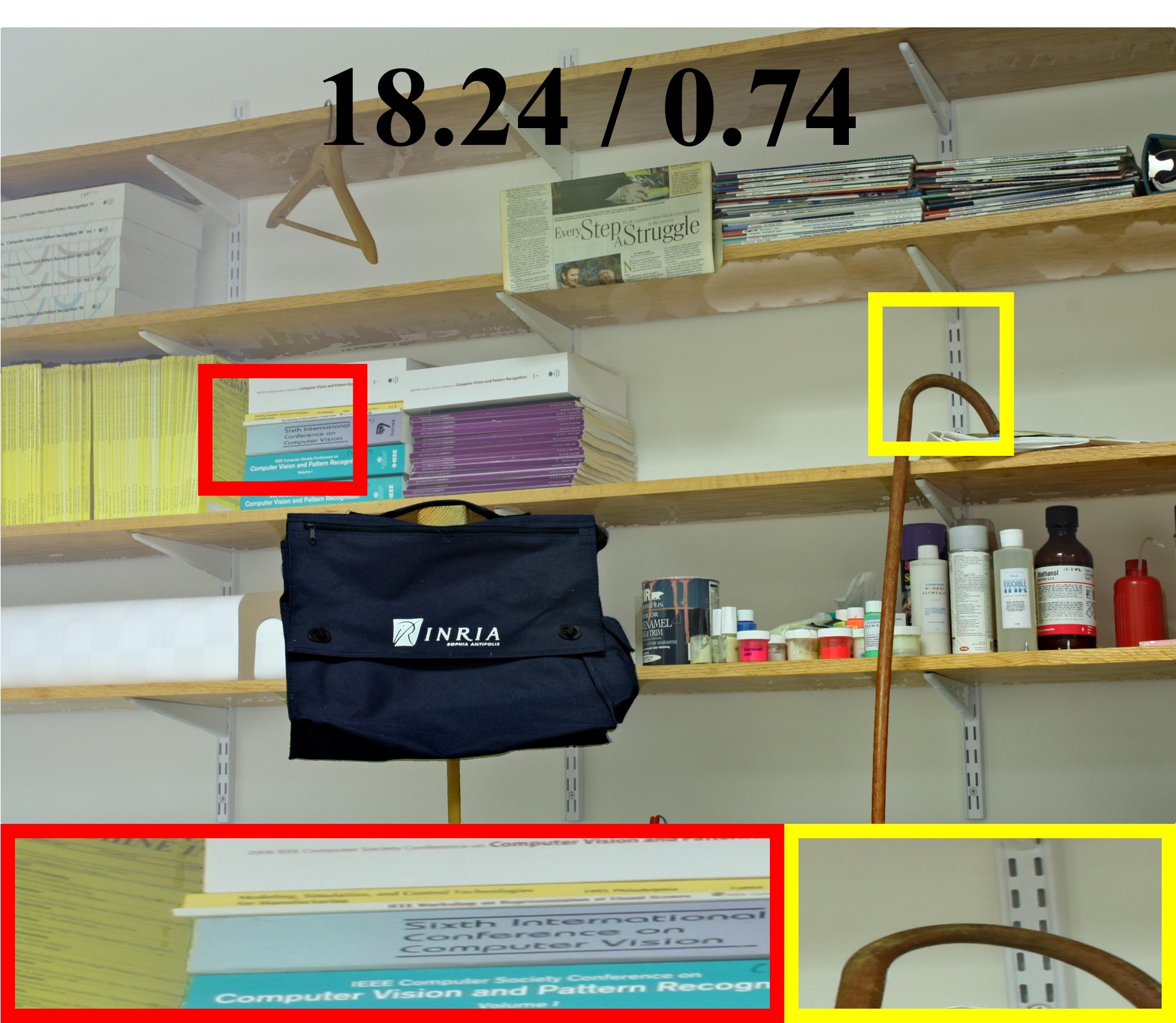}} &
			\adjustbox{valign=m}{\includegraphics[width=\imgw]{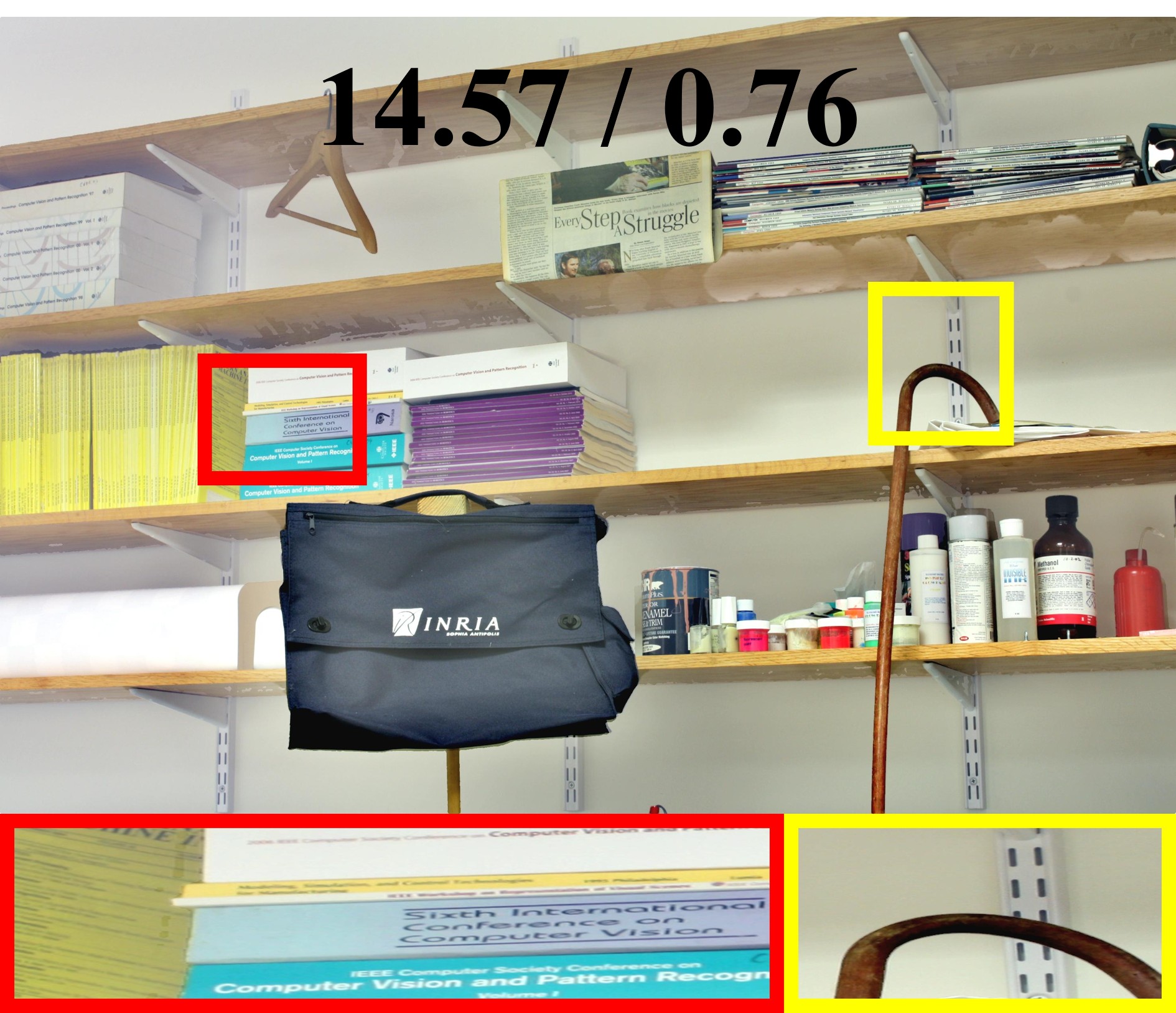}} &
			\adjustbox{valign=m}{\includegraphics[width=\imgw]{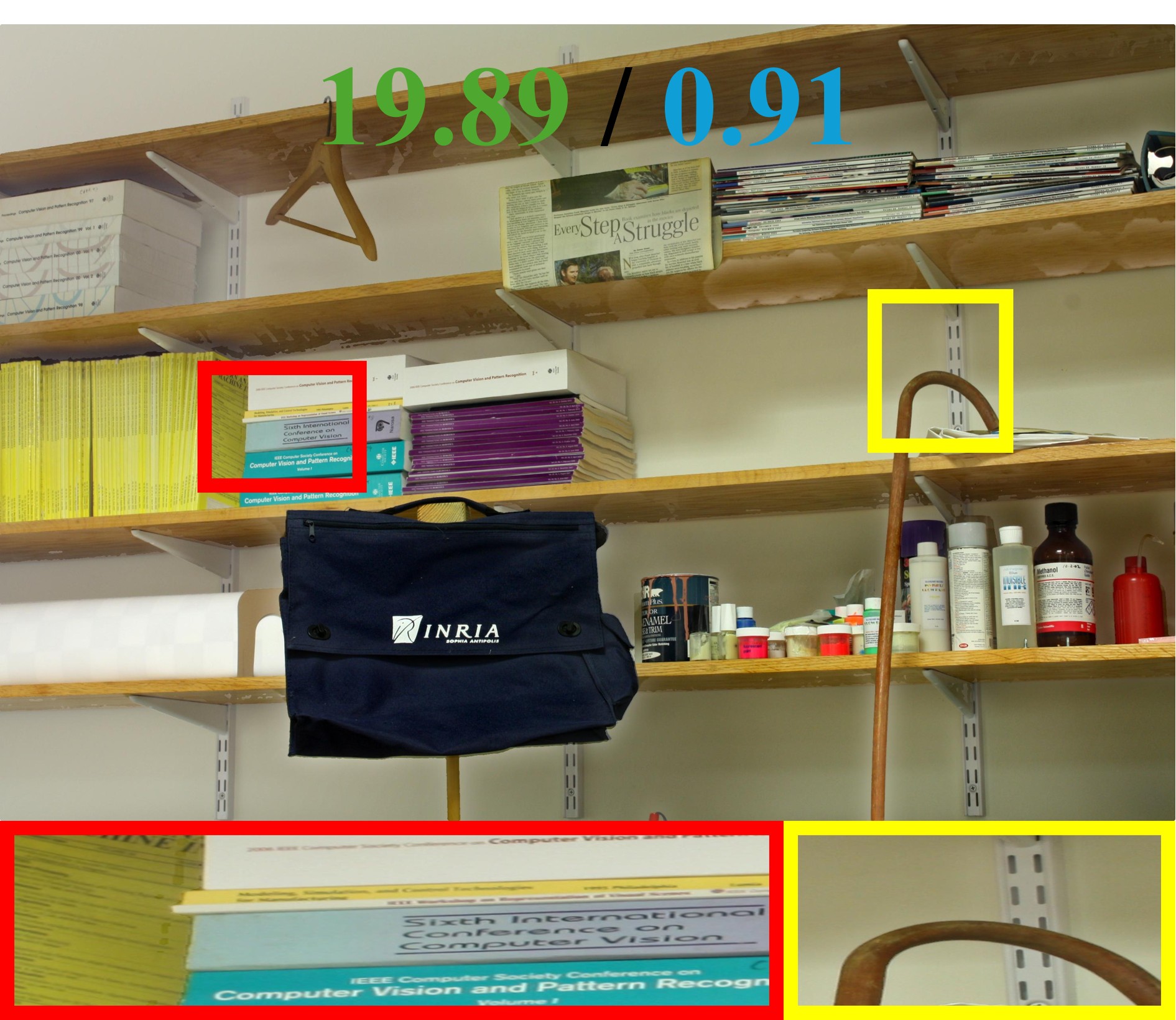}} &
			\adjustbox{valign=m}{\includegraphics[width=\imgw]{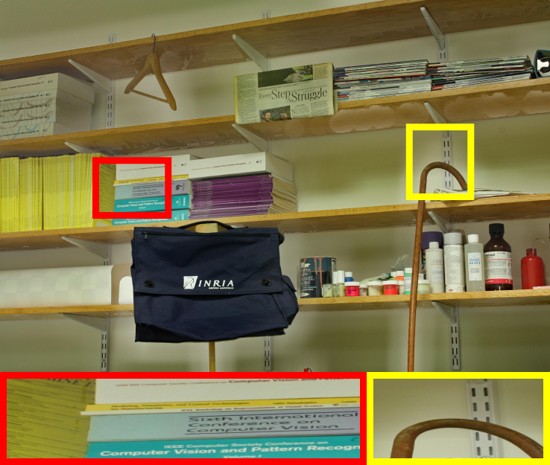}} &
			\adjustbox{valign=m}{\includegraphics[width=\imgw]{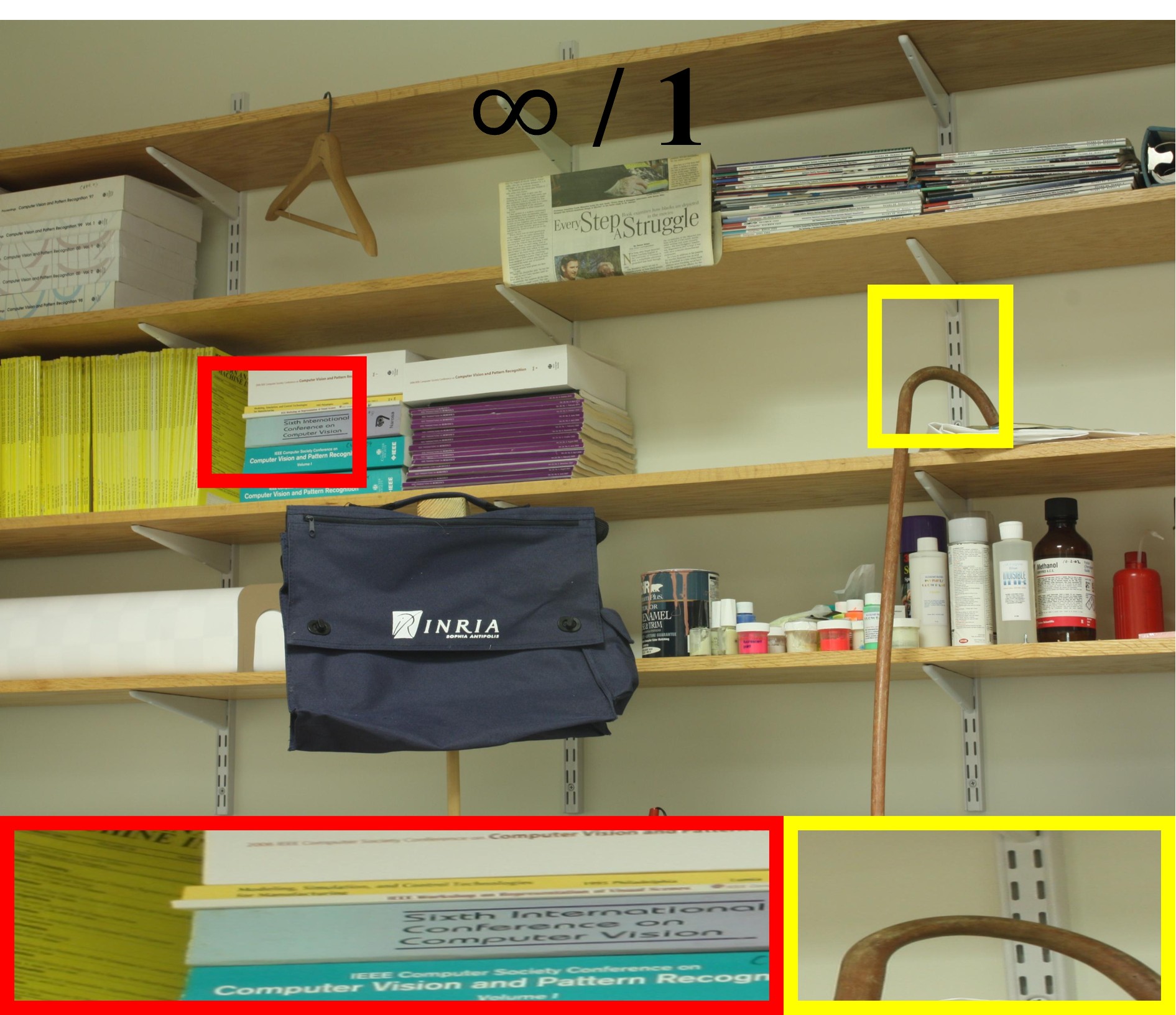}}
		\end{tabular}
	\end{adjustbox}
	\caption{Visual comparison of different prior-based dehazing methods on the SOTS, HSTS, I-HAZE, O-HAZE, and D-HAZY datasets. The proposed ASAP preserves clearer boundaries, restores more natural contrast, and suppresses halo-like artifacts across diverse scene types.}
	\label{fig:asap_visual}
\end{figure*}

Fig.~\ref{fig:asap_visual} presents representative visual comparisons. Compared with existing prior-based methods, ASAP better preserves structural boundaries, suppresses halo-like artifacts, and restores more natural contrast. These qualitative observations are consistent with the quantitative results in Table~\ref{tab:msadcp_sots_dhazy}.

\subsection{Performance of PP-Net$_{\mathrm{P}}$ on RESIDE-ITS and RESIDE-6K}
\label{subsec:hybrid_pipeline_eval}

We next evaluate PP-Net$_{\mathrm{P}}$ on RESIDE-ITS and RESIDE-6K to verify whether ASAP and GF-Net can effectively cooperate under paired synthetic supervision.

\begin{figure}[htbp]
	\centering
	\includegraphics[width=\columnwidth]{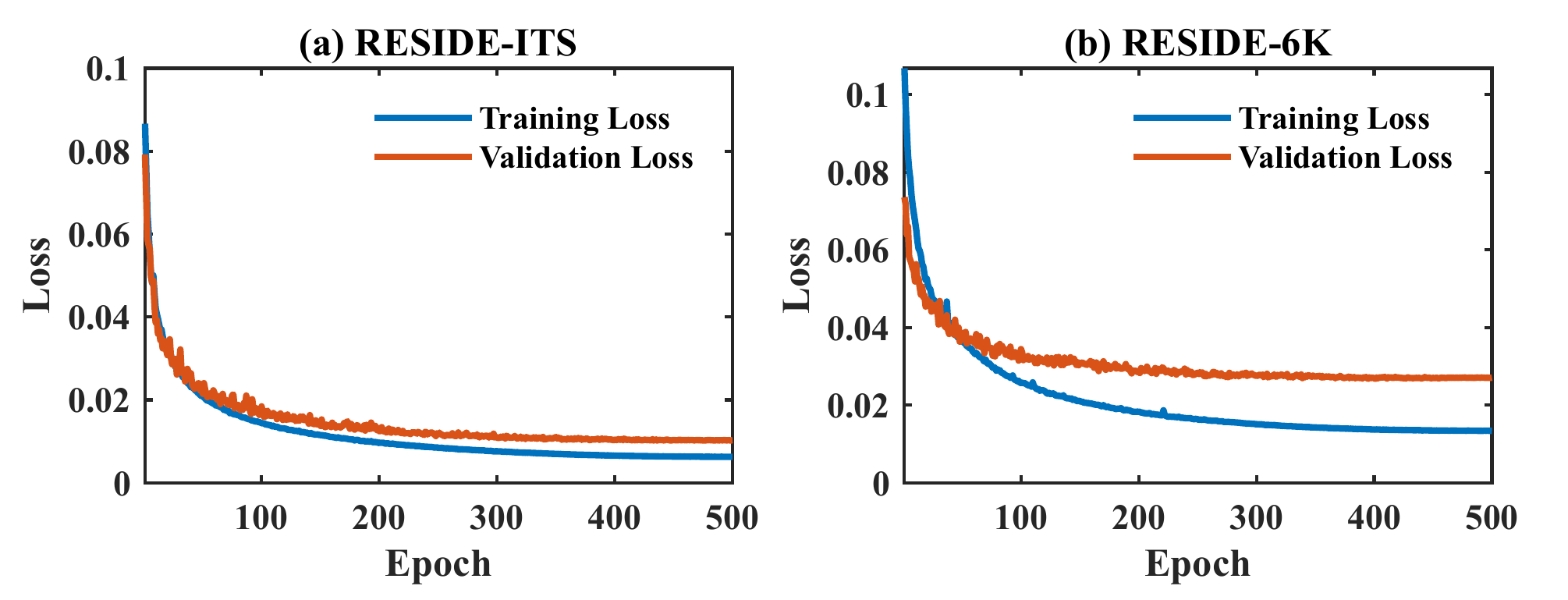}
	\caption{Training and validation loss curves of PP-Net$_{\mathrm{P}}$ on RESIDE-ITS and RESIDE-6K under paired synthetic supervision. Both datasets are trained for 500 epochs using the Charbonnier loss.}
	\label{fig:ppnetp_loss}
\end{figure}

To further examine the optimization process, Fig.~\ref{fig:ppnetp_loss} shows the training and validation loss curves of PP-Net$_{\mathrm{P}}$ on RESIDE-ITS and RESIDE-6K over 500 epochs. The losses decrease rapidly in the early stage and then converge smoothly. The validation curves follow the training curves with a similar decreasing trend, indicating stable optimization and generalization during training.

\begin{table}[htbp]
	\centering
	\caption{Quantitative comparison between PP-Net$_{\mathrm{P}}$ and representative learning-based dehazing methods on ITS and RESIDE-6K under \textbf{the haze-only setting}. Higher PSNR/SSIM indicates better quality. {\color{red}\textbf{Red}}, {\color{darkgreen}\textbf{green}}, and {\color{blue}\textbf{blue}} indicate the {\color{red}\textbf{best}}, {\color{darkgreen}\textbf{second-best}}, and {\color{blue}\textbf{third-best}} results.}
	
	\label{tab:ppnetp_its_reside6k}
	\scriptsize
	\setlength{\tabcolsep}{3.5pt}
	\renewcommand{\arraystretch}{1.10}
	\resizebox{\columnwidth}{!}{%
		\begin{tabular}{c|cc|cc}
			\hline
			\hline
			\multicolumn{5}{c}{\textbf{Noise-Free Haze Case}} \\
			\hline
			\multirow{2}{*}{Method}
			& \multicolumn{2}{c|}{ITS \cite{RESIDE}}
			& \multicolumn{2}{c}{RESIDE-6K \cite{RESIDE}} \\
			\cline{2-5}
			& SSIM$\uparrow$ & PSNR$\uparrow$
			& SSIM$\uparrow$ & PSNR$\uparrow$ \\
			\hline
			(ECCV'16) MSCNN \cite{MSCNN}    & 0.8342 & 19.8443 & 0.8262 & 22.8021 \\
			(ICCV'17) AOD-Net \cite{AOD-Net}  & 0.8164 & 20.5132 & 0.8554 & 20.2754 \\
			(CVPR'18) GFN \cite{GFN}      & 0.8802 & 22.3023 & 0.9053 & 23.5245 \\
			(CVPR'20) MSBDN \cite{MSBDN}     & {\color{darkgreen}\textbf{0.9852}} & {\color{blue}\textbf{33.6725}} & {\color{blue}\textbf{0.9661}} & {\color{blue}\textbf{28.5632}} \\
			(ECCV'20) PFDN \cite{PFDN}      & 0.9761 & 32.6802 & 0.9621 & 28.1546 \\
			(AAAI'21) FFA-Net \cite{FFA-Net}  & {\color{blue}\textbf{0.9765}} & {\color{darkgreen}\textbf{36.3913}} & {\color{darkgreen}\textbf{0.9731}} & {\color{darkgreen}\textbf{29.9623}} \\
			(TETCI'24) TBN \cite{TBN}     & 0.8510 & 18.3860 & 0.8614 & 19.0325 \\
			(TIM'25) CDVA \cite{CDVA}      & 0.7333 & 16.0261 & 0.7805 & 15.7067 \\
			(TITS'25) IDB \cite{IDB}      & 0.6176 & 18.8099 & 0.6090 & 20.9200 \\
			(AAAI'25) MPMF-Net \cite{MPMF} & 0.8421 & 20.2619 & 0.8352 & 23.0786 \\
			\textbf{(Ours) PP-Net$_{\mathrm{P}}$} & {\color{red}\textbf{0.9901}} & {\color{red}\textbf{37.6547}} & {\color{red}\textbf{0.9792}} & {\color{red}\textbf{30.2354}} \\
			\hline
			\hline
		\end{tabular}%
	}
\end{table}

As shown in Table~\ref{tab:ppnetp_its_reside6k}, PP-Net$_{\mathrm{P}}$ achieves the best PSNR and SSIM on both datasets. This result demonstrates that the proposed physical-prior-guided refinement strategy effectively combines scattering-map estimation, prior-map recovery, and learning-based detail reconstruction on paired synthetic data.

\begin{figure*}[htbp]
	\centering
	\setlength{\tabcolsep}{2pt}
	\renewcommand{\arraystretch}{1.0}
	
	\newcommand{\imgw}{0.13\textwidth}
	\newcommand{\rlabel}[1]{\adjustbox{valign=m}{\rotatebox[origin=c]{90}{\scriptsize\textbf{#1}}}}
	\newcommand{\chead}[1]{\makebox[\imgw][c]{\footnotesize\textbf{#1}}}
	\newcommand{\cheadtwo}[2]{\makebox[\imgw][c]{\shortstack{\footnotesize\textbf{#1}\\[-0.4ex]\footnotesize\textbf{#2}}}}
	
	\begin{adjustbox}{max totalsize={\textwidth}{0.95\textheight},center}
		\begin{tabular}{c ccccccc}
			& \chead{Hazy} &
			\chead{AOD-Net\cite{AOD-Net}} &
			\chead{FFA-Net\cite{FFA-Net}} &
			\chead{MPMF-Net\cite{MPMF}} &
			\chead{TBN\cite{TBN}} &
			\chead{PP-Net$_{\mathrm{P}}$ (Ours)} &
			\chead{GT} \\
			
			\rlabel{ITS} &
			\adjustbox{valign=m}{\includegraphics[width=\imgw]{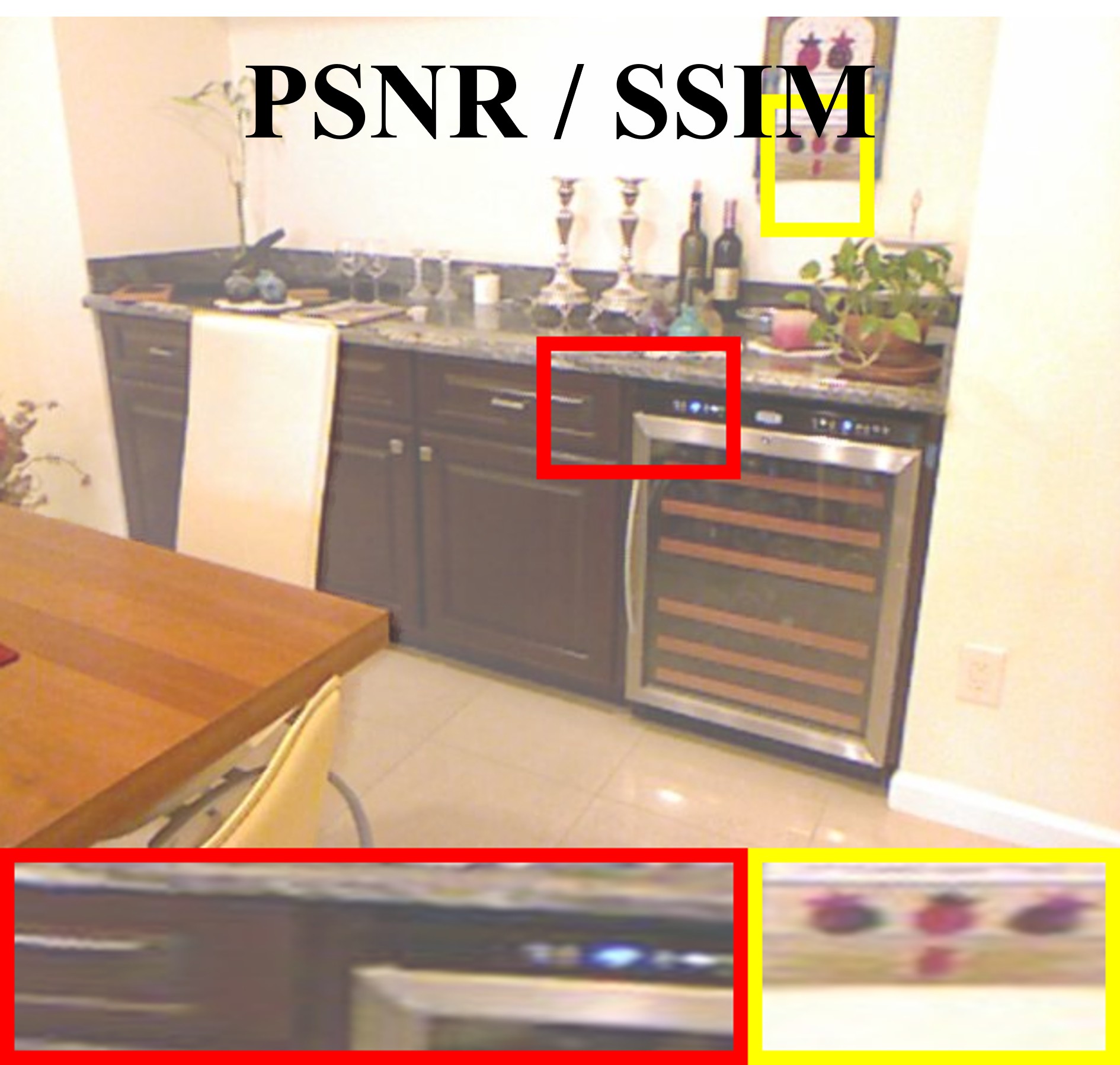}} &
			\adjustbox{valign=m}{\includegraphics[width=\imgw]{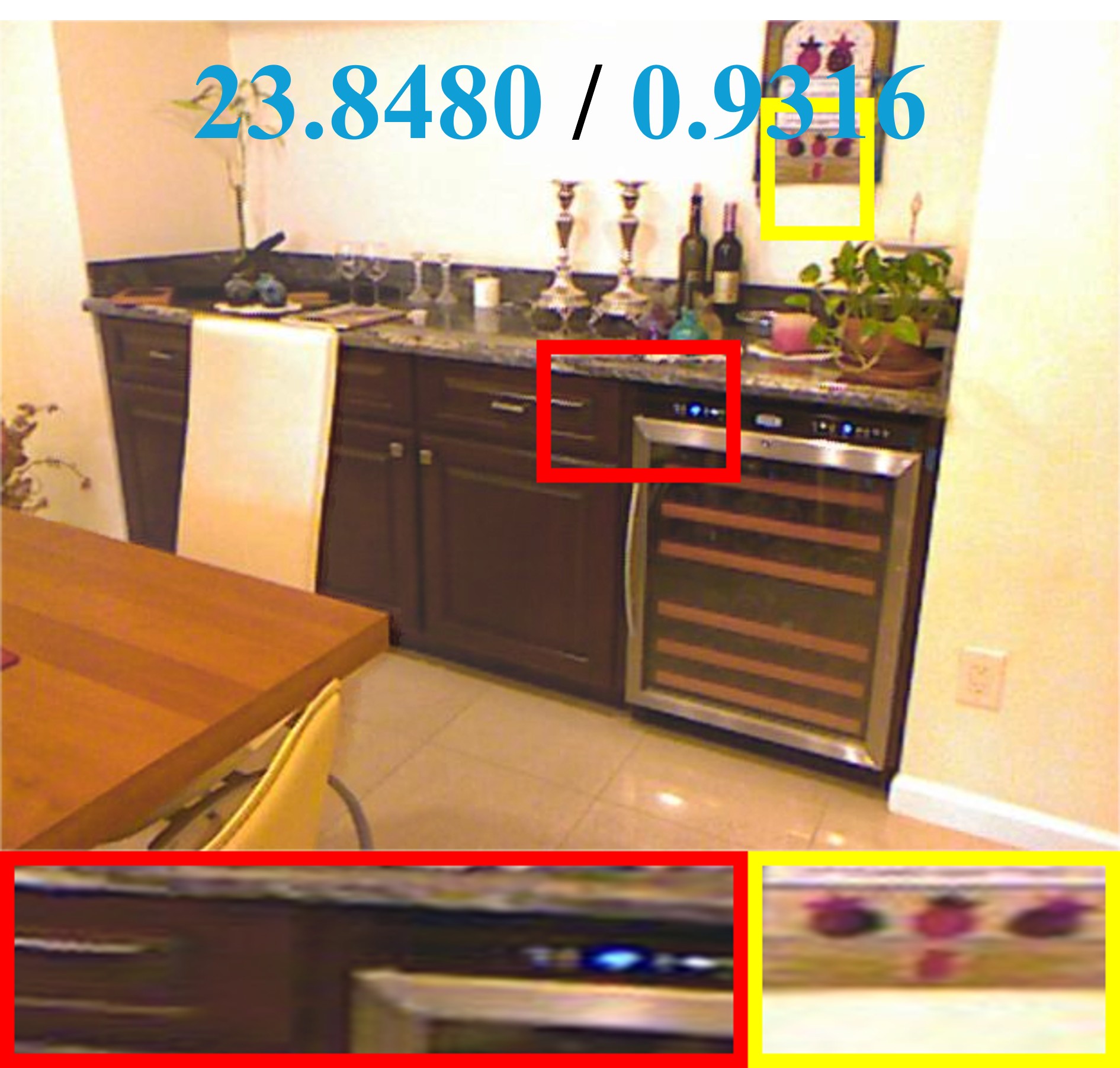}} &
			\adjustbox{valign=m}{\includegraphics[width=\imgw]{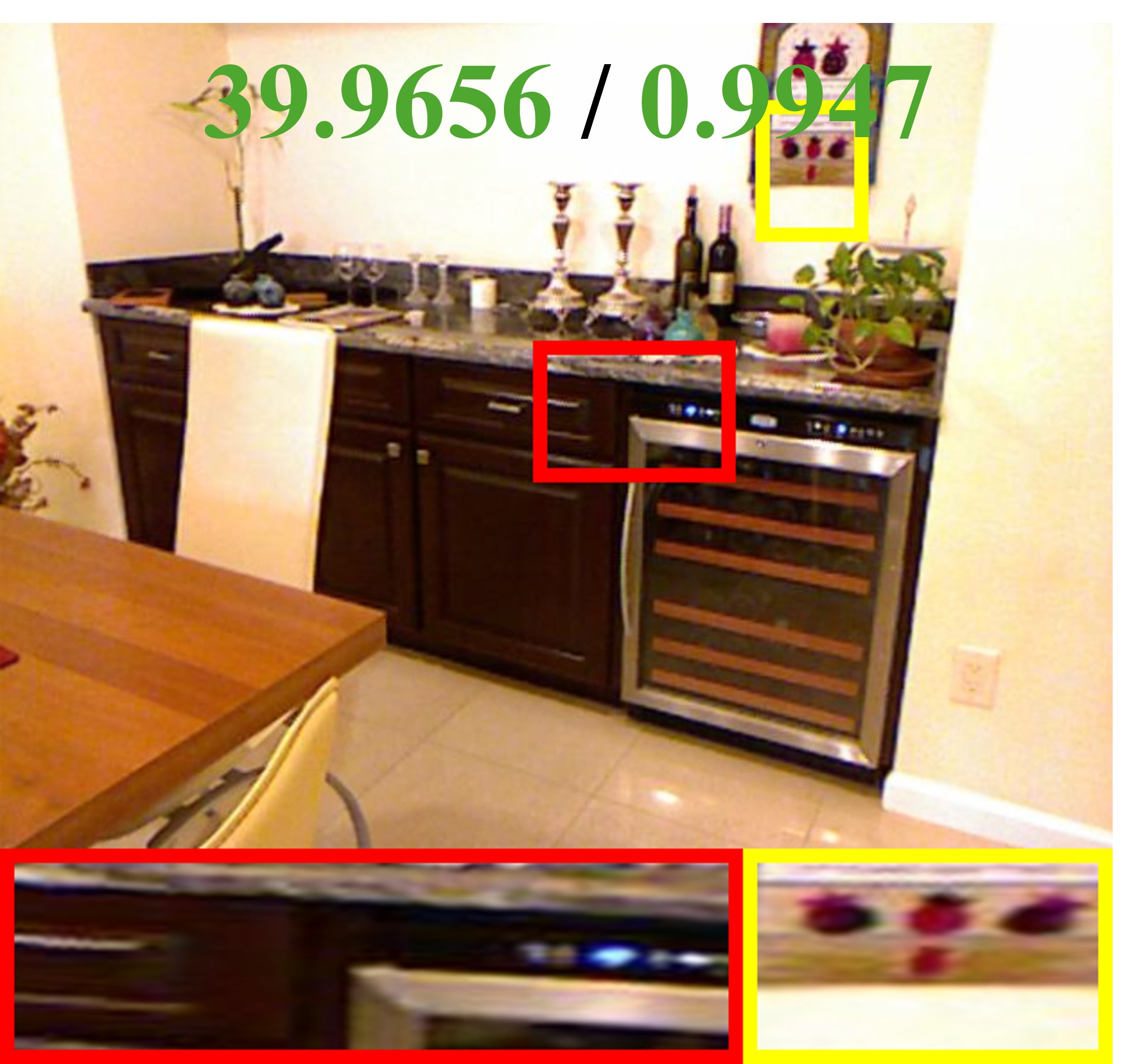}} &
			\adjustbox{valign=m}{\includegraphics[width=\imgw]{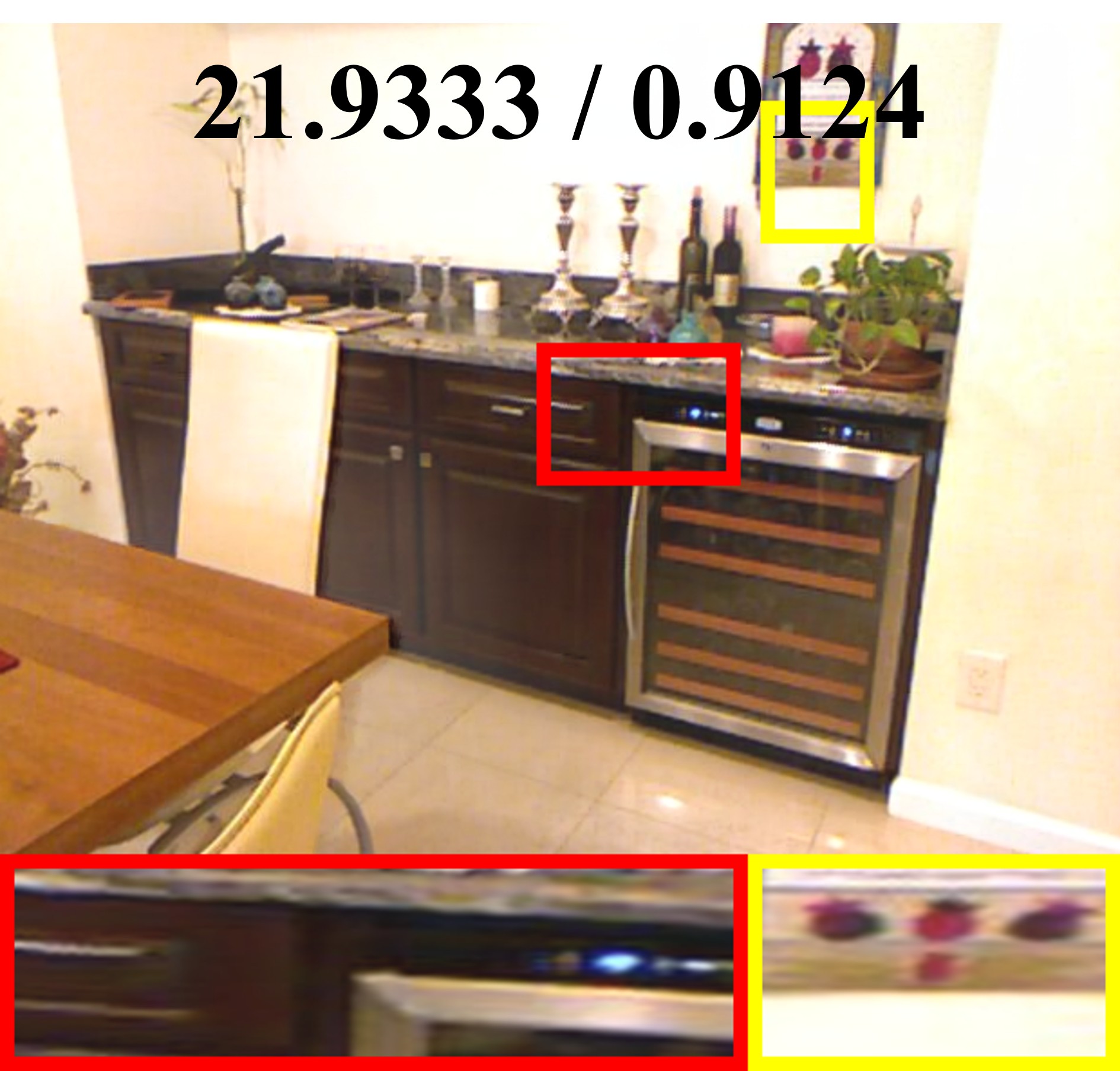}} &
			\adjustbox{valign=m}{\includegraphics[width=\imgw]{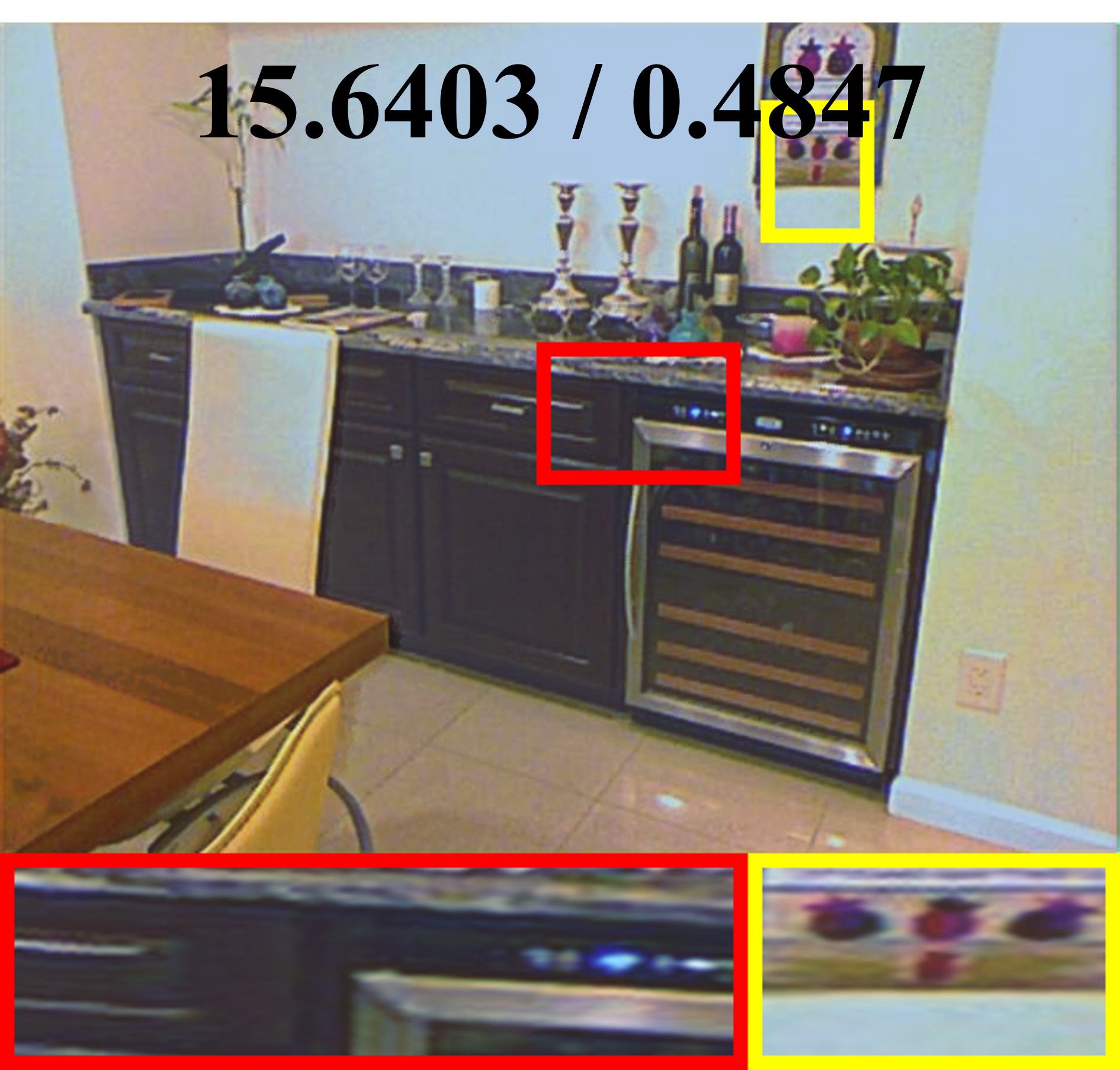}} &
			\adjustbox{valign=m}{\includegraphics[width=\imgw]{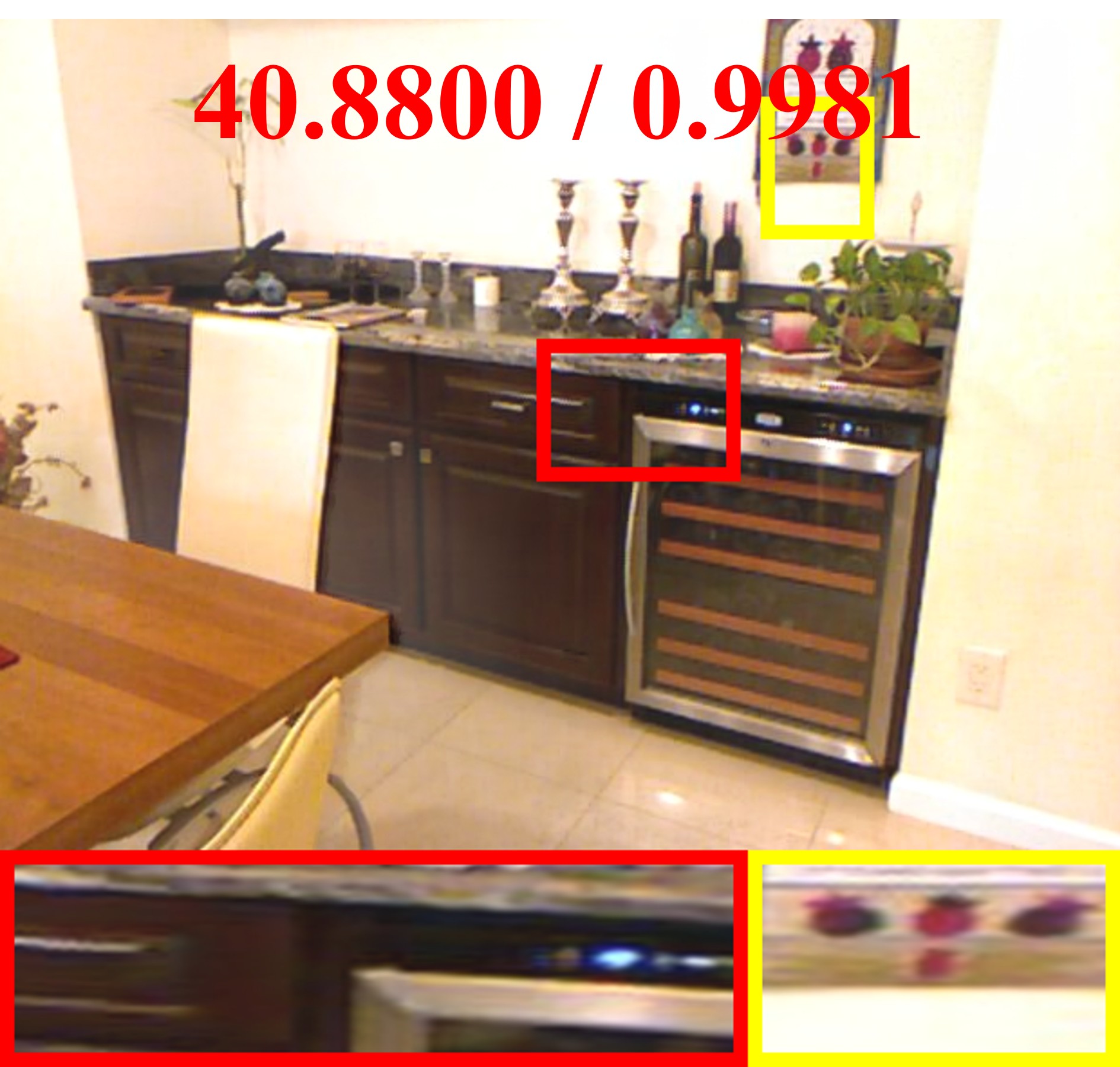}} &
			\adjustbox{valign=m}{\includegraphics[width=\imgw]{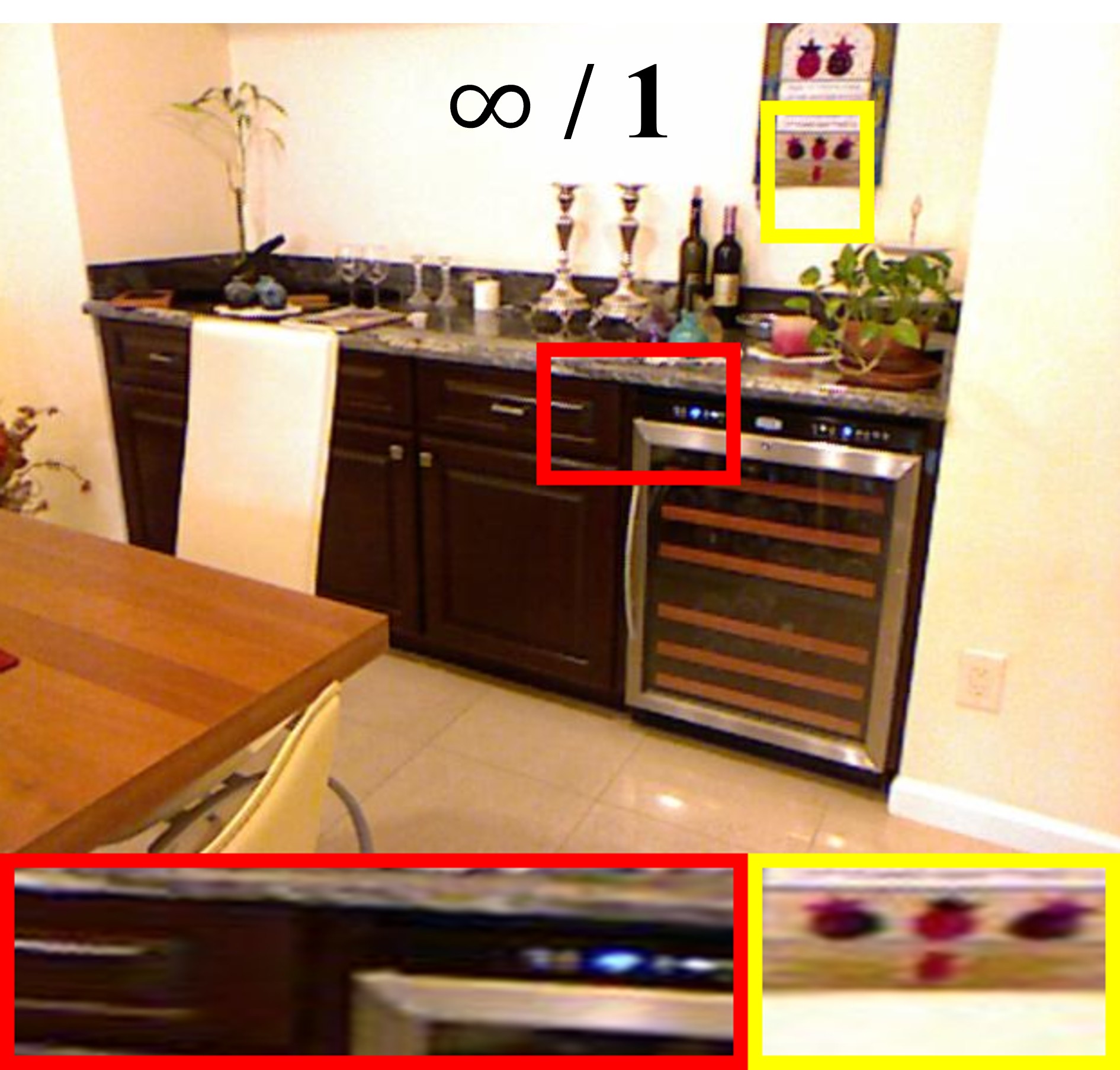}} \\
			
			\rlabel{RESIDE-6K} &
			\adjustbox{valign=m}{\includegraphics[width=\imgw]{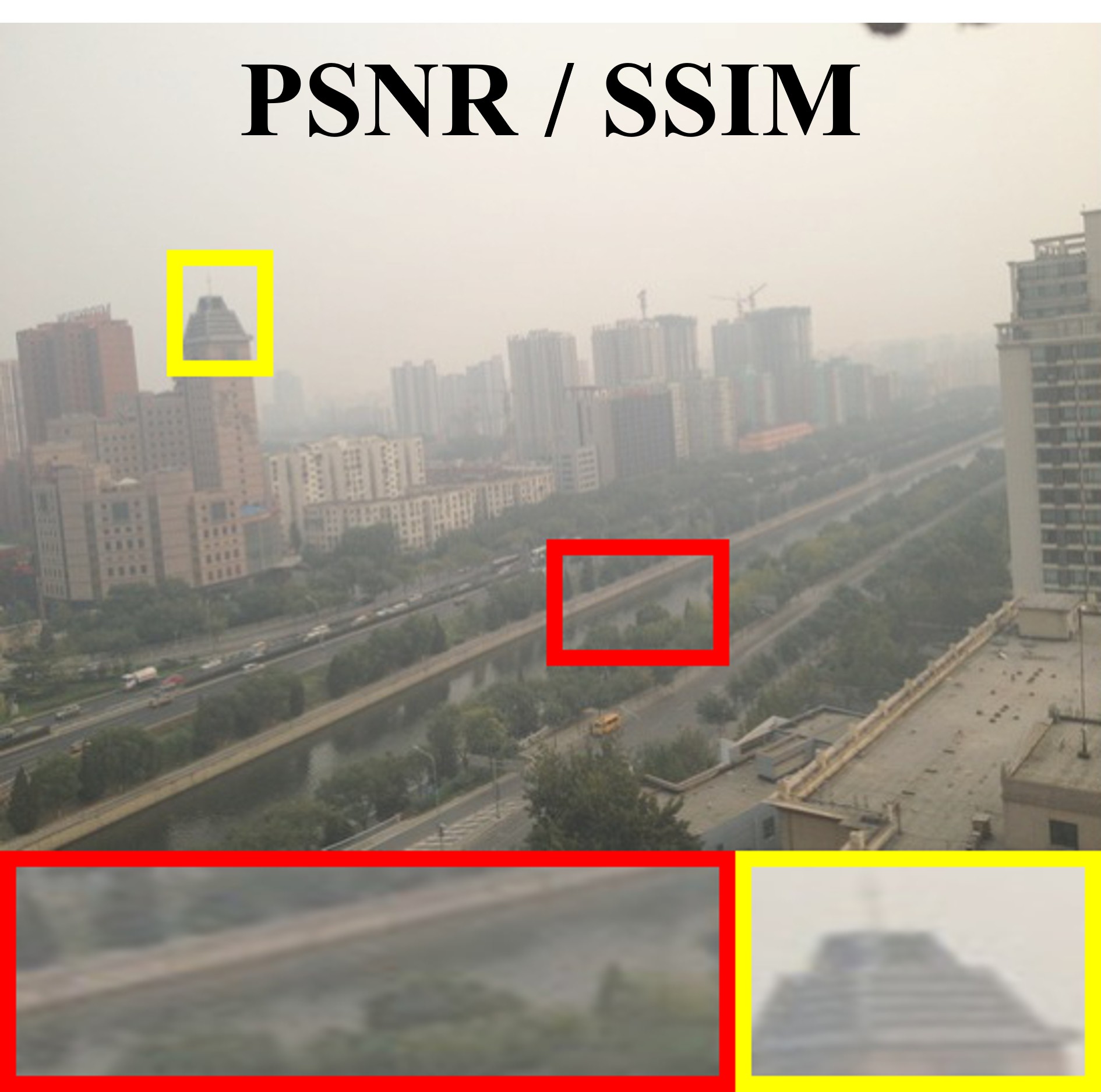}} &
			\adjustbox{valign=m}{\includegraphics[width=\imgw]{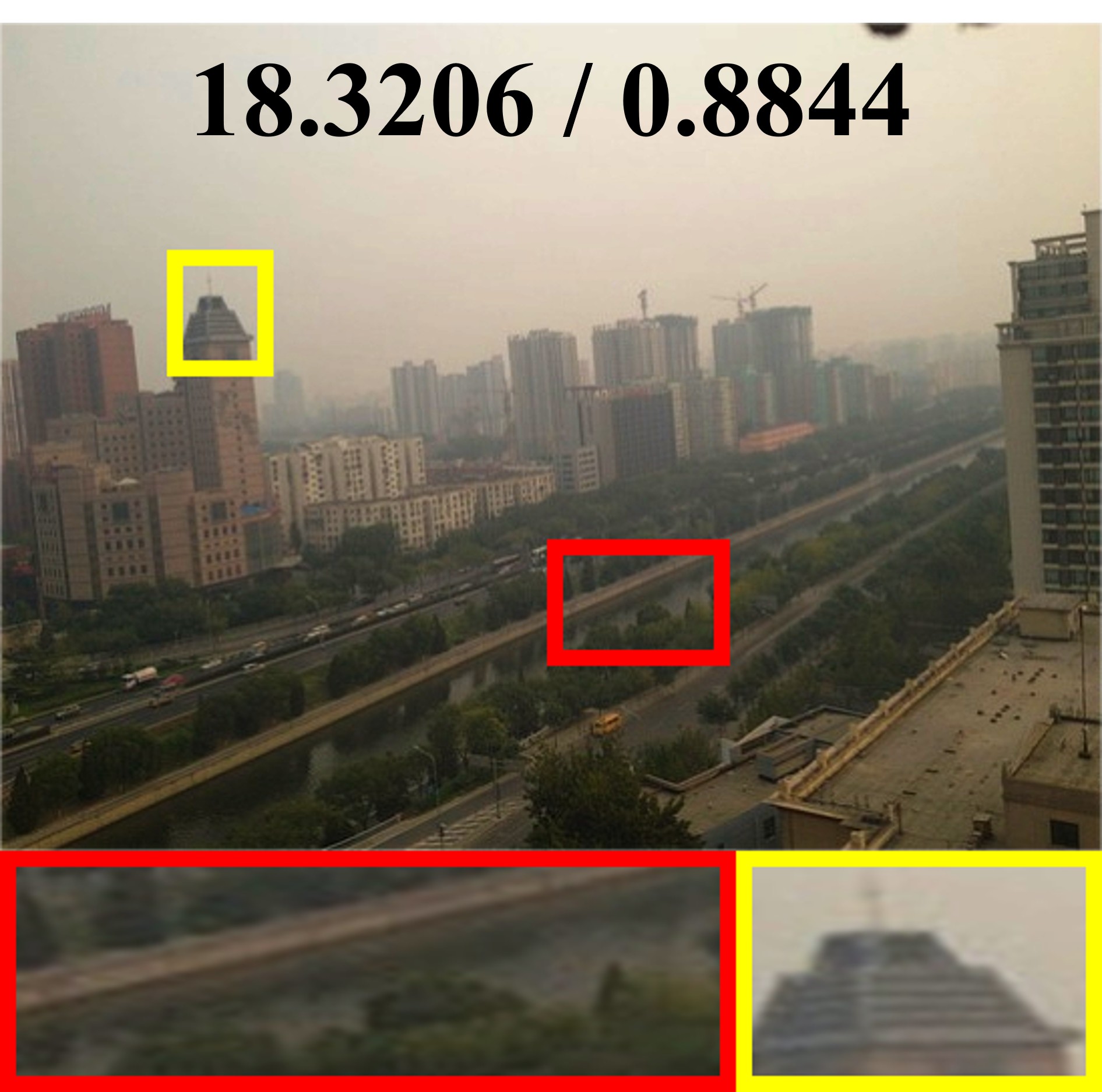}} &
			\adjustbox{valign=m}{\includegraphics[width=\imgw]{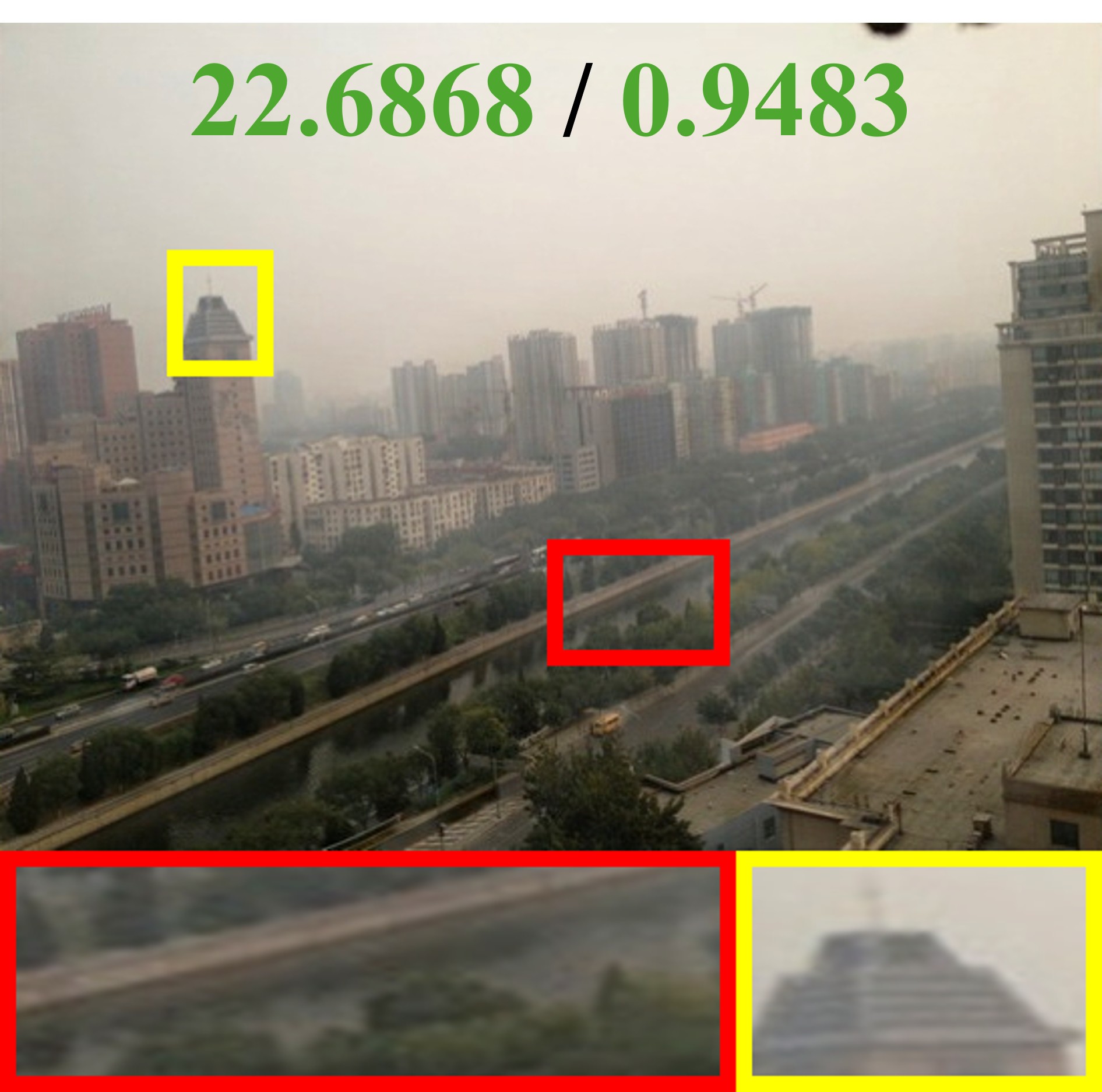}} &
			\adjustbox{valign=m}{\includegraphics[width=\imgw]{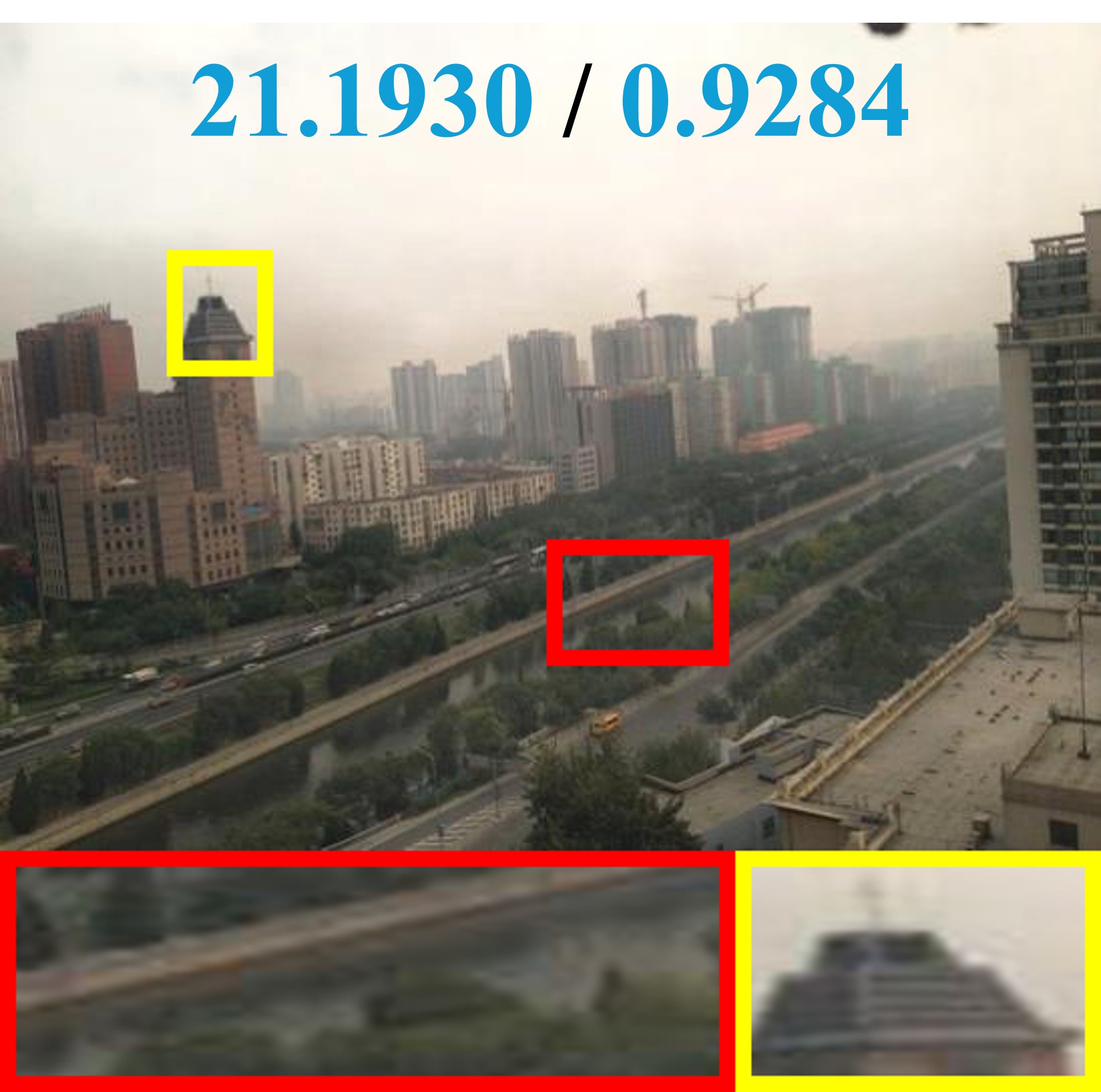}} &
			\adjustbox{valign=m}{\includegraphics[width=\imgw]{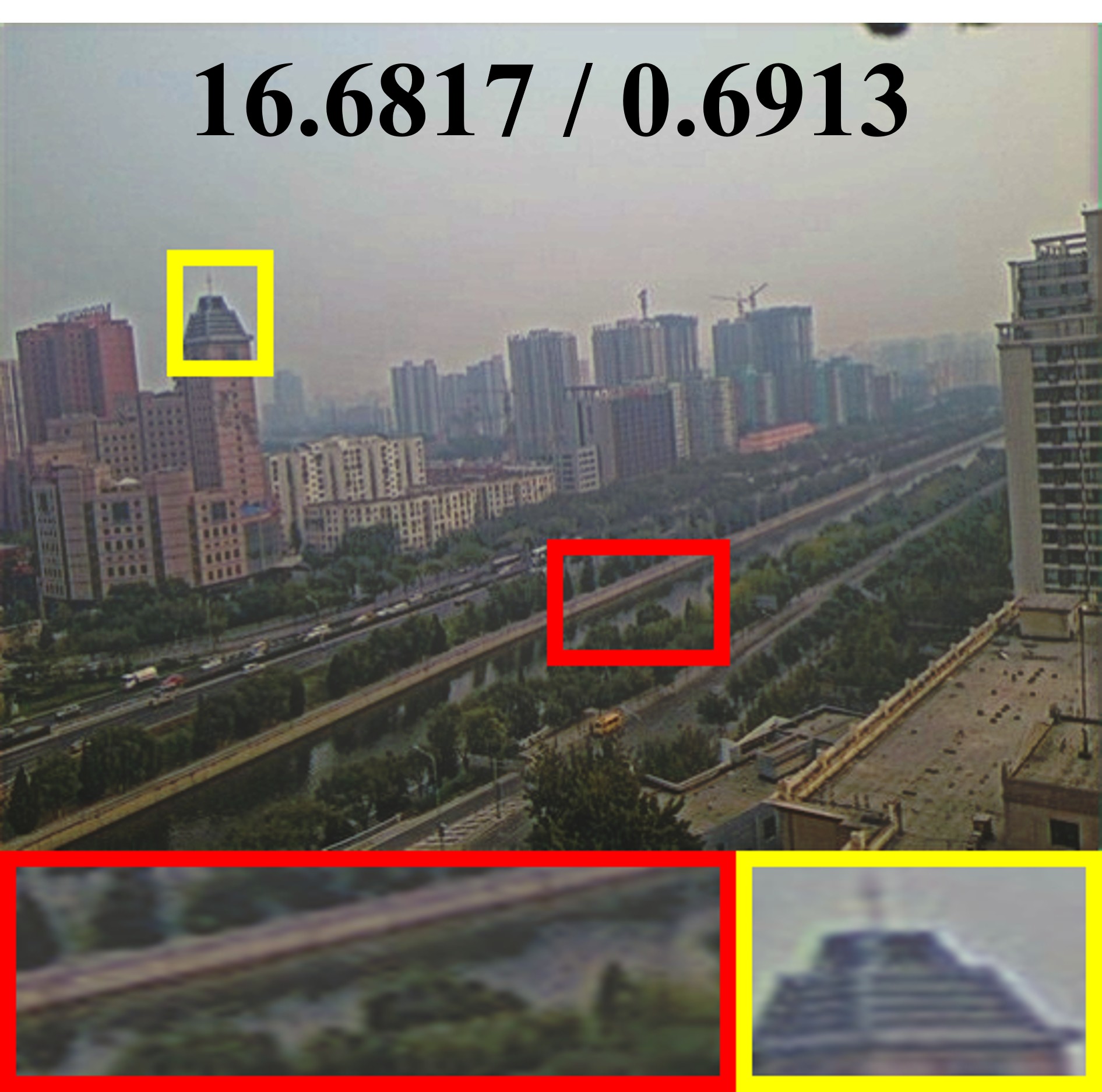}} &
			\adjustbox{valign=m}{\includegraphics[width=\imgw]{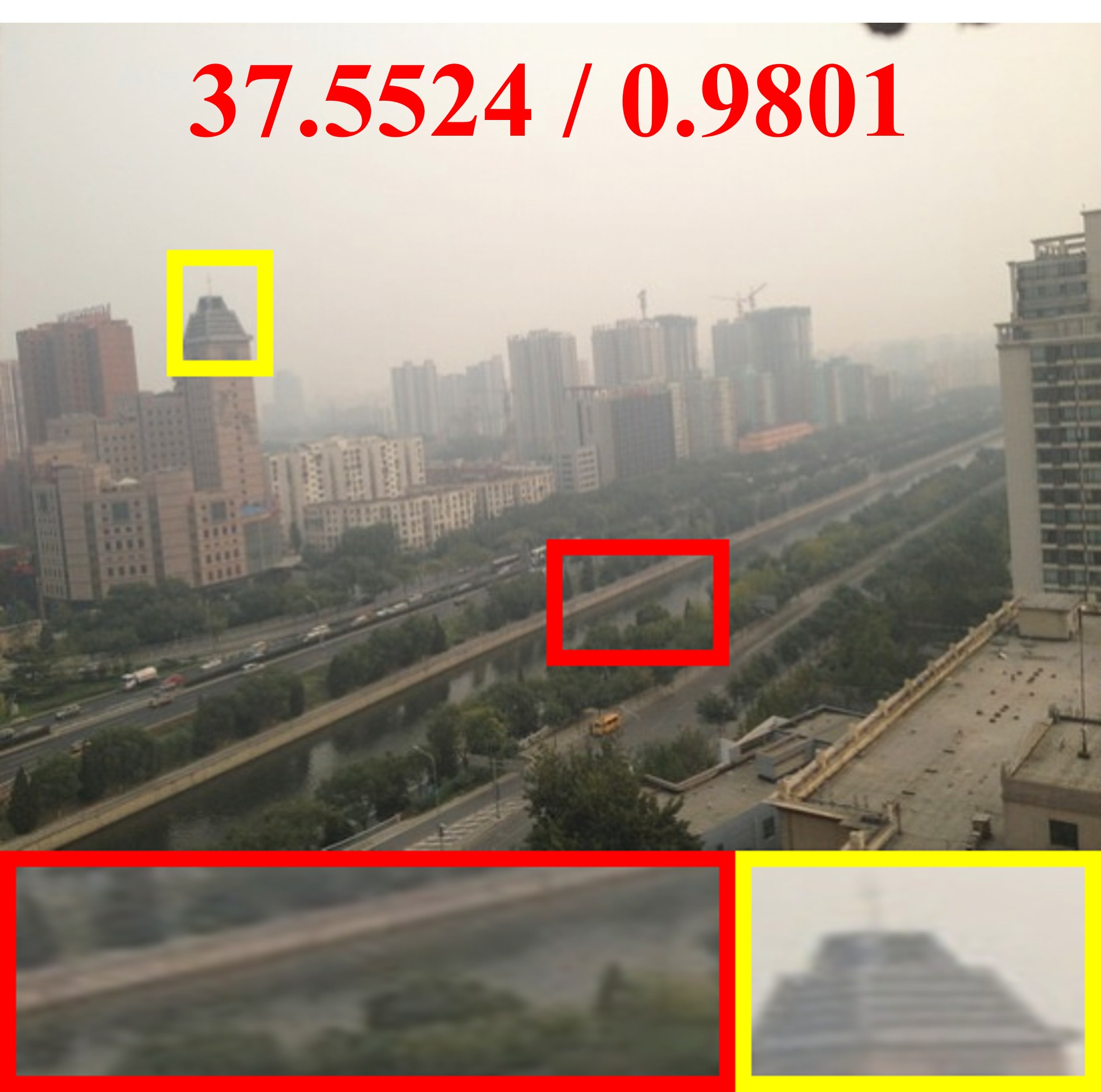}} &
			\adjustbox{valign=m}{\includegraphics[width=\imgw]{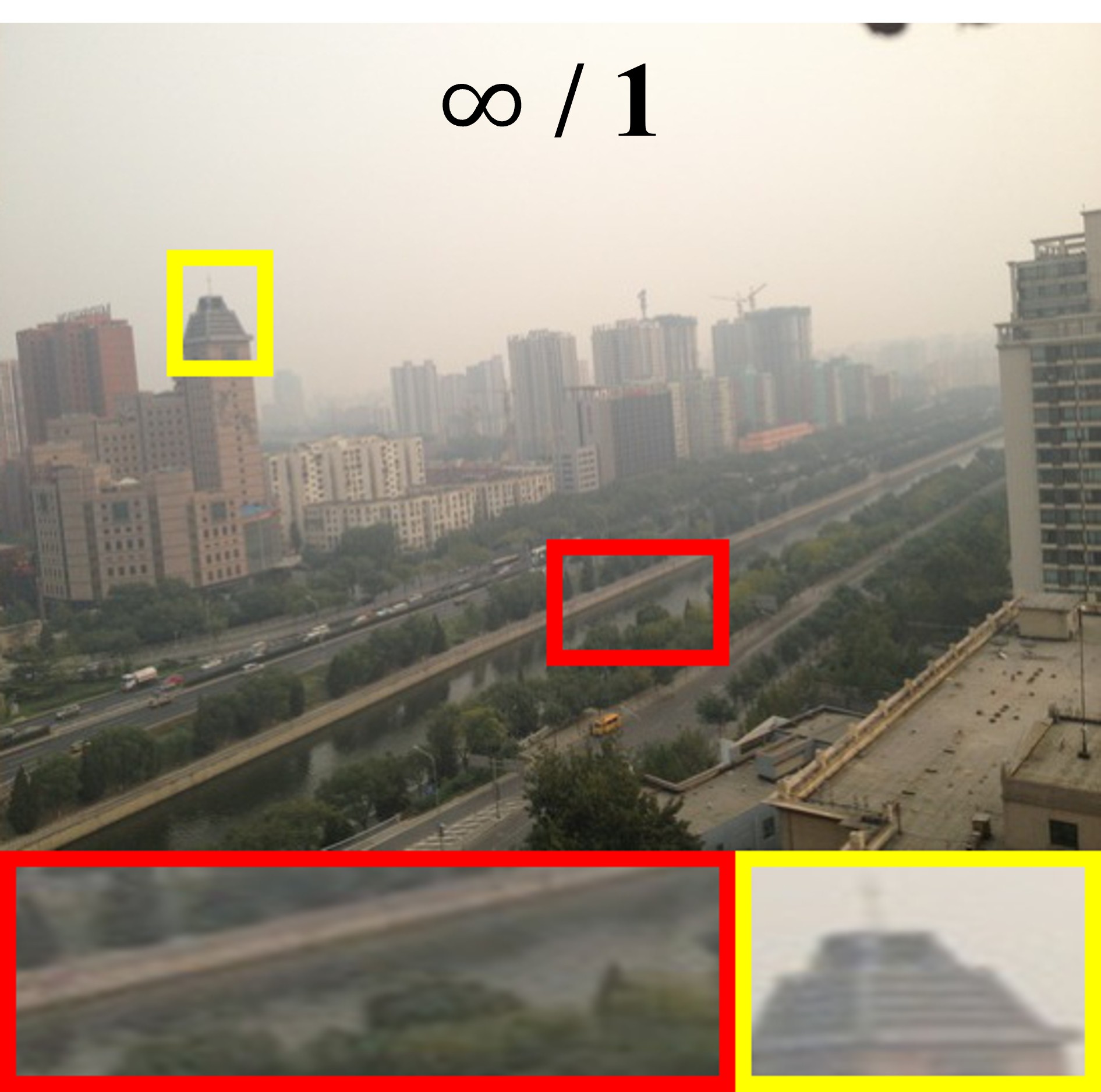}}
		\end{tabular}
	\end{adjustbox}
	\caption{Visual comparison of different dehazing networks on the ITS and RESIDE-6K datasets. The first and second rows correspond to ITS and RESIDE-6K, respectively. PP-Net$_{\mathrm{P}}$ produces clearer structures, more faithful textures, and fewer residual haze artifacts.}
	\label{fig:ppnetp_visual}
\end{figure*}

Fig.~\ref{fig:ppnetp_visual} shows representative qualitative examples. Compared with the competing dehazing networks, PP-Net$_{\mathrm{P}}$ produces clearer structures, more faithful textures, and fewer residual haze artifacts, especially in dense haze regions.

\subsection{Robustness of PP-Net Under Noisy Degradation}
\label{subsec:noisy_eval}

To evaluate robustness under joint haze-and-noise degradation, we further test PP-Net on noisy ITS and noisy RESIDE-6K. This setting corresponds to the second-stage training process, where mixed Poisson--Gaussian noise is introduced to simulate practical biomedical acquisition noise.

\begin{table}[htbp]
	\centering
	\caption{Quantitative comparison between PP-Net and representative learning-based dehazing methods on ITS and RESIDE-6K under \textbf{the haze-and-noise setting}. Higher PSNR/SSIM indicates better quality. {\color{red}\textbf{Red}}, {\color{darkgreen}\textbf{green}}, and {\color{blue}\textbf{blue}} indicate the {\color{red}\textbf{best}}, {\color{darkgreen}\textbf{second-best}}, and {\color{blue}\textbf{third-best}} results.}
	\label{tab:ppnet_noisy}
	\scriptsize
	\setlength{\tabcolsep}{3.5pt}
	\renewcommand{\arraystretch}{1.10}
	\resizebox{\columnwidth}{!}{
	\begin{tabular}{c|cc|cc}
		\hline
		\hline
		\multicolumn{5}{c}{\textbf{Noisy Haze Case}} \\
		\hline
		\multirow{2}{*}{Method}
		& \multicolumn{2}{c|}{ITS \cite{RESIDE}}
		& \multicolumn{2}{c}{RESIDE-6K \cite{RESIDE}} \\
		\cline{2-5}
		& SSIM$\uparrow$ & PSNR$\uparrow$
		& SSIM$\uparrow$ & PSNR$\uparrow$ \\
		\hline
		(ECCV'16) MSCNN \cite{MSCNN}    & 0.1023 & 11.0782 & 0.0728 & 12.1120 \\
		(ICCV'17) AOD-Net \cite{AOD-Net}  & 0.0638 & 12.6801 & 0.1199 & 12.4083 \\
		(CVPR'18) GFN \cite{GFN}      & 0.0892 & 10.2018 & 0.0924 & 11.0352 \\
		(CVPR'20) MSBDN \cite{MSBDN}     & {\color{blue}\textbf{0.1246}} & 11.3412 & 0.1021 & 11.4012 \\
		(ECCV'20) PFDN \cite{PFDN}      & 0.1103 & 12.0119 & 0.1240 & 10.2147 \\
		(AAAI'21) FFA-Net \cite{FFA-Net}  & 0.0423 & 10.5311 & 0.0928 & 10.1249 \\
		(TETCI'24) TBN \cite{TBN}     & 0.1089 & {\color{darkgreen}\textbf{14.0227}} & {\color{darkgreen}\textbf{0.1994}} & {\color{darkgreen}\textbf{14.4773}} \\
		(TIM'25) CDVA \cite{CDVA}      & 0.1023 & 11.3123 & {\color{blue}\textbf{0.1732}} & {\color{blue}\textbf{13.6241}} \\
		(TITS'25) IDB \cite{IDB}      & {\color{darkgreen}\textbf{0.1325}} & {\color{blue}\textbf{13.5427}} & 0.1324 & 11.4512 \\
		(AAAI'25) MPMF-Net \cite{MPMF} & 0.0963 & 12.3733 & 0.1533 & 11.9649 \\
		\textbf{(Ours) PP-Net} & {\color{red}\textbf{0.8093}} & {\color{red}\textbf{24.9645}} & {\color{red}\textbf{0.8236}} & {\color{red}\textbf{25.3423}} \\
		\hline
		\hline
	\end{tabular}%
}
\end{table}

Table~\ref{tab:ppnet_noisy} shows that PP-Net significantly outperforms all compared methods on both datasets. The result verifies the importance of introducing DFN-Net before scattering-map estimation, prior-map recovery, and GF-Net refinement. Front-end noise suppression provides more reliable inputs for ASAP-based scattering-map estimation and GF-Net refinement, improving PP-Net stability when noise and scattering coexist.

\begin{figure*}[htbp]
	\centering
	\includegraphics[width=0.98\textwidth]{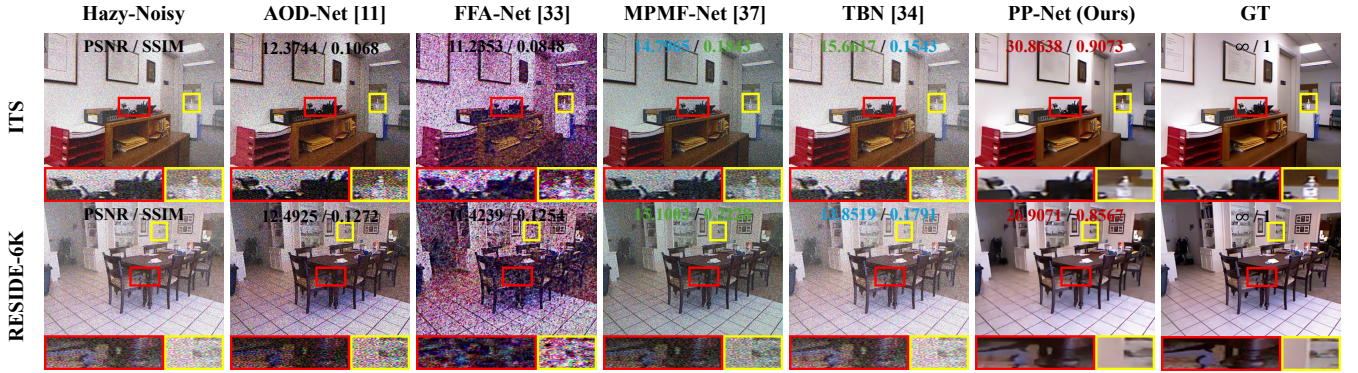}
	\caption{Visual comparison under joint haze-and-noise degradation on the ITS and RESIDE-6K datasets. The first and second rows correspond to ITS and RESIDE-6K, respectively. The proposed method effectively suppresses haze and noise while preserving clearer structures and more faithful details.}
	\label{fig:ppnet_noisy_visual}
\end{figure*}

Fig.~\ref{fig:ppnet_noisy_visual} provides representative qualitative comparisons. PP-Net simultaneously removes haze and suppresses noise, producing cleaner structures and more stable visual restoration than the competing methods.

\subsection{No-Reference Evaluation on Real Biomedical Images}
\label{subsec:biomedical_eval}

Finally, we evaluate PP-Net on the W2S biomedical dataset under different averaging/noise levels, including avg1, avg4, avg16, and avg400. These settings represent different acquisition conditions and are used to examine the robustness of the proposed framework on real biomedical images. Since paired scattering-free ground truth is unavailable, PP-Net is compared with the degraded inputs and representative prior-based baselines using no-reference quality metrics.

NIQE and BRISQUE are adopted as no-reference metrics, where lower values indicate better perceptual quality. As reported in Table~\ref{tab:w2s_no_reference}, PP-Net achieves the best NIQE scores across all four averaging levels, indicating improved naturalness and perceptual quality according to the NIQE criterion. The BRISQUE results are less consistent across different averaging levels, suggesting that different no-reference metrics may emphasize different image statistics. Therefore, the no-reference evaluation on real biomedical images is interpreted together with qualitative visual evidence.

\begin{table*}[htbp]
	\centering
	\caption{No-reference evaluation on the W2S dataset under different averaging/noise levels. Lower NIQE and BRISQUE values indicate better perceptual quality. {\color{red}\textbf{Red}}, {\color{darkgreen}\textbf{green}}, and {\color{blue}\textbf{blue}} denote the {\color{red}\textbf{best}}, {\color{darkgreen}\textbf{second-best}}, and {\color{blue}\textbf{third-best}} results, respectively.}
	\label{tab:w2s_no_reference}
	\scriptsize
	\setlength{\tabcolsep}{4.5pt}
	\renewcommand{\arraystretch}{1.2}
	\resizebox{\textwidth}{!}{%
		\begin{tabular}{c|cc|cc|cc|cc}
			\hline
			\hline
			\multirow{3}{*}{Method}
			& \multicolumn{8}{c}{(ECCVW'20) Widefield2SIM\cite{ws2_dataset}} \\
			\cline{2-9}
			& \multicolumn{2}{c|}{avg1}
			& \multicolumn{2}{c|}{avg4}
			& \multicolumn{2}{c|}{avg16}
			& \multicolumn{2}{c}{avg400} \\
			\cline{2-9}
			& NIQE$\downarrow$ & BRISQUE$\downarrow$
			& NIQE$\downarrow$ & BRISQUE$\downarrow$
			& NIQE$\downarrow$ & BRISQUE$\downarrow$
			& NIQE$\downarrow$ & BRISQUE$\downarrow$ \\
			\hline
			INPUT                    & 15.4081 & {\color{blue}\textbf{41.4164}} & 13.8807 & {\color{darkgreen}\textbf{37.4247}} & 11.1614 & {\color{darkgreen}\textbf{31.7543}} & 6.9195 & {\color{blue}\textbf{32.8347}} \\
			(TPAMI'11) DCP           & 15.3282 & {\color{darkgreen}\textbf{41.3114}} & {\color{darkgreen}\textbf{13.7268}} & {\color{red}\textbf{37.1064}} & {\color{darkgreen}\textbf{10.9883}} & {\color{red}\textbf{31.1809}} & {\color{darkgreen}\textbf{6.8975}} & 33.3567 \\
			(ISBI'23) HDCP           & {\color{darkgreen}\textbf{12.3006}} & 41.4668 & {\color{blue}\textbf{13.7673}} & {\color{blue}\textbf{37.7673}} & {\color{blue}\textbf{11.0959}} & {\color{blue}\textbf{31.8450}} & 6.9037 & {\color{darkgreen}\textbf{32.2912}} \\
			(IEEE Access'25) QDCP    & {\color{blue}\textbf{15.3019}} & 41.4695 & 13.7730 & 37.5326 & 11.0971 & 31.8459 & {\color{blue}\textbf{6.9012}} & {\color{red}\textbf{32.2662}} \\
			(Nature Methods'25) Dark & 17.8489 & 43.2703 & 19.9137 & 43.5090 & 17.1220 & 43.4596 & 8.2914 & 35.3332 \\
			\textbf{(Ours) PP-Net} & {\color{red}\textbf{5.5820}}  & {\color{red}\textbf{40.8959}} & {\color{red}\textbf{6.1972}}  & 44.9863 & {\color{red}\textbf{6.4928}}  & 46.4840 & {\color{red}\textbf{6.6210}} & 46.5043 \\
			\hline
			\hline
		\end{tabular}%
	}
\end{table*}

\begin{figure*}[htbp]
	\centering
	\setlength{\tabcolsep}{2pt}
	\renewcommand{\arraystretch}{1.0}
	
	\newcommand{\imgw}{0.145\textwidth}
	\newcommand{\rlabel}[1]{\adjustbox{valign=m}{\rotatebox[origin=c]{90}{\scriptsize\textbf{#1}}}}
	\newcommand{\chead}[1]{\makebox[\imgw][c]{\footnotesize\textbf{#1}}}
	\newcommand{\cheadtwo}[2]{\makebox[\imgw][c]{\shortstack{\footnotesize\textbf{#1}\\[-0.4ex]\footnotesize\textbf{#2}}}}
	
	\begin{adjustbox}{max totalsize={\textwidth}{0.95\textheight},center}
		\begin{tabular}{c cccccc}
			& \chead{Original} &
			\chead{DCP\cite{DCP}} &
			\chead{HDCP\cite{HDCP}} &
			\chead{QDCP\cite{QDCP}} &
			\chead{Dark\cite{Dark}} &
			\chead{PP-Net (Ours)} \\
			
			\rlabel{avg1} &
			\adjustbox{valign=m}{\includegraphics[width=\imgw]{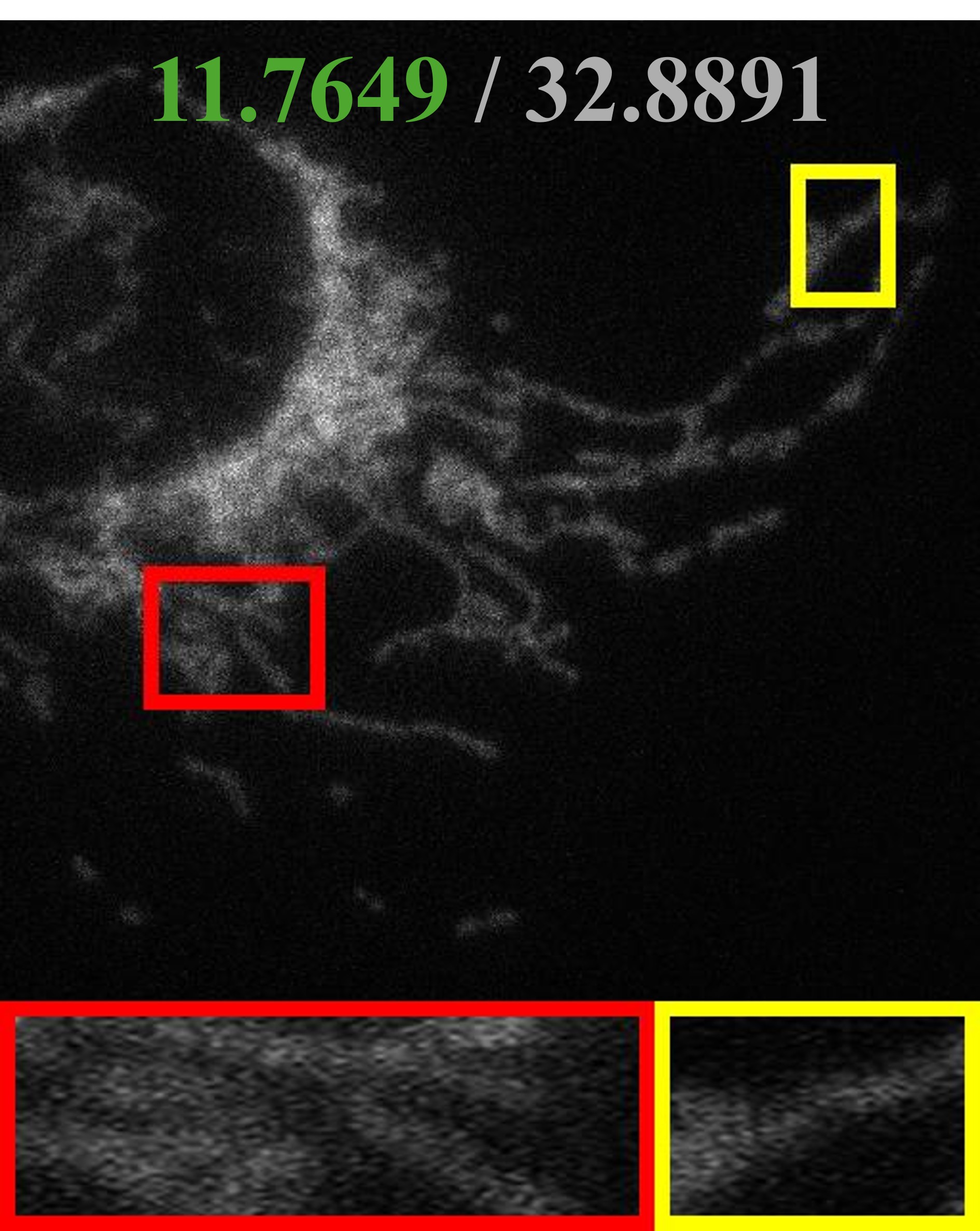}} &
			\adjustbox{valign=m}{\includegraphics[width=\imgw]{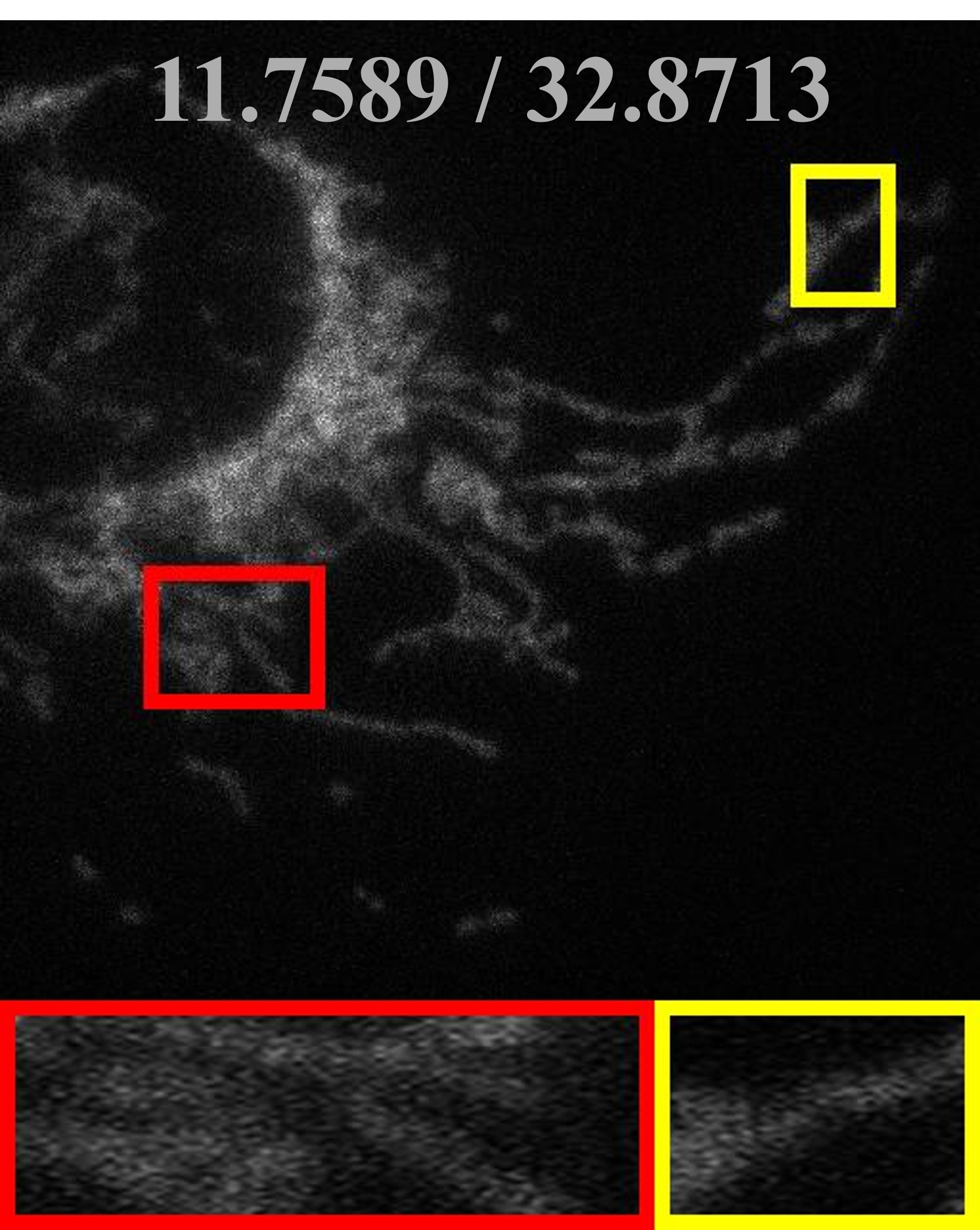}} &
			\adjustbox{valign=m}{\includegraphics[width=\imgw]{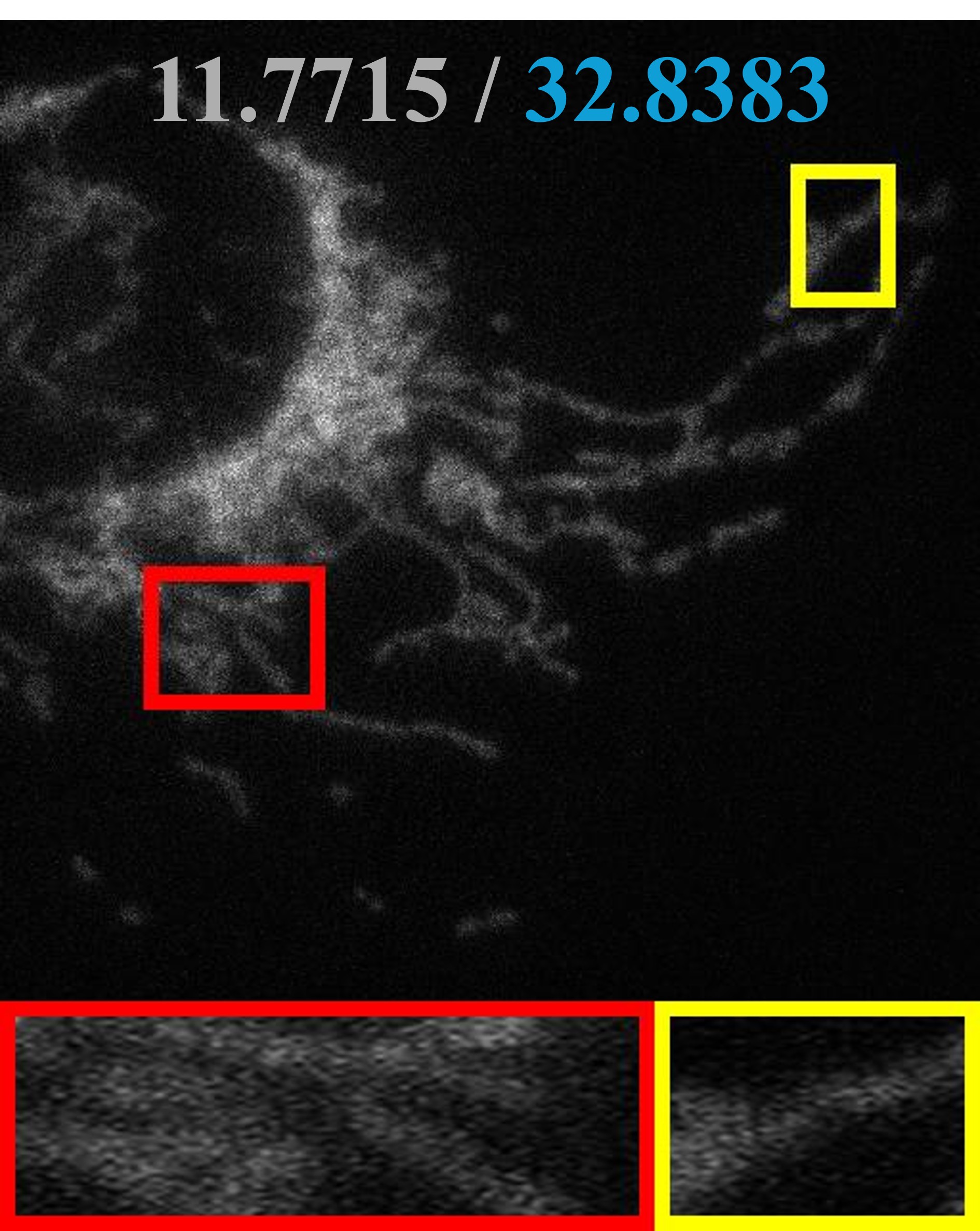}} &
			\adjustbox{valign=m}{\includegraphics[width=\imgw]{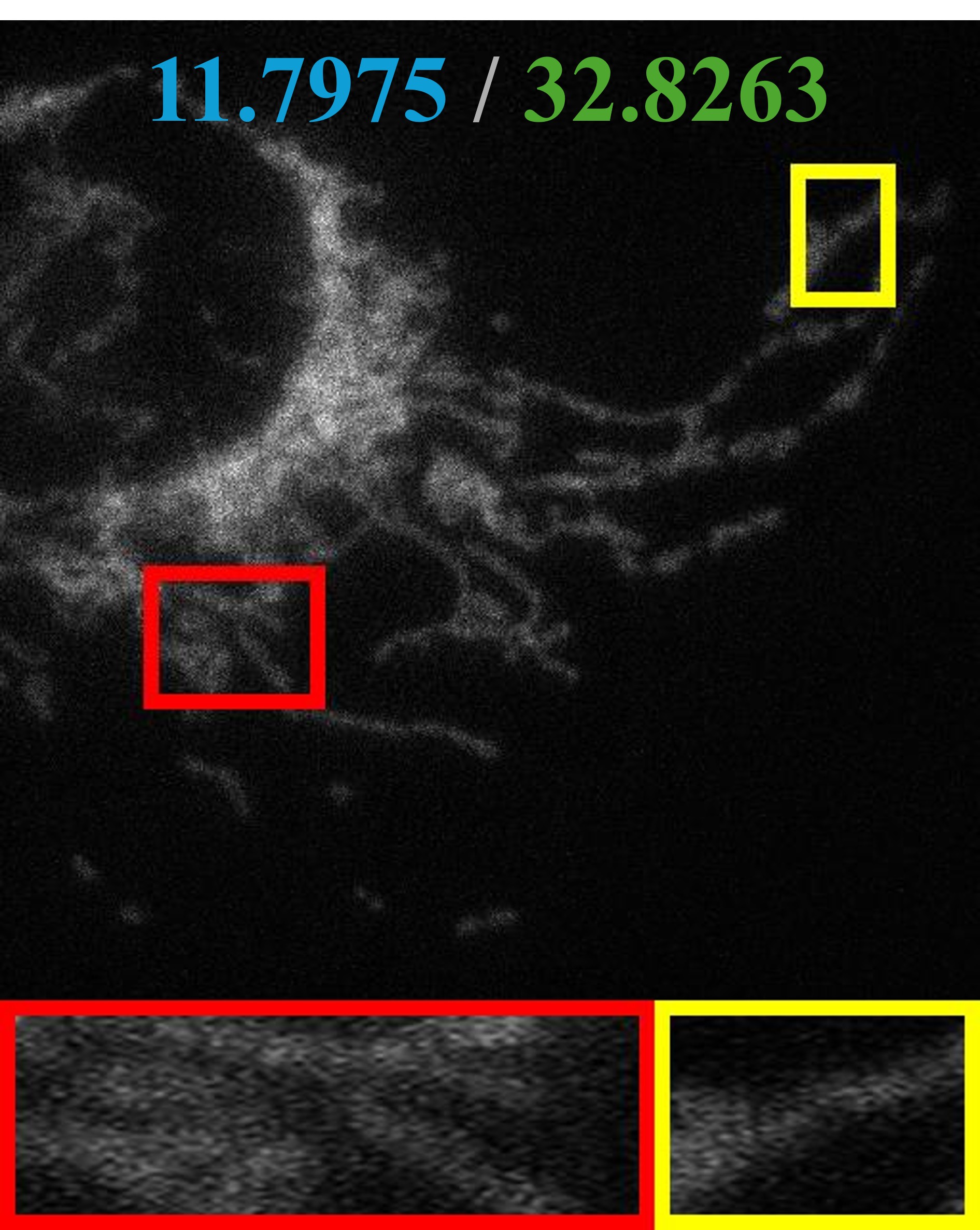}} &
			\adjustbox{valign=m}{\includegraphics[width=\imgw]{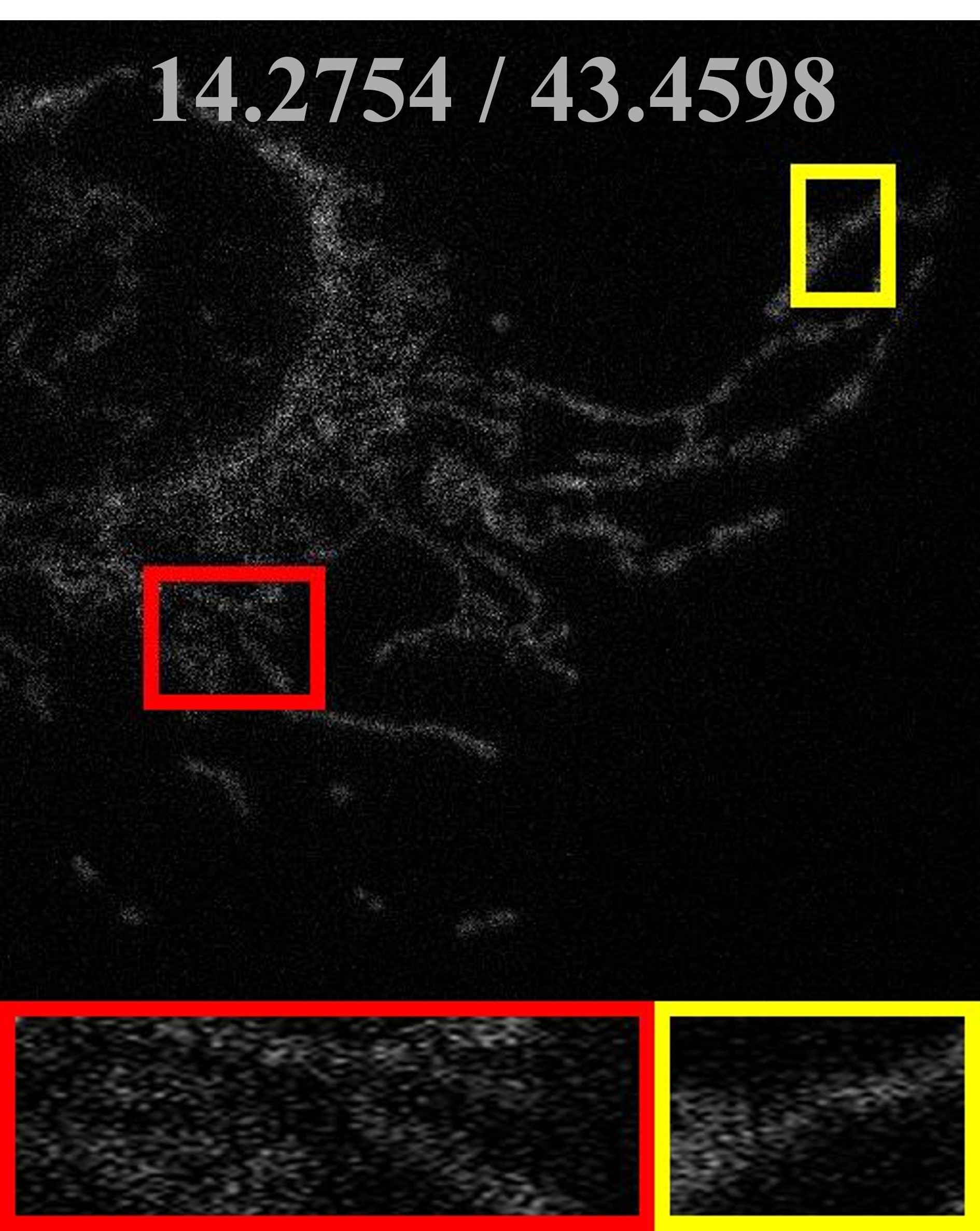}} &
			\adjustbox{valign=m}{\includegraphics[width=\imgw]{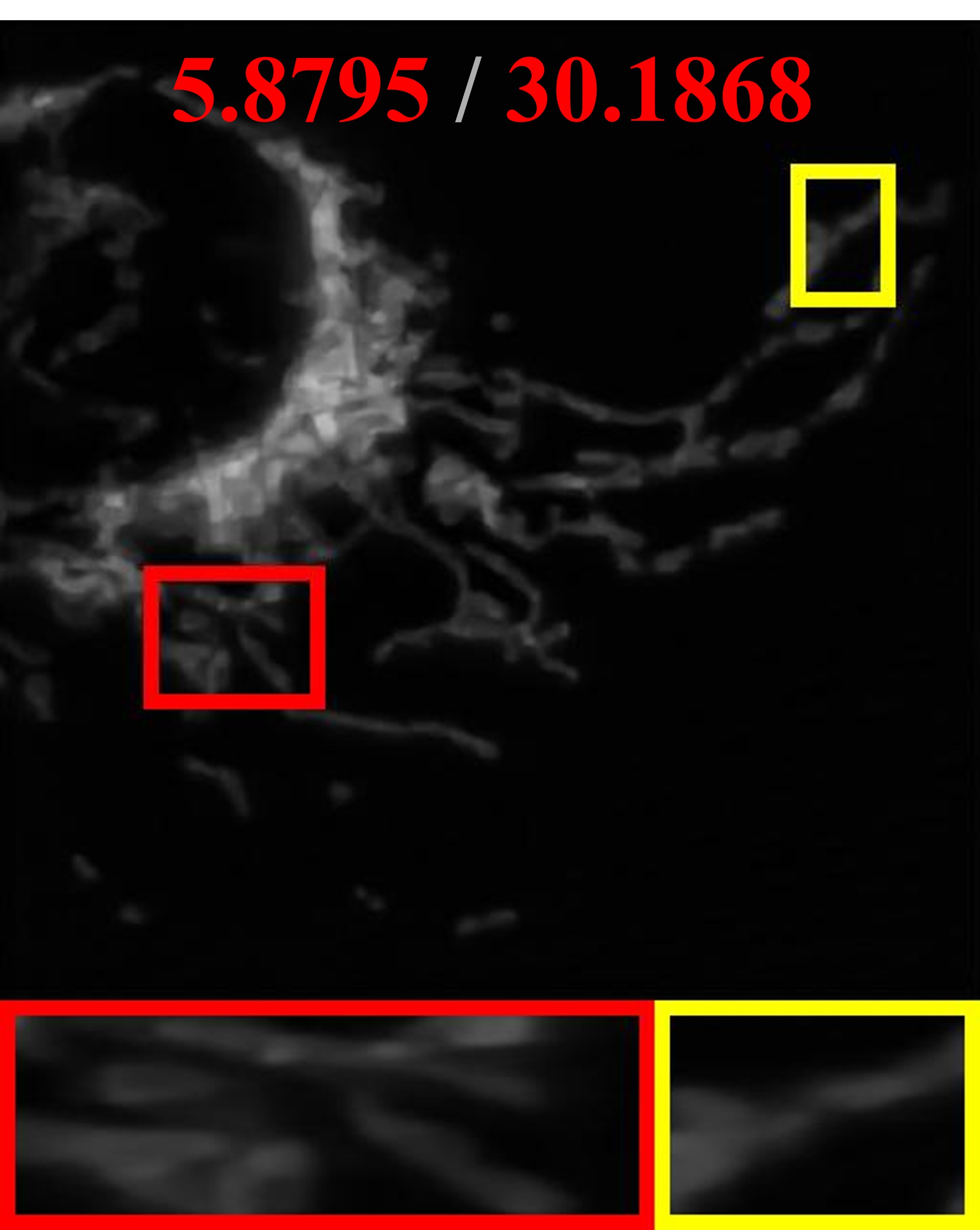}} \\
			
			\rlabel{avg4} &
			\adjustbox{valign=m}{\includegraphics[width=\imgw]{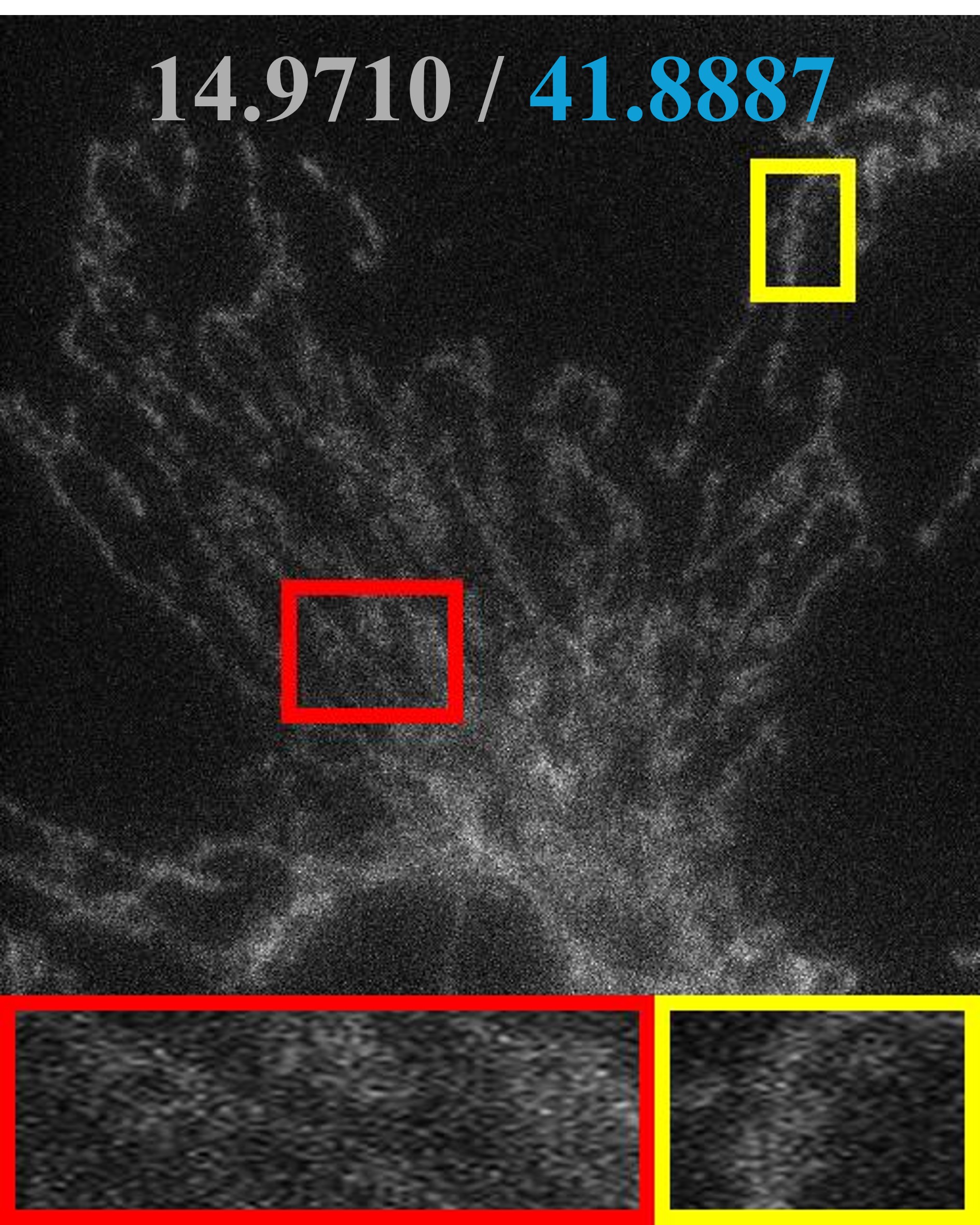}} &
			\adjustbox{valign=m}{\includegraphics[width=\imgw]{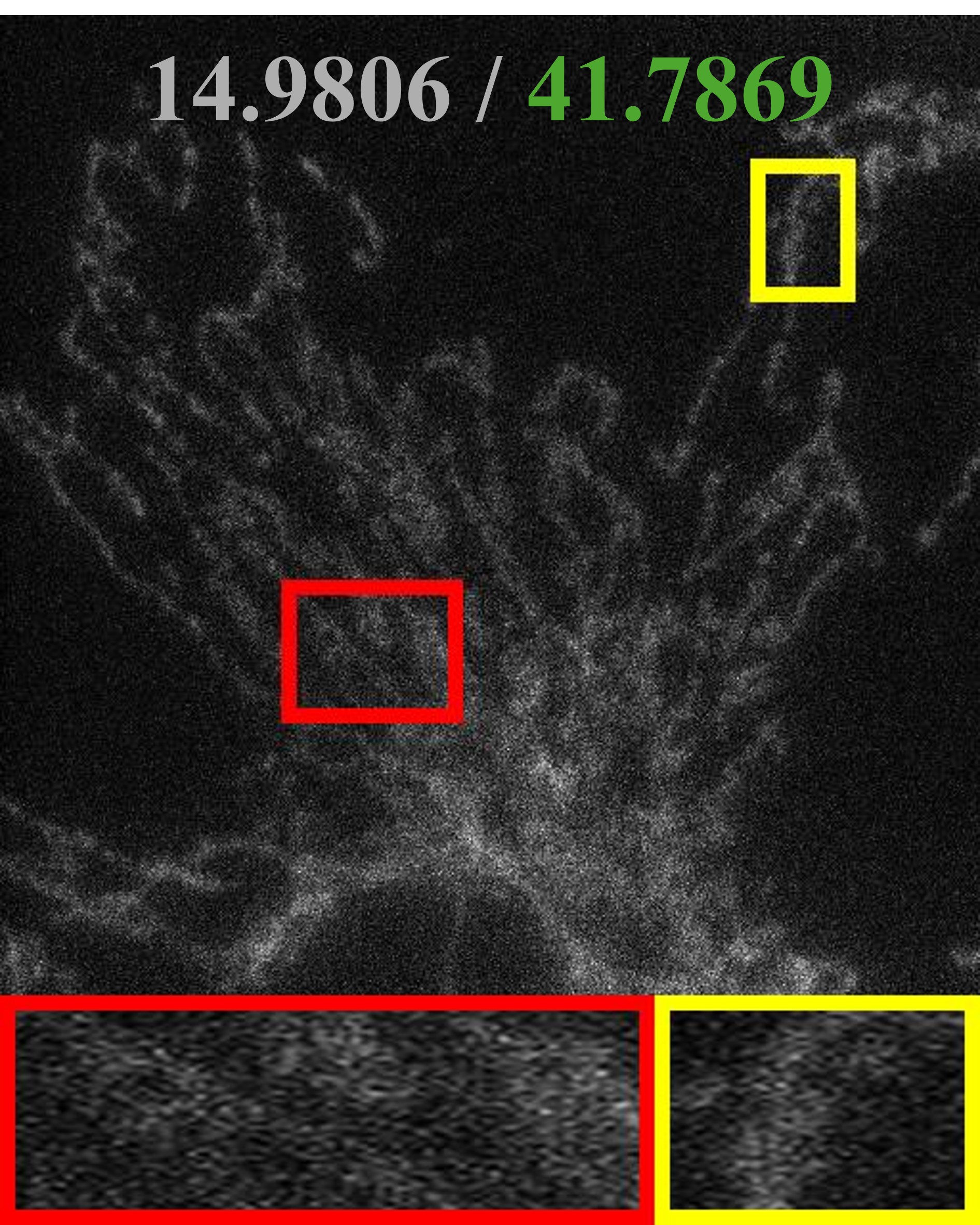}} &
			\adjustbox{valign=m}{\includegraphics[width=\imgw]{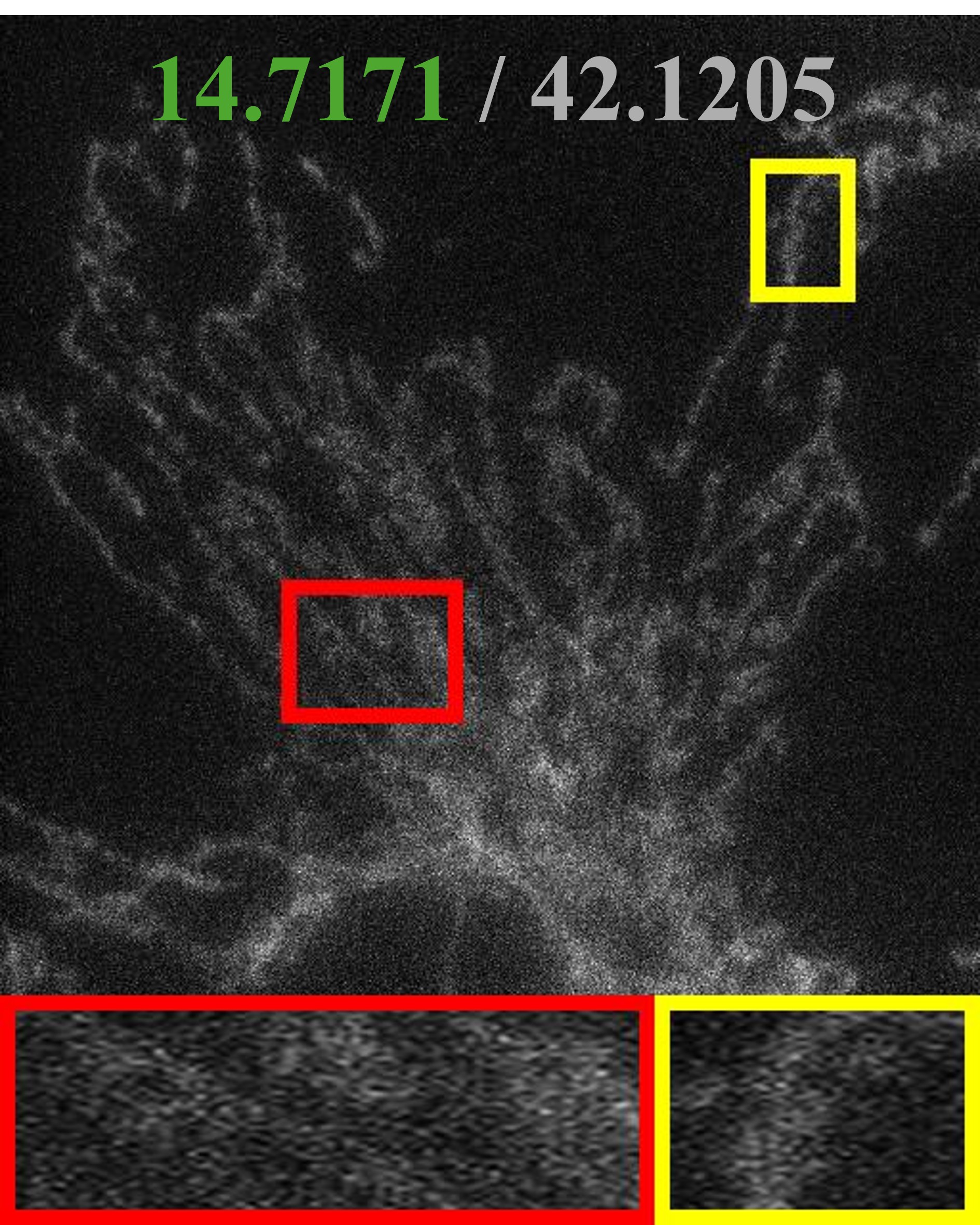}} &
			\adjustbox{valign=m}{\includegraphics[width=\imgw]{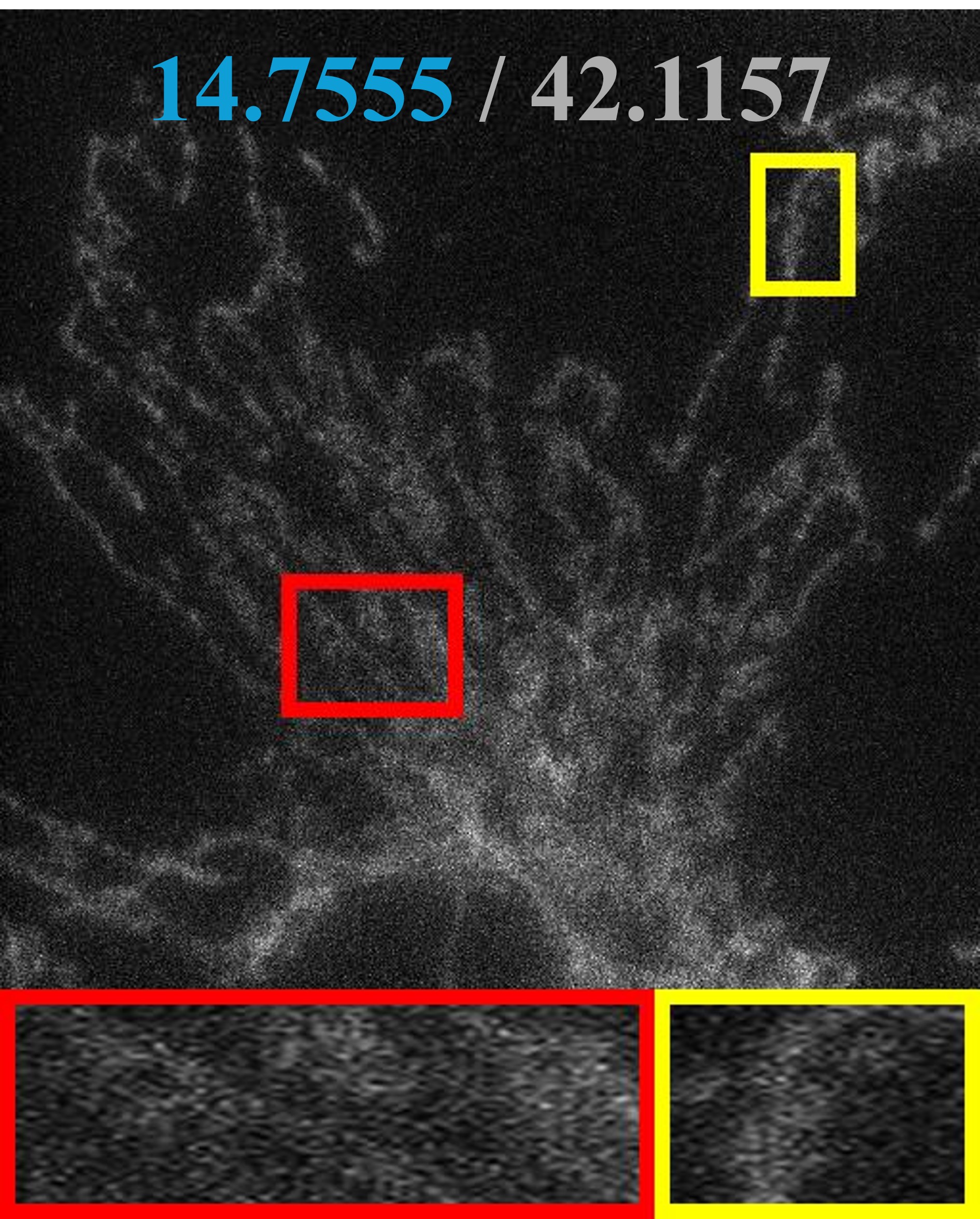}} &
			\adjustbox{valign=m}{\includegraphics[width=\imgw]{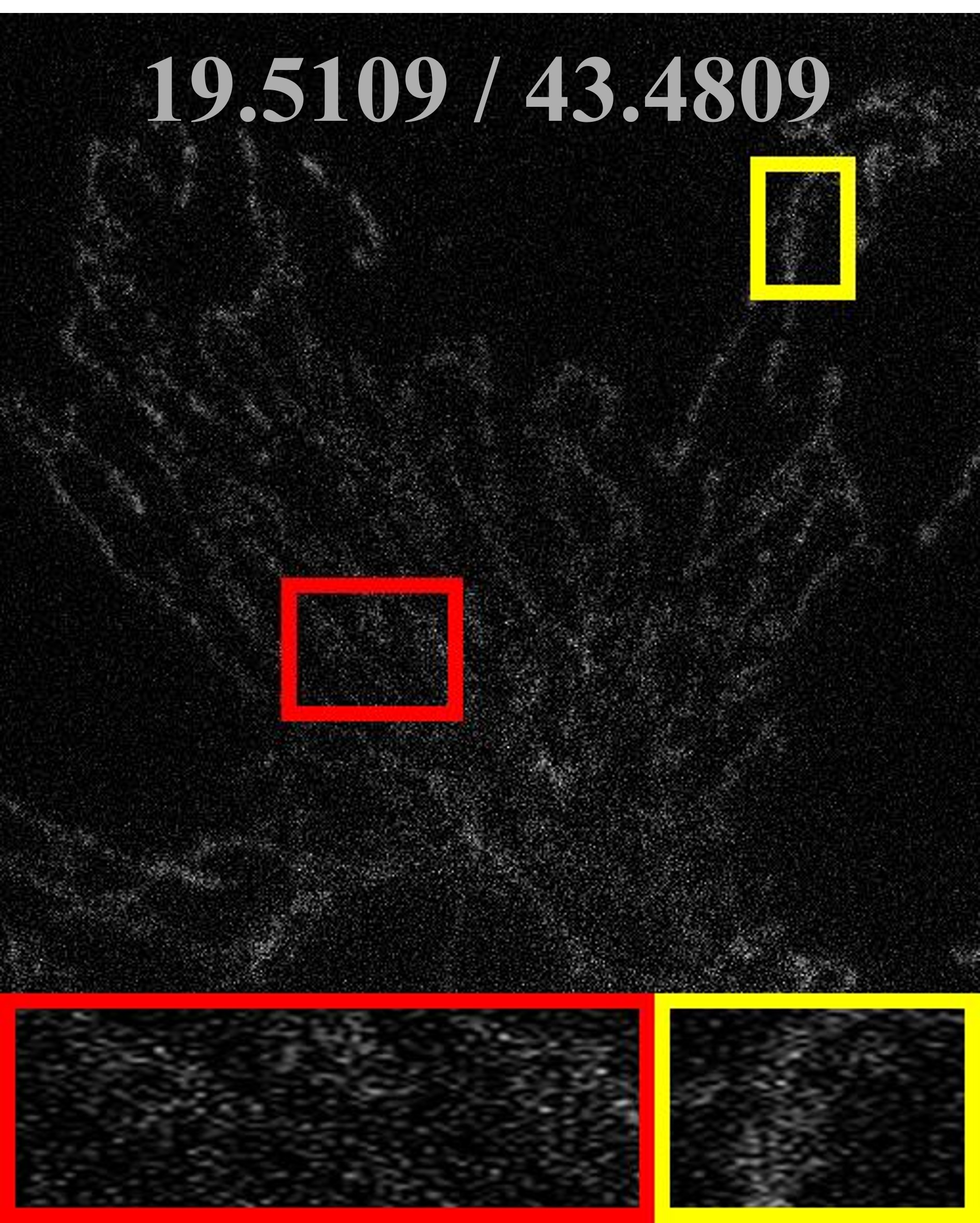}} &
			\adjustbox{valign=m}{\includegraphics[width=\imgw]{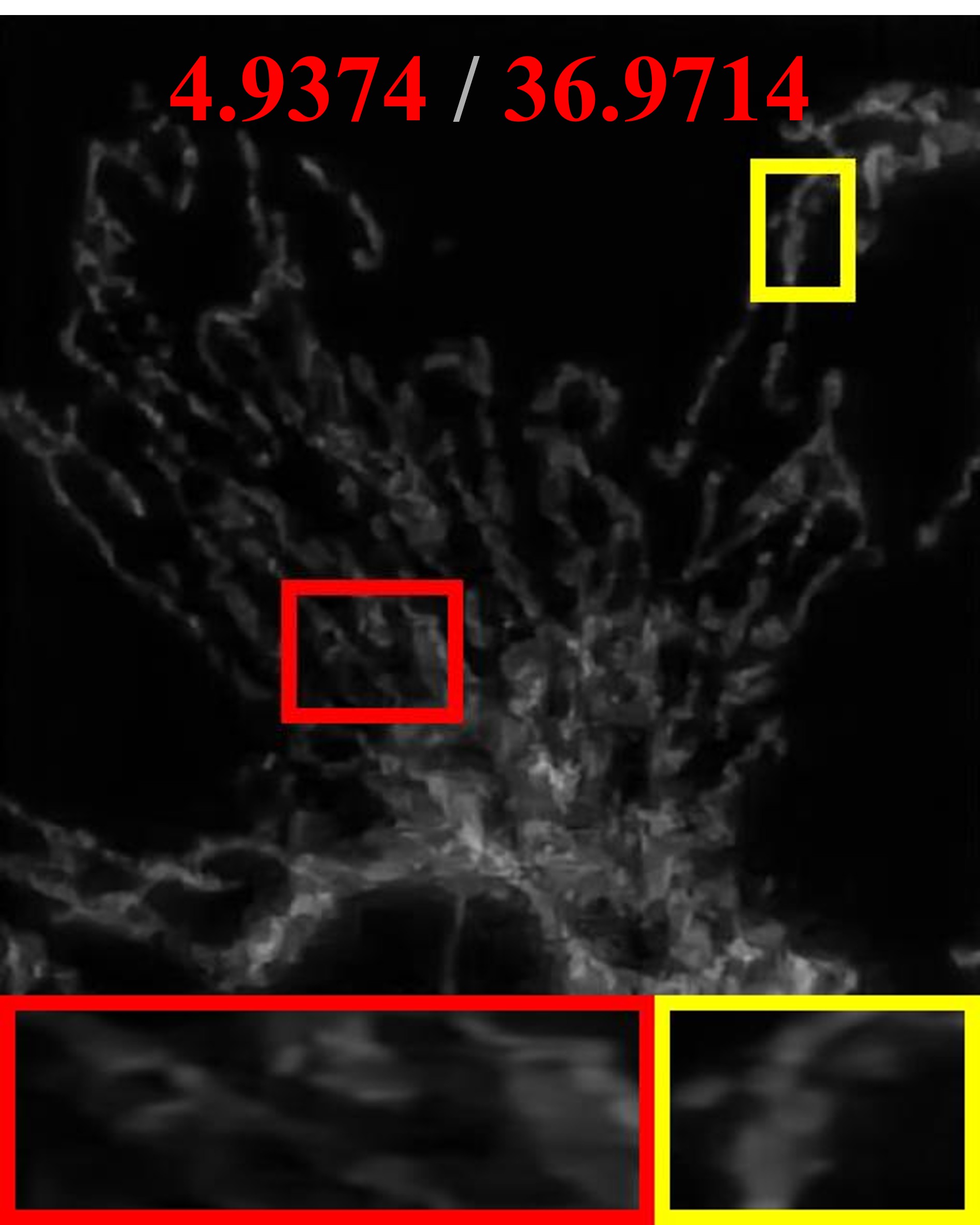}} \\
			
			\rlabel{avg16} &
			\adjustbox{valign=m}{\includegraphics[width=\imgw]{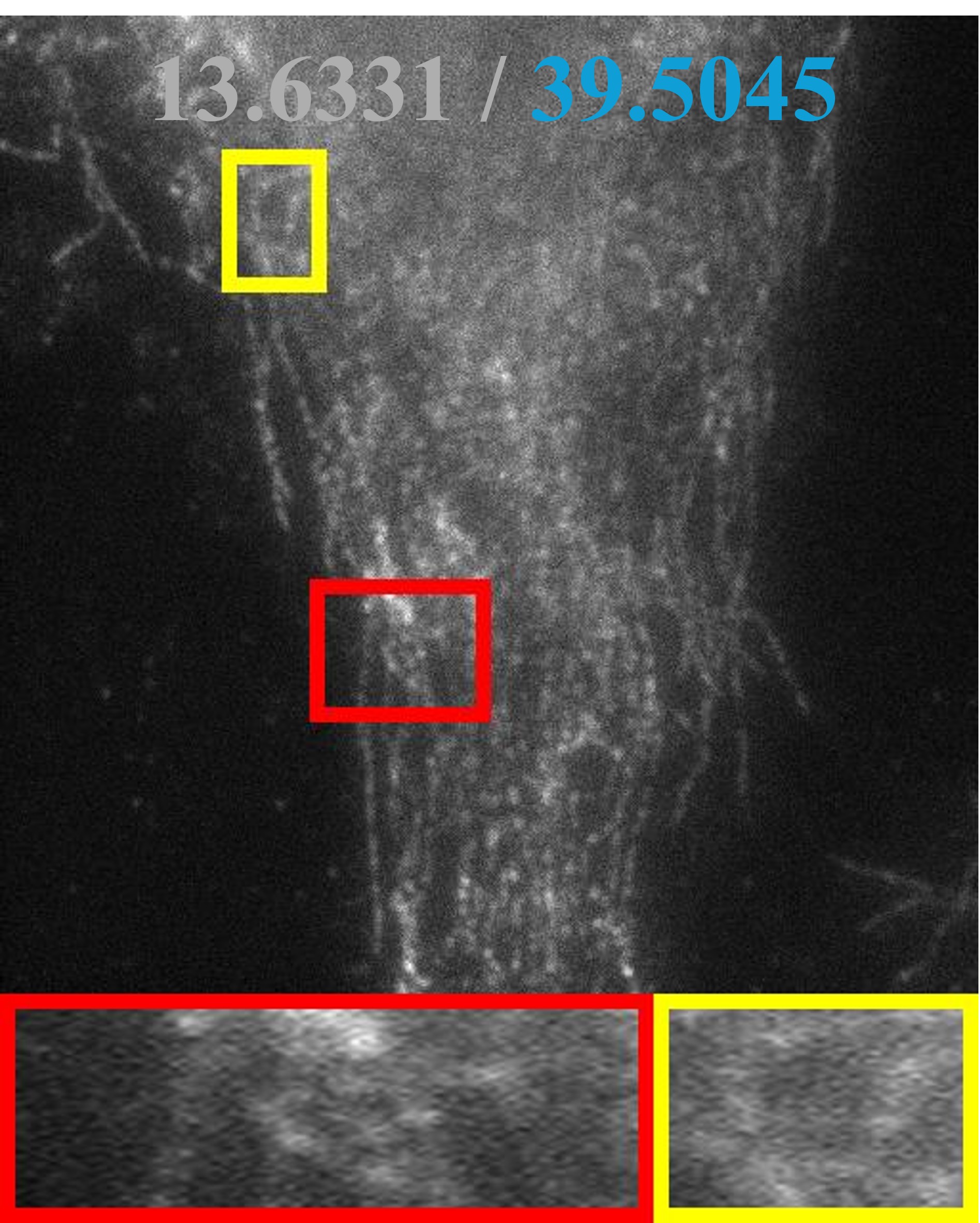}} &
			\adjustbox{valign=m}{\includegraphics[width=\imgw]{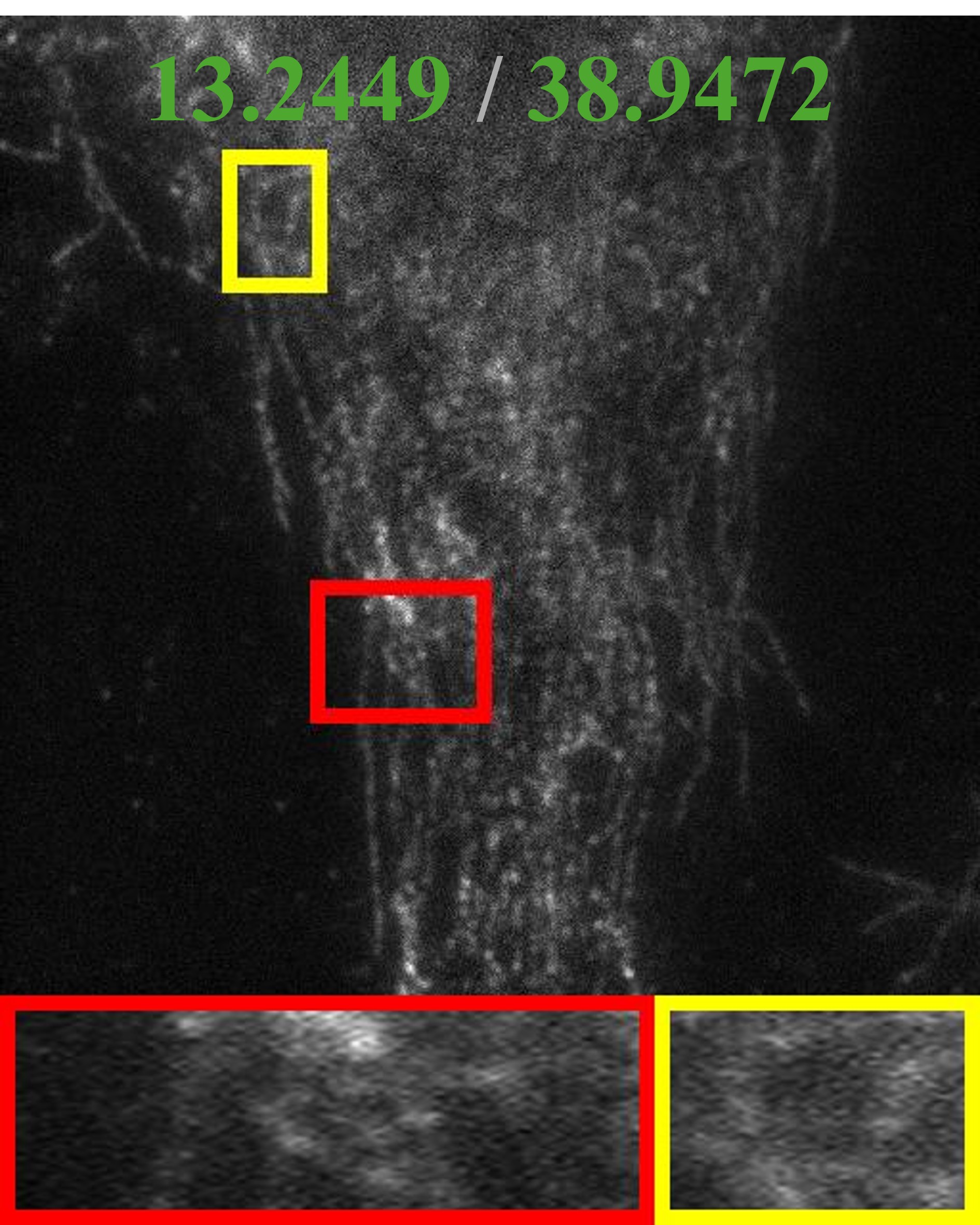}} &
			\adjustbox{valign=m}{\includegraphics[width=\imgw]{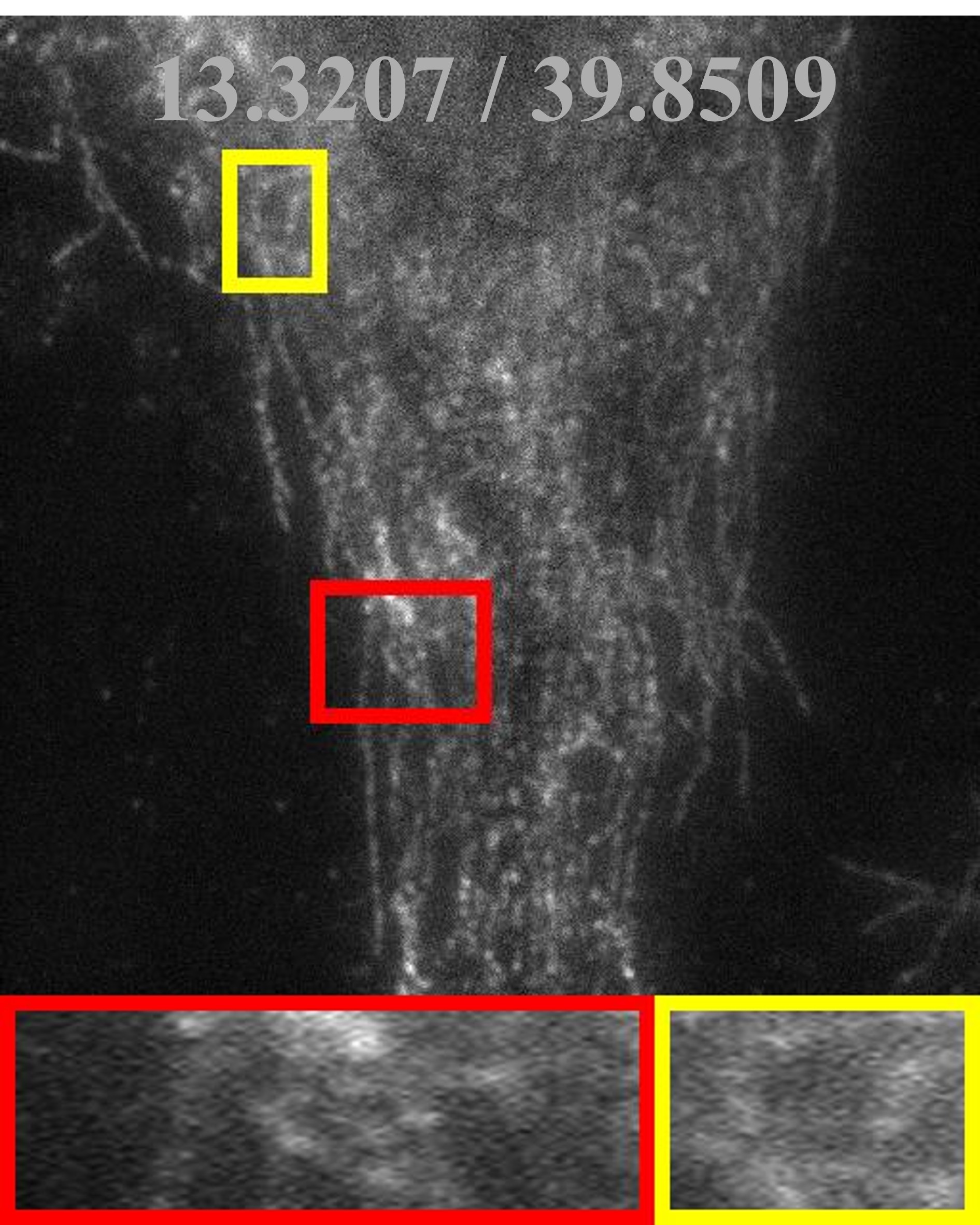}} &
			\adjustbox{valign=m}{\includegraphics[width=\imgw]{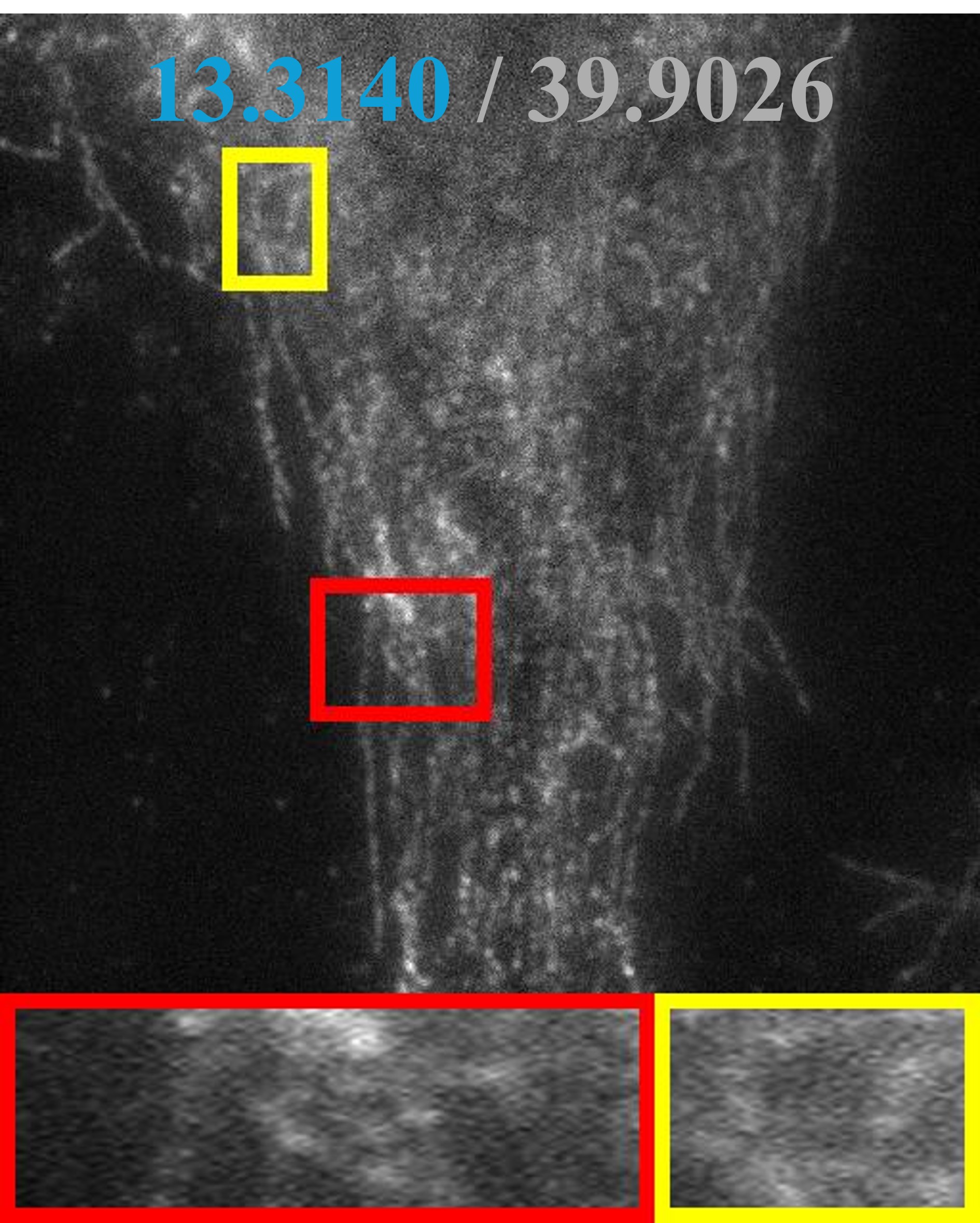}} &
			\adjustbox{valign=m}{\includegraphics[width=\imgw]{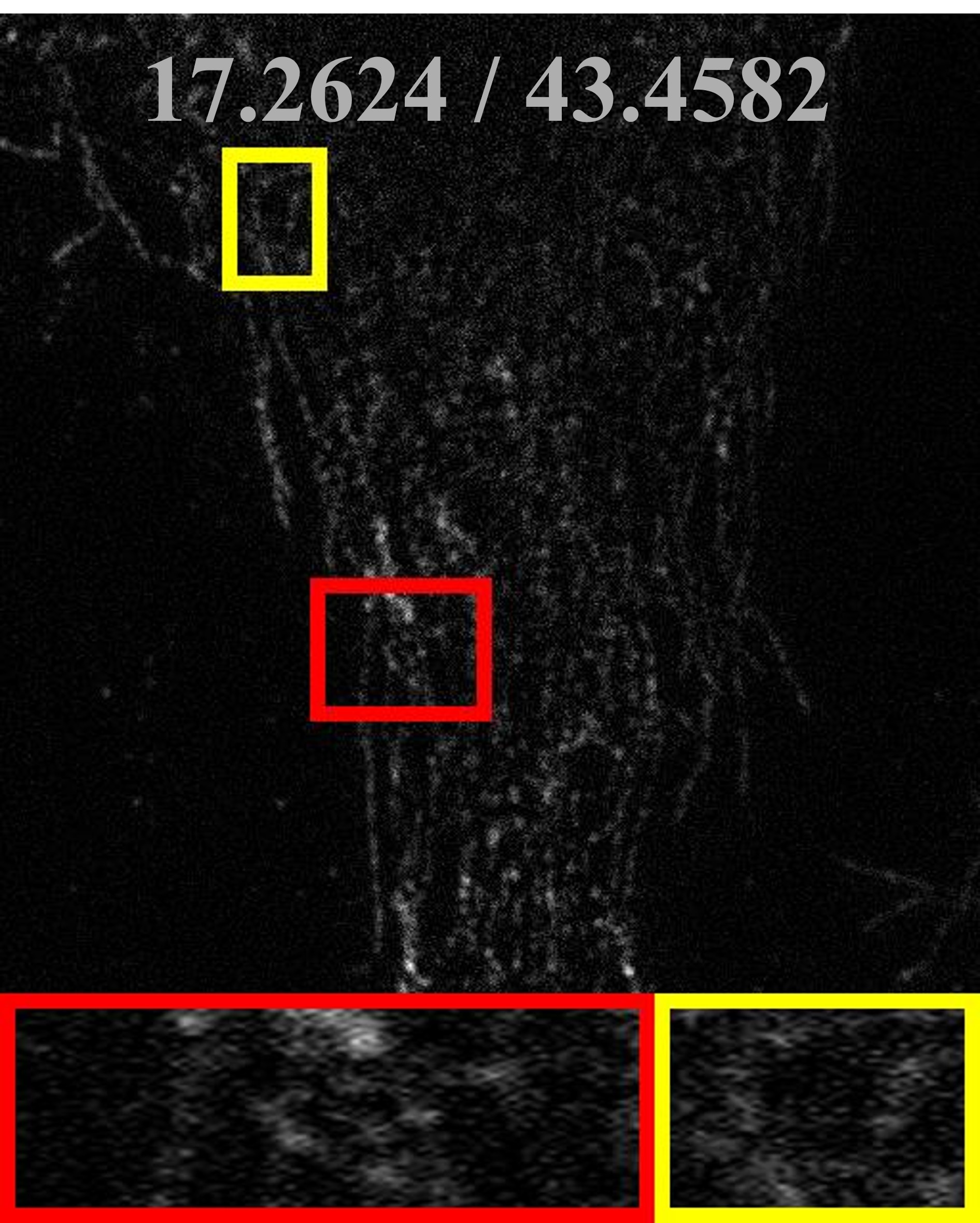}} &
			\adjustbox{valign=m}{\includegraphics[width=\imgw]{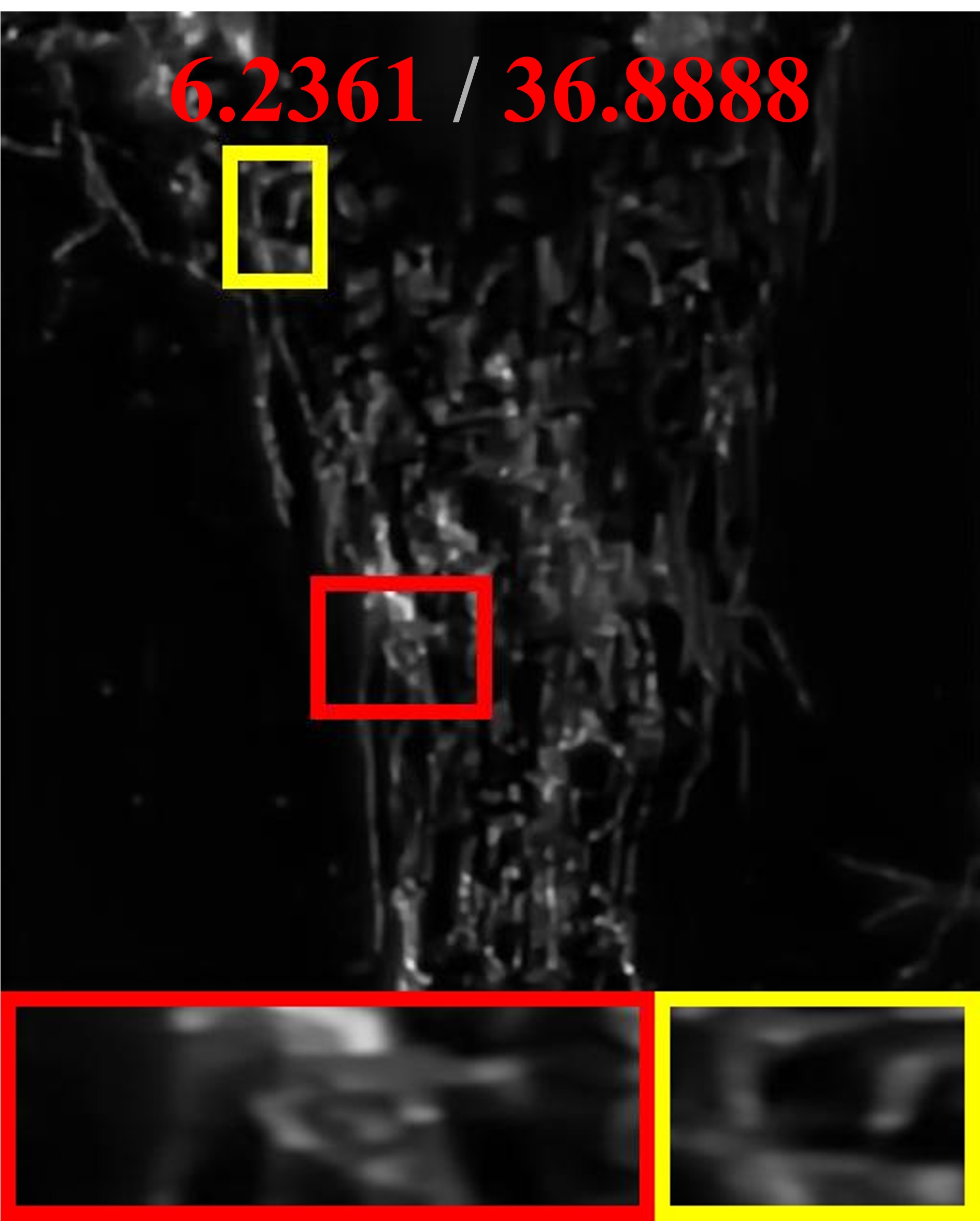}} \\
			
			\rlabel{avg400} &
			\adjustbox{valign=m}{\includegraphics[width=\imgw]{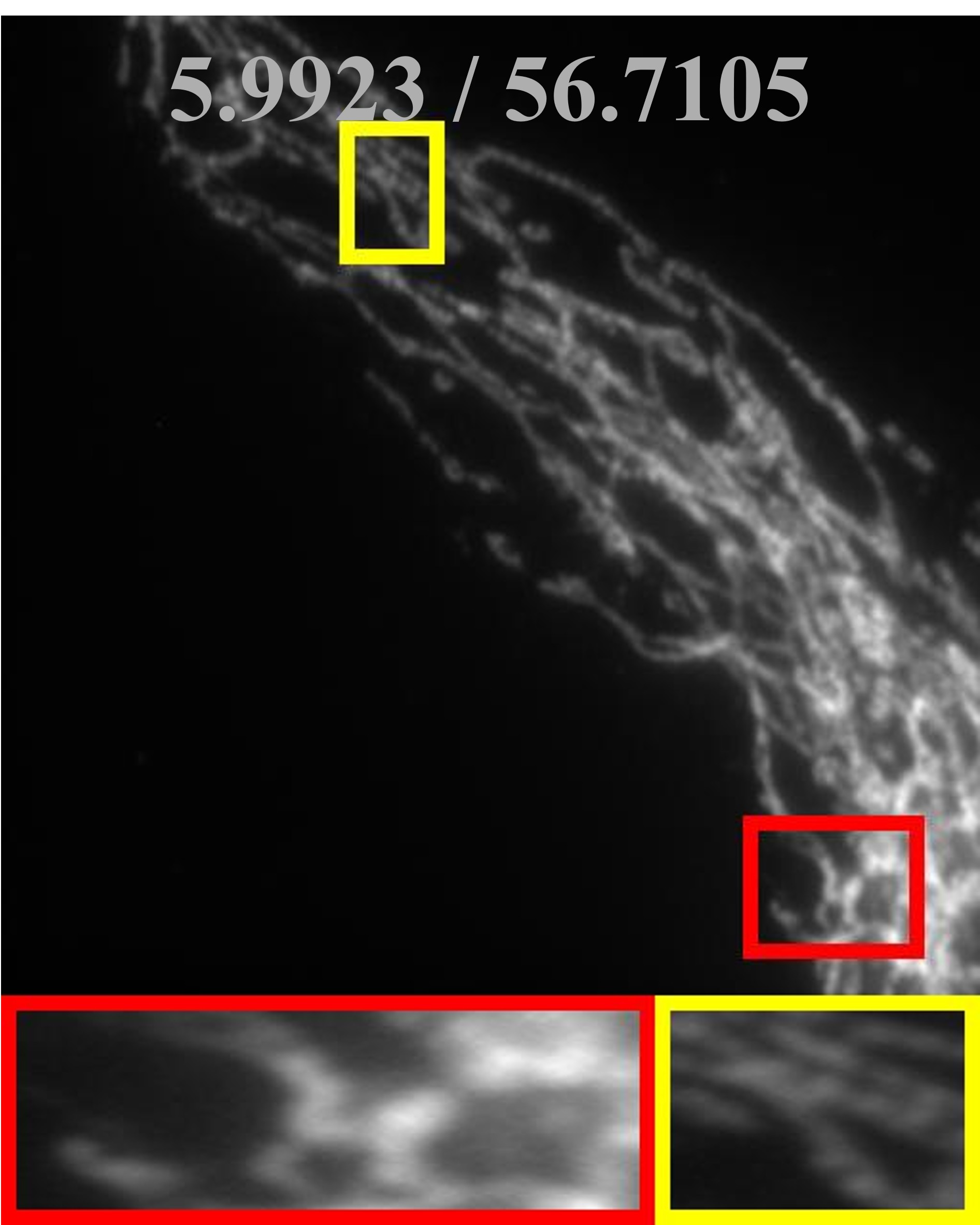}} &
			\adjustbox{valign=m}{\includegraphics[width=\imgw]{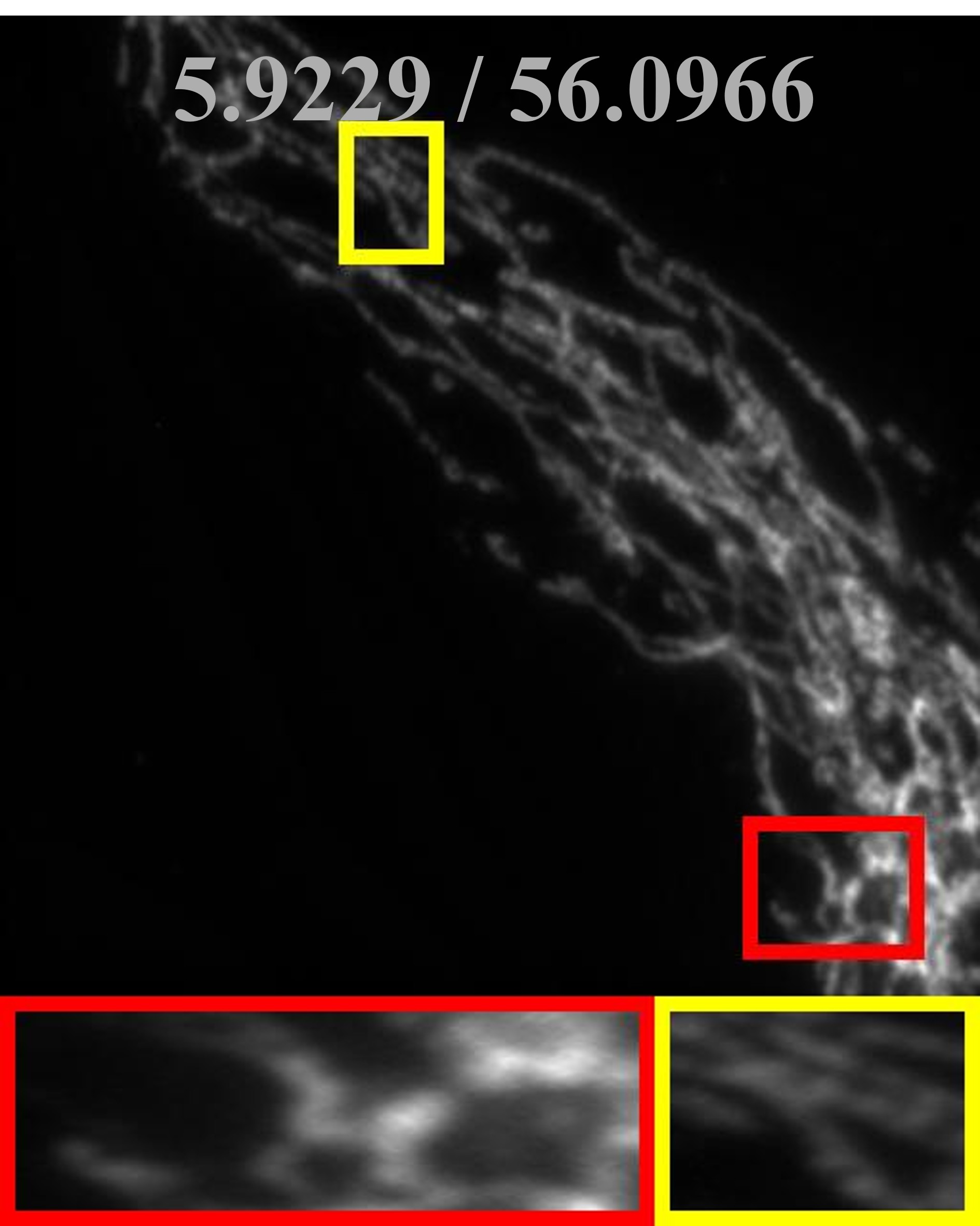}} &
			\adjustbox{valign=m}{\includegraphics[width=\imgw]{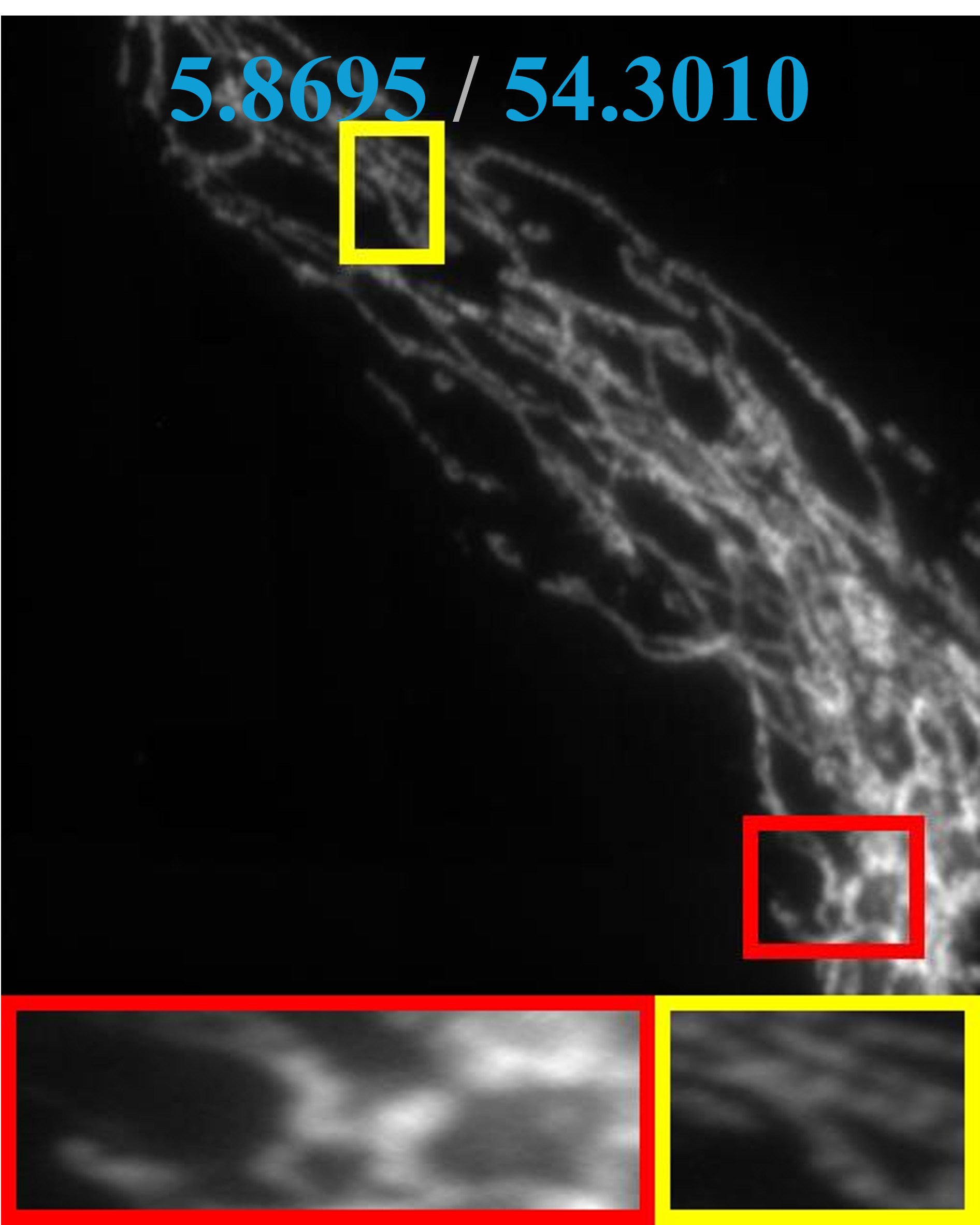}} &
			\adjustbox{valign=m}{\includegraphics[width=\imgw]{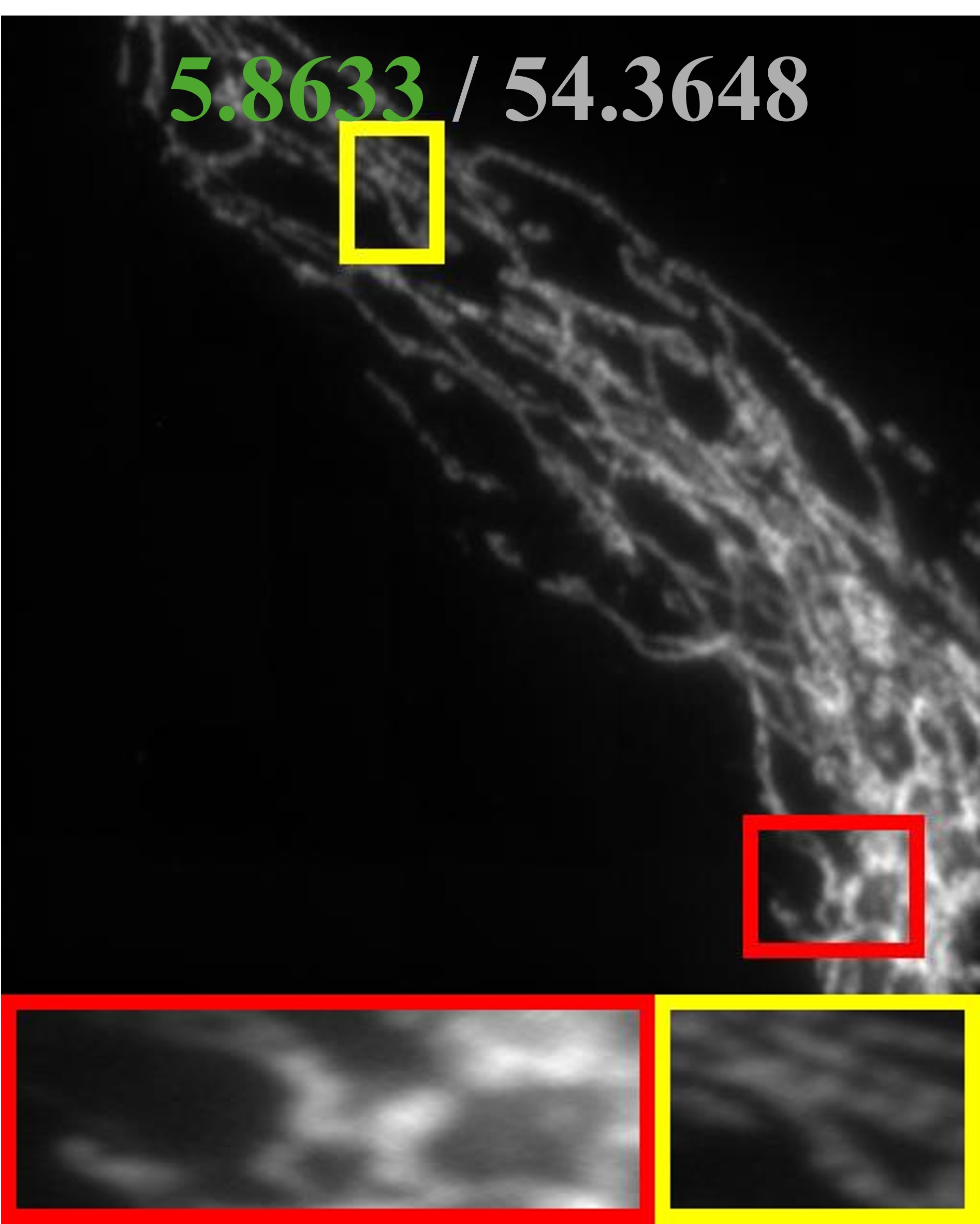}} &
			\adjustbox{valign=m}{\includegraphics[width=\imgw]{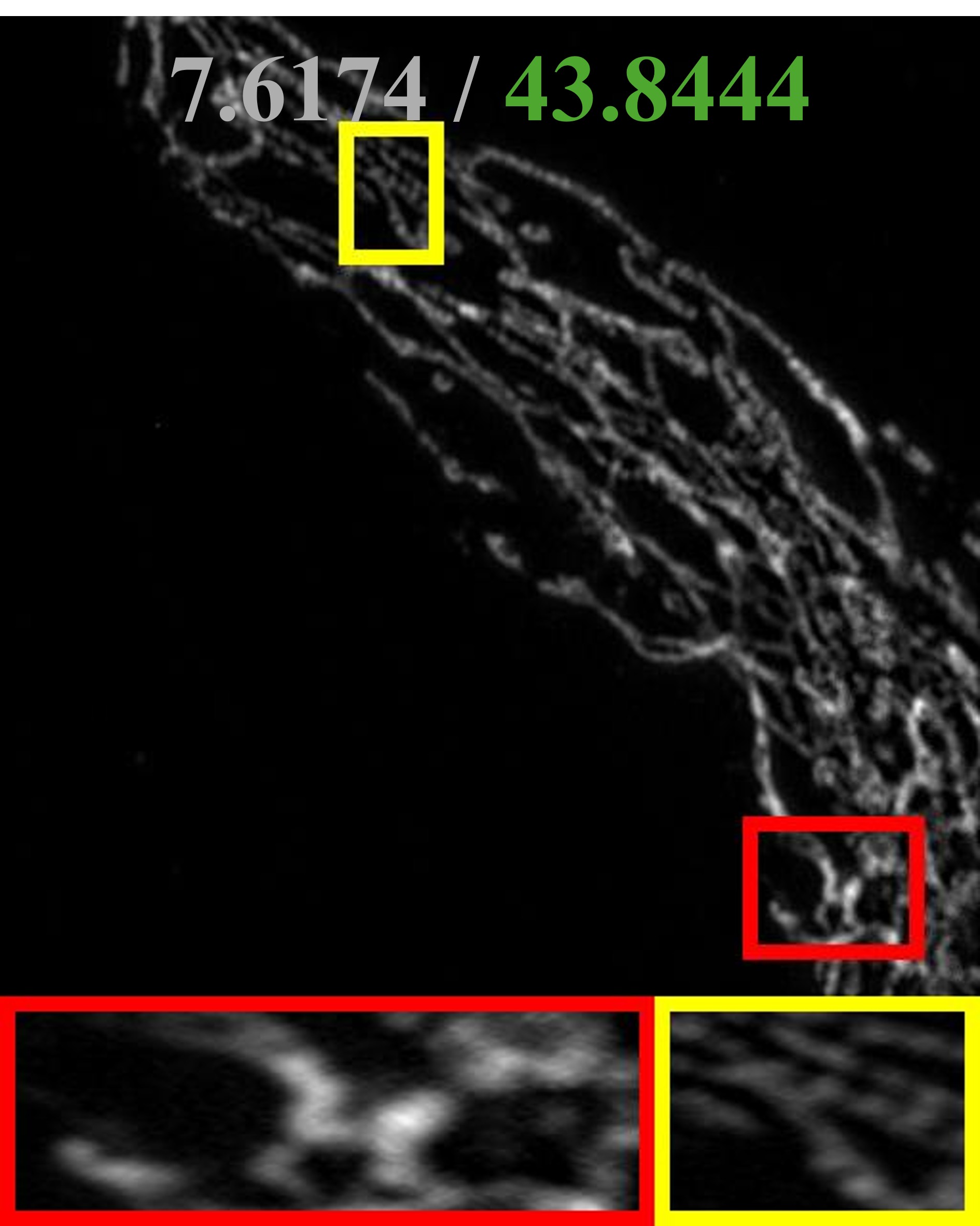}} &
			\adjustbox{valign=m}{\includegraphics[width=\imgw]{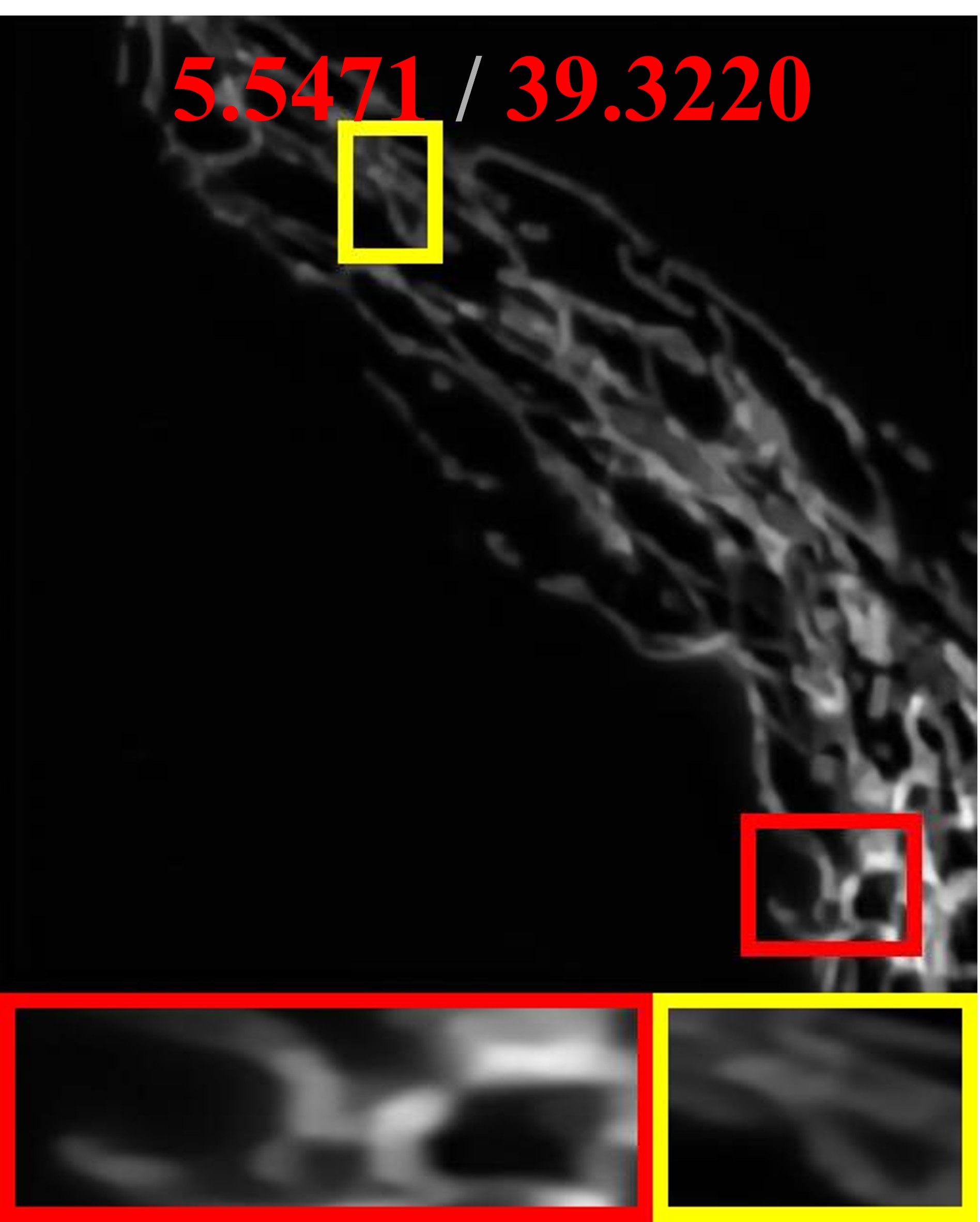}}
		\end{tabular}
	\end{adjustbox}
	
	\caption{Qualitative comparison on the W2S dataset under different averaging levels. The values shown in each image denote NIQE/BRISQUE scores, where lower values indicate better perceptual quality. PP-Net improves structural visibility and preserves finer biomedical details across different averaging levels.}
	\label{fig:biomedical_images}
\end{figure*}

Fig.~\ref{fig:biomedical_images} presents representative W2S visual results under avg1, avg4, avg16, and avg400. Compared with the degraded inputs and selected baselines, PP-Net improves structural visibility and preserves finer biomedical details, especially in the highlighted local regions.

\subsection{Ablation Study}
\label{subsec:ablation}
We further conduct ablation experiments to verify the main design choices of the proposed framework. The ablation study includes pipeline-level ablation, prior-design ablation, and training-strategy ablation.

\subsubsection{Pipeline-Level Ablation}
\label{subsubsec:pipeline_ablation}

Since the proposed framework consists of DFN-Net, ASAP, and GF-Net, we conduct module-level ablation experiments to analyze their individual effects and joint contribution.

We first evaluate the clean synthetic setting on the ITS dataset, where PP-Net$_{\mathrm{P}}$ is used without DFN-Net. As shown in Table~\ref{tab:ablation_modules_clean}, both ASAP and GF-Net contribute to restoration performance, while their combination achieves the best result. This verifies the effectiveness of coupling adaptive prior estimation with GF-Net refinement.

\begin{table}[htbp]
	\centering
	\caption{Module-level ablation of PP-Net$_{\mathrm{P}}$ on the clean ITS dataset, where higher PSNR and SSIM indicate better restoration quality.}
	\label{tab:ablation_modules_clean}
	\scriptsize
	\setlength{\tabcolsep}{14pt}
	\renewcommand{\arraystretch}{1.20}
	\begin{tabular}{c c c|cc}
		\hline
		\hline
		DFN-Net & ASAP & GF-Net & PSNR$\uparrow$ & SSIM$\uparrow$ \\
		\hline
		\xmark & \xmark & \cmark & 16.6585 & 0.8107 \\
		\xmark & \cmark & \xmark &19.5059  & 0.8415 \\
		\xmark & \cmark & \cmark & \textbf{37.6547} & \textbf{0.9901} \\
		\hline
		\hline
	\end{tabular}
\end{table}

\begin{table}[htbp]
	\centering
	\caption{Module-level ablation of PP-Net on the noisy ITS dataset, where higher PSNR and SSIM indicate better restoration quality.}
	\label{tab:ablation_modules_noisy}
	\scriptsize
	\setlength{\tabcolsep}{14pt}
	\renewcommand{\arraystretch}{1.20}
	\begin{tabular}{c c c|cc}
		\hline
		\hline
		DFN-Net & ASAP & GF-Net & PSNR$\uparrow$ & SSIM$\uparrow$ \\
		\hline
		\xmark & \xmark & \cmark & 12.2658 & 0.1346 \\
		\xmark & \cmark & \cmark & 12.3694 & 0.1324 \\
		\cmark & \xmark & \cmark & 13.9053 & 0.6573 \\
		\cmark & \cmark & \xmark & 18.7490 & 0.7258 \\
		\cmark & \cmark & \cmark & \textbf{24.9645} & \textbf{0.8093} \\
		\hline
		\hline
	\end{tabular}
\end{table}

We then evaluate the noisy synthetic setting on the noisy ITS dataset, where the complete PP-Net is used. Table~\ref{tab:ablation_modules_noisy} shows that the complete framework achieves the best performance. In particular, the comparison confirms that DFN-Net is important for stabilizing restoration under noisy degradation, while ASAP and GF-Net provide complementary benefits in prior-map recovery and detail refinement.

\subsubsection{Prior-Design Ablation}
\label{subsubsec:prior_ablation}

We next replace ASAP with alternative prior formulations while keeping the GF-Net refinement backbone unchanged.

\begin{table}[htbp]
\centering
\caption{Ablation study on prior design. The refinement backbone is fixed as GF-Net, and only the prior estimation strategy is changed. Higher PSNR and SSIM indicate better restoration performance.}
\label{tab:ablation_prior}
\scriptsize
\setlength{\tabcolsep}{22.5pt}
\renewcommand{\arraystretch}{1.1}
\begin{tabular}{c|cc}
	\hline
	\hline
	Prior Variant & SSIM$\uparrow$ & PSNR$\uparrow$ \\
	\hline
	DCP + GF-Net & 0.8234 & 17.1612 \\
	HDCP + GF-Net & 0.9742 & 36.6547 \\
	QDCP + GF-Net & 0.9832 & 33.8157 \\
	ASAP + GF-Net & \textbf{0.9901} & \textbf{37.6547} \\
	\hline
	\hline
\end{tabular}%
\end{table}

Table~\ref{tab:ablation_prior} shows that the proposed adaptive prior provides the most effective guidance among the compared prior variants. This result shows that the performance gain comes from both GF-Net refinement and improved ASAP-based scattering-map estimation, confirming the complementary roles of physical prior estimation and neural refinement.

\subsubsection{Training-Strategy Ablation}
\label{subsubsec:training_ablation}

Finally, we evaluate the proposed progressive synthetic training strategy.

\begin{table}[htbp]
\centering
\caption{Ablation study on training strategy. Higher PSNR and SSIM indicate better performance on noisy synthetic data, while lower NIQE and BRISQUE indicate better perceptual quality on real biomedical images.}
\label{tab:ablation_training}
\scriptsize
\setlength{\tabcolsep}{3.2pt}
\renewcommand{\arraystretch}{1.20}
\resizebox{\columnwidth}{!}{%
	\begin{tabular}{c|cc|cc}
		\hline
		\hline
		\multirow{2}{*}{Training Strategy}
		& \multicolumn{2}{c|}{Noisy ITS}
		& \multicolumn{2}{c}{W2S avg1} \\
		\cline{2-5}
		& PSNR$\uparrow$ & SSIM$\uparrow$
		& NIQE$\downarrow$ & BRISQUE$\downarrow$ \\
		\hline
		Clean pretraining & 12.3694 & 0.1324 & 14.4642 & 41.9512 \\
		Noisy-only training & 22.4543 & 0.7821 & 9.5643 & 41.0512 \\
		Progressive training (ours) & \textbf{24.9645} & \textbf{0.8093} & \textbf{5.5820} & \textbf{40.8959} \\
		\hline
		\hline
	\end{tabular}%
}
\end{table}

As shown in Table~\ref{tab:ablation_training}, progressive two-stage training achieves the best overall trade-off between noisy synthetic restoration performance and biomedical transferability. This result validates the effectiveness of progressively bridging synthetic haze removal, noisy degradation handling, and real biomedical image inference.

Overall, the experiments demonstrate the effectiveness of ASAP, the strong restoration performance of PP-Net$_{\mathrm{P}}$ and PP-Net, and the necessity of the major design choices. In the next section, we further investigate the practical deployment feasibility of PP-Net on on Embedded Devices.

\section{PP-Net on Embedded Devices}
\label{sec:edge_deployment}

To further evaluate the practical feasibility of PP-Net on embedded devices, we deploy the trained model on an RK3588-based platform. Rather than treating RK3588 as the only target hardware, we use it as a representative embedded device to examine whether the proposed network can be converted, quantized, and executed efficiently for edge-side biomedical image enhancement.
\subsection{Model Conversion and Quantization Workflow}
\label{subsec:conversion_quantization}

The deployment follows a staged model-conversion workflow. The trained PP-Net model is first evaluated in PyTorch with FP32 precision, and the corresponding output is used as the desktop-side reference. The model is then exported to ONNX format and further converted into RKNN format for edge-side inference. The overall deployment route is summarized as follows:

\newcommand{\boxHb}{0.65cm}
\newcommand{\gapWb}{0.45cm}      
\newcommand{\arrowLWb}{0.75pt}   

\begin{center}
	\begin{tikzpicture}[
		>=Latex,
		node distance=0cm,
		every node/.style={
			font=\rmfamily\footnotesize
		},
		proc/.style={
			draw,
			rounded corners=2pt,
			minimum height=\boxHb,
			align=center,
			inner xsep=2pt,
			inner ysep=1pt
		}
		]
		
		\node[proc, minimum width=1.85cm] (pt) {PyTorch\\FP32};
		\node[proc, minimum width=1.15cm, right=\gapWb of pt] (onnx) {ONNX};
		\node[proc, minimum width=1.15cm, right=\gapWb of onnx] (rknn) {RKNN};
		\node[proc, minimum width=1.85cm, right=\gapWb of rknn] (int8) {RK3588\\INT8};
		
		\draw[->, line width=\arrowLWb] (pt) -- (onnx);
		\draw[->, line width=\arrowLWb] (onnx) -- (rknn);
		\draw[->, line width=\arrowLWb] (rknn) -- (int8);
		
	\end{tikzpicture}
\end{center}

After model conversion, INT8 post-training quantization is adopted for edge acceleration. Since biomedical image restoration is sensitive to intensity shifts, contrast distortion, and local structural artifacts, representative W2S images from different averaging levels are used for quantization calibration. This domain-specific calibration helps cover diverse noise levels and intensity distributions, thereby reducing quantization-induced degradation when the model is applied to real biomedical inputs.

Before final deployment, the converted RKNN model is checked against the original PyTorch implementation to ensure visual consistency. This step is important for biomedical image enhancement because small structural distortions introduced during model conversion or quantization may affect the visibility of local tissue details.

\subsection{Deployment Configuration and Edge Inference Results}
\label{subsec:deployment_results}

The deployment configuration is summarized in Table~\ref{tab:rk3588_deployment}. The PP-Net model is executed on the RK3588 platform with INT8 quantization enabled. At an input resolution of $512\times512$, the average single-image inference latency is approximately 200 ms over 360 test images, supporting efficient edge-side biomedical enhancement on the RK3588 platform.

\begin{table}[htbp]
	\centering
	\caption{Deployment configuration and performance of PP-Net on the RK3588 platform.}
	\label{tab:rk3588_deployment}
	\scriptsize
	\setlength{\tabcolsep}{3pt}
	\renewcommand{\arraystretch}{1.08}
	\begin{tabular}{p{0.43\columnwidth}|p{0.47\columnwidth}}
		\hline
		\hline
		\textbf{Item} & \textbf{Configuration} \\
		\hline
		Deployment model & PP-Net \\
		Training framework & PyTorch / Python \\
		Deployment framework & RKNN \\
		Hardware platform & TOP-EET iTOP-RK3588 \\
		Processor & RK3588 \\
		Input resolution & $512 \times 512$ \\
		Inference mode & Single-image inference \\
		Average inference latency per image & $\sim$200 ms over 360 images \\
		Quantization & INT8-enabled \\
		\hline
		\hline
	\end{tabular}
\end{table}

\begin{figure}[htbp]
	\centering
	\setlength{\tabcolsep}{1.5pt}
	\renewcommand{\arraystretch}{1.0}
	
	\newcommand{\figNineImgW}{0.45\columnwidth}
	\newcommand{\figNineLabel}[1]{\adjustbox{valign=m}{\rotatebox[origin=c]{90}{\textbf{#1}}}}
	\newcommand{\figNineHead}[1]{\makebox[\figNineImgW][c]{\textbf{#1}}}
	
	\begin{adjustbox}{max totalsize={\columnwidth}{0.95\textheight},center}
		\begin{tabular}{c ccc}
			& \figNineHead{Input} &
			\figNineHead{PyTorch PP-Net} &
			\figNineHead{RK3588 PP-Net} \\
			
			\figNineLabel{avg1} &
			\adjustbox{valign=m}{\includegraphics[width=\figNineImgW]{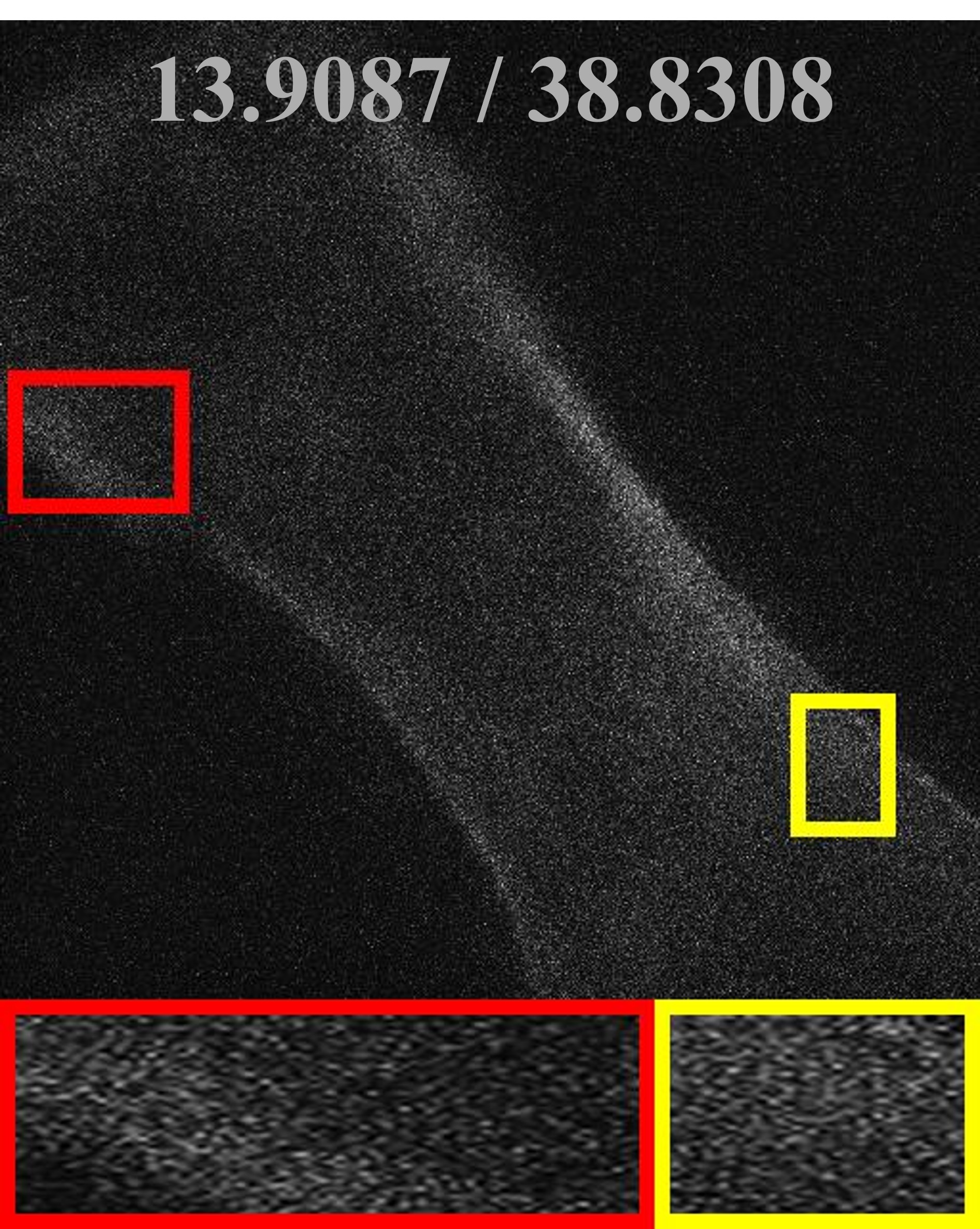}} &
			\adjustbox{valign=m}{\includegraphics[width=\figNineImgW]{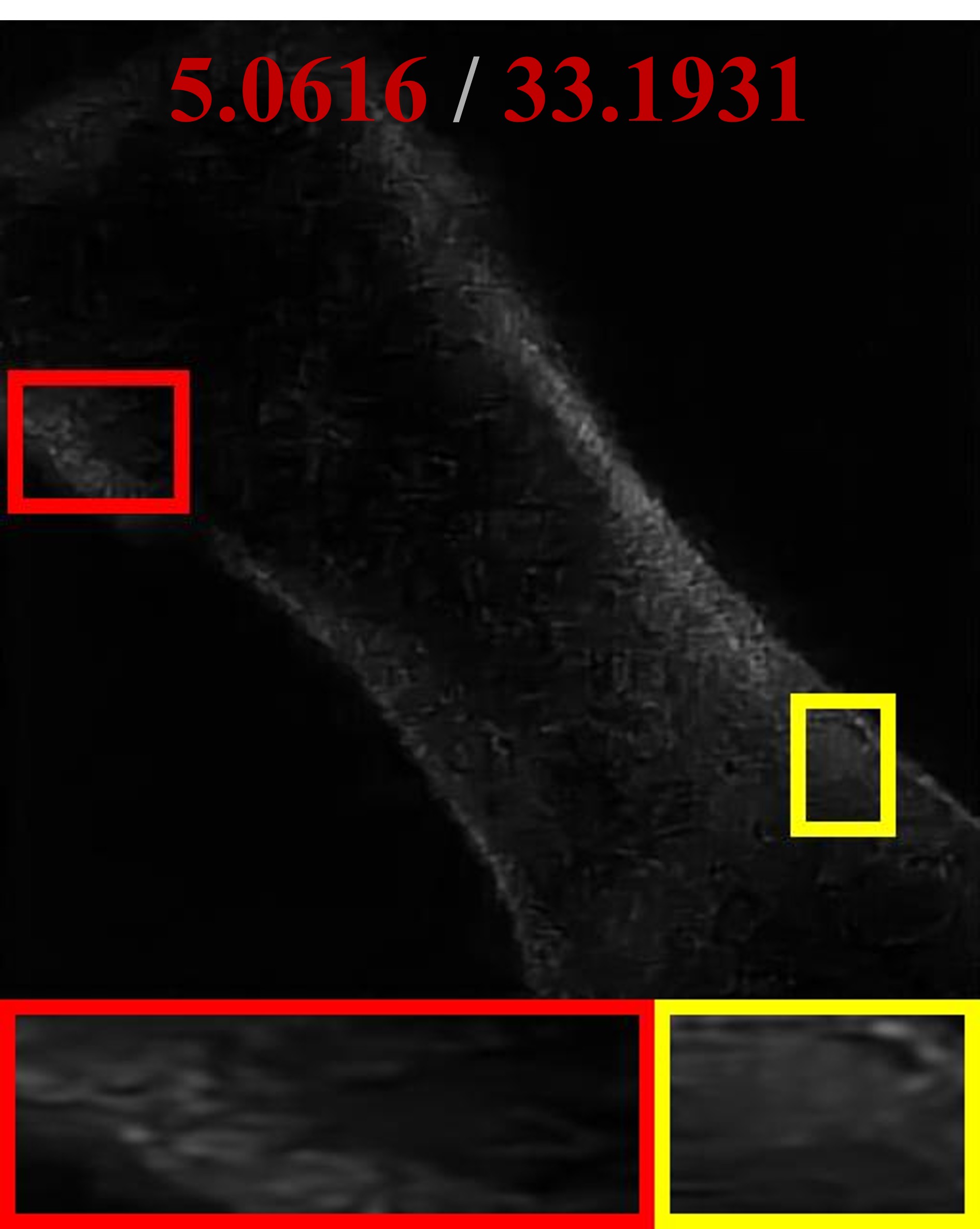}} &
			\adjustbox{valign=m}{\includegraphics[width=\figNineImgW]{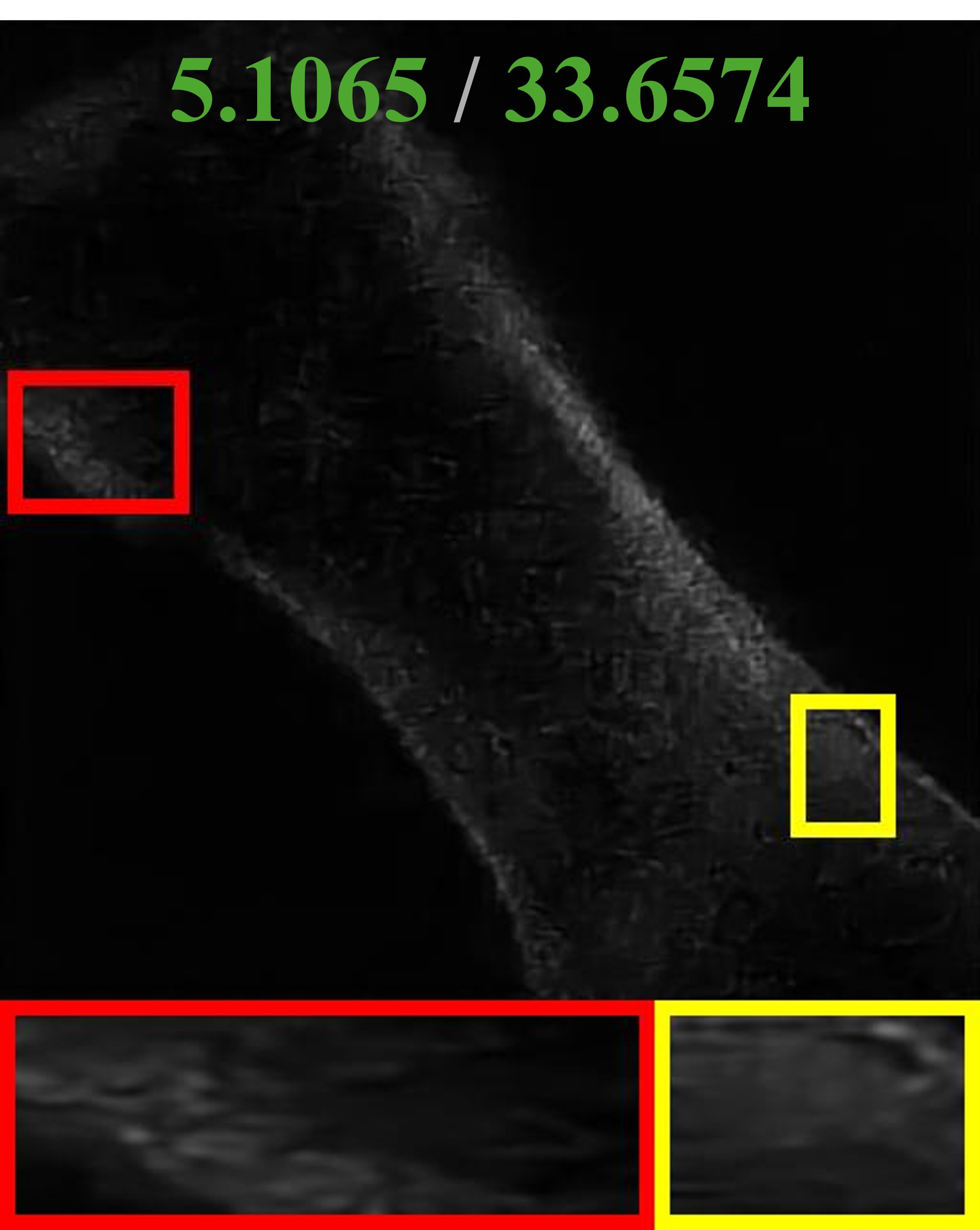}} \\
			
			\figNineLabel{avg16} &
			\adjustbox{valign=m}{\includegraphics[width=\figNineImgW]{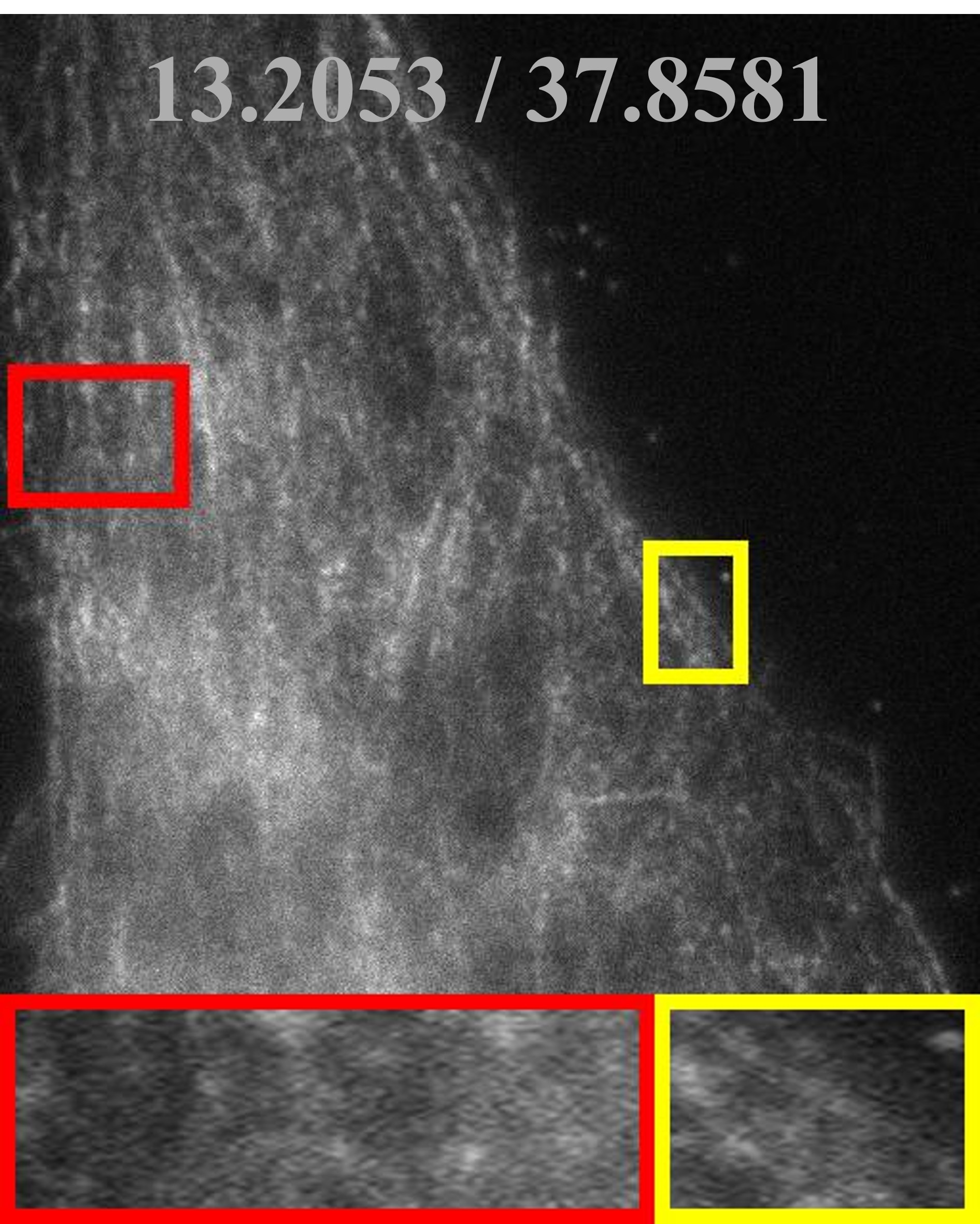}} &
			\adjustbox{valign=m}{\includegraphics[width=\figNineImgW]{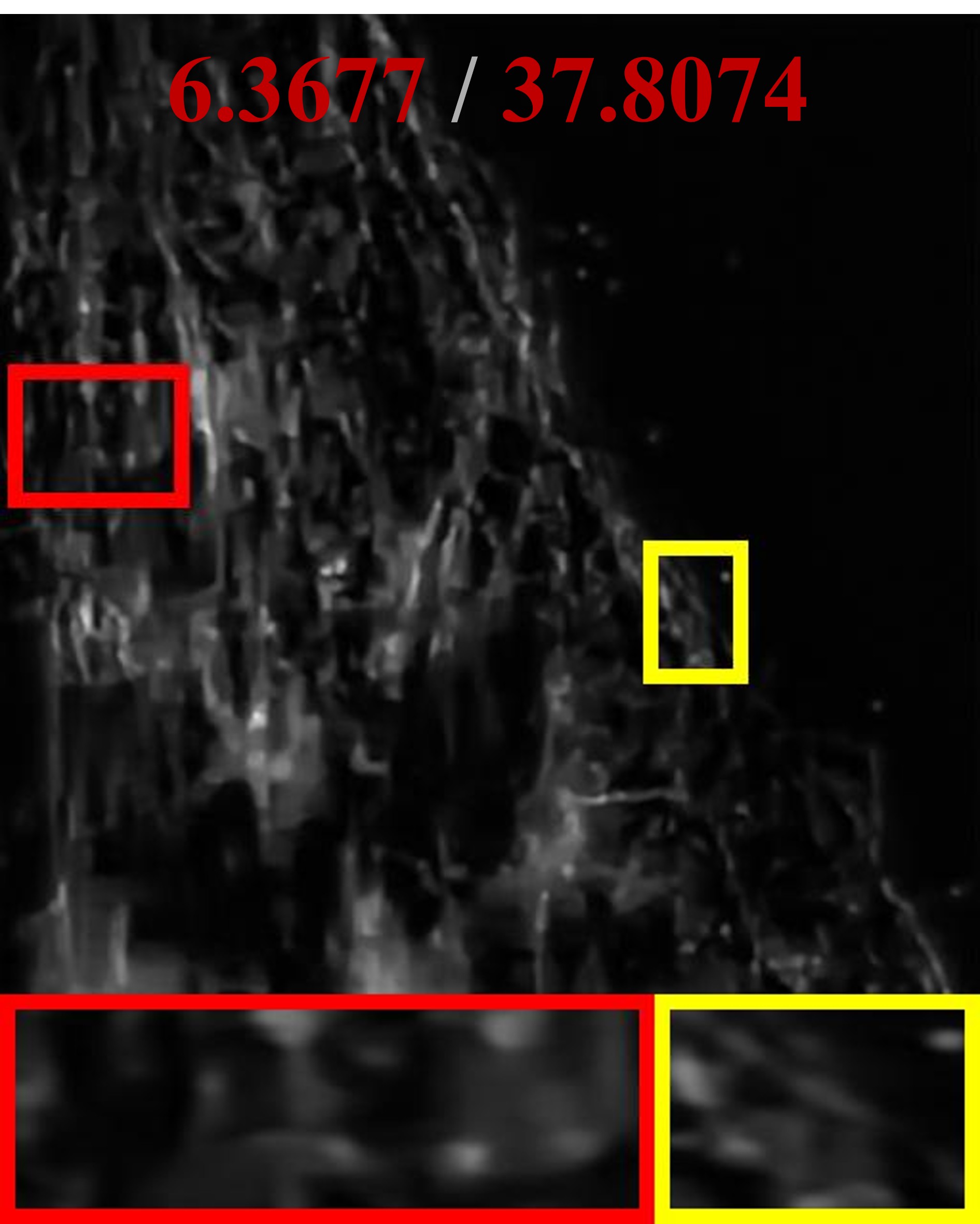}} &
			\adjustbox{valign=m}{\includegraphics[width=\figNineImgW]{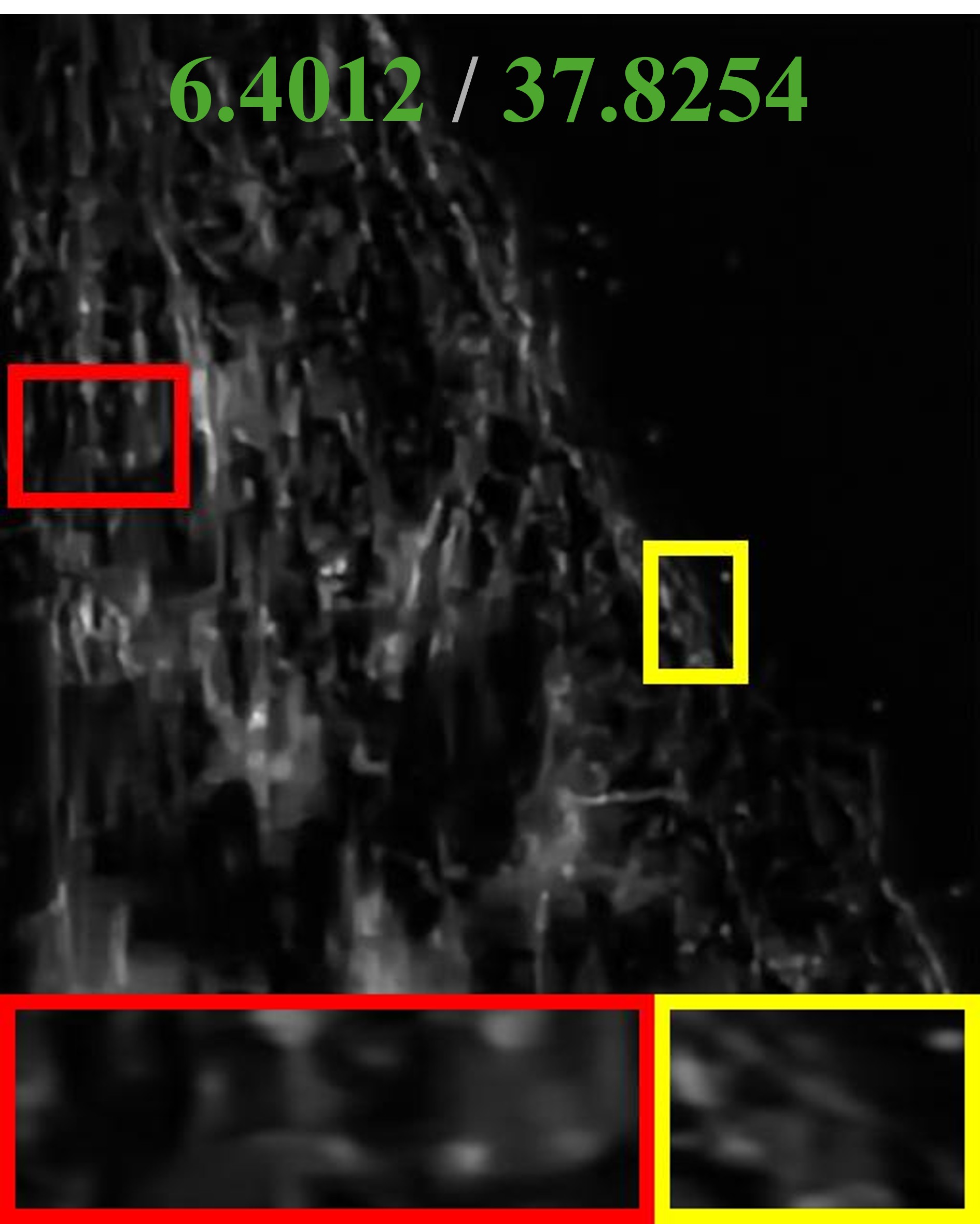}} \\
			
			\figNineLabel{avg400} &
			\adjustbox{valign=m}{\includegraphics[width=\figNineImgW]{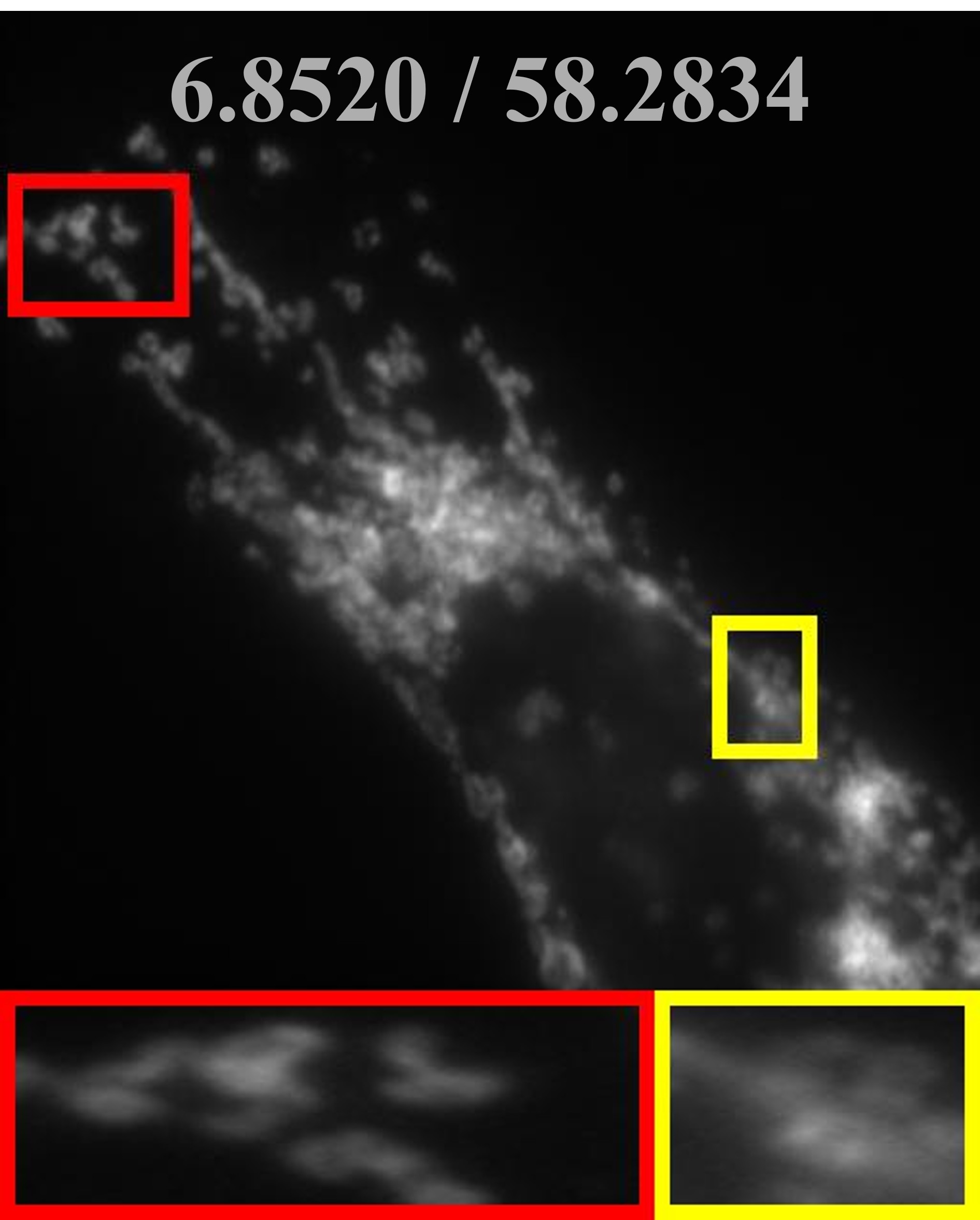}} &
			\adjustbox{valign=m}{\includegraphics[width=\figNineImgW]{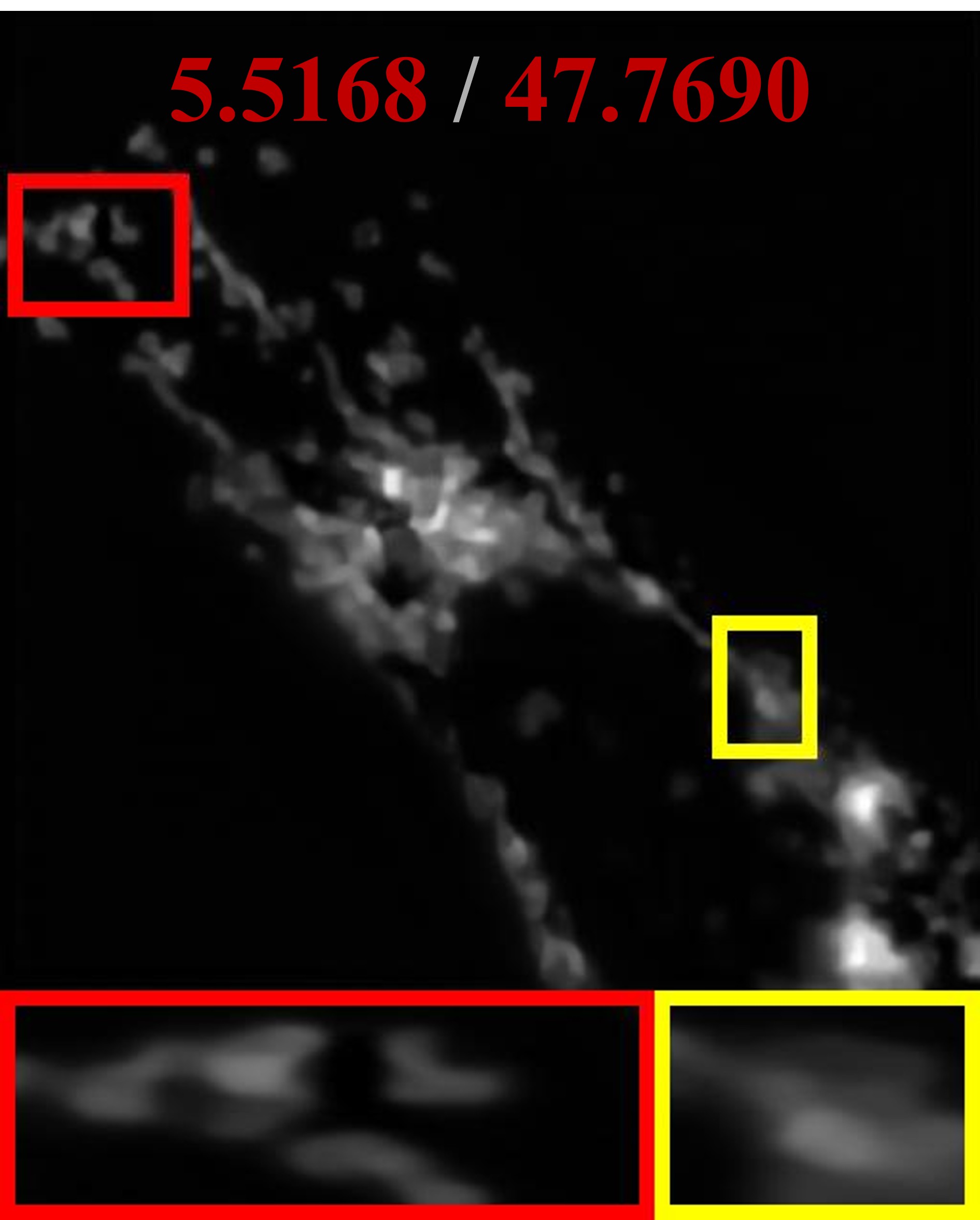}} &
			\adjustbox{valign=m}{\includegraphics[width=\figNineImgW]{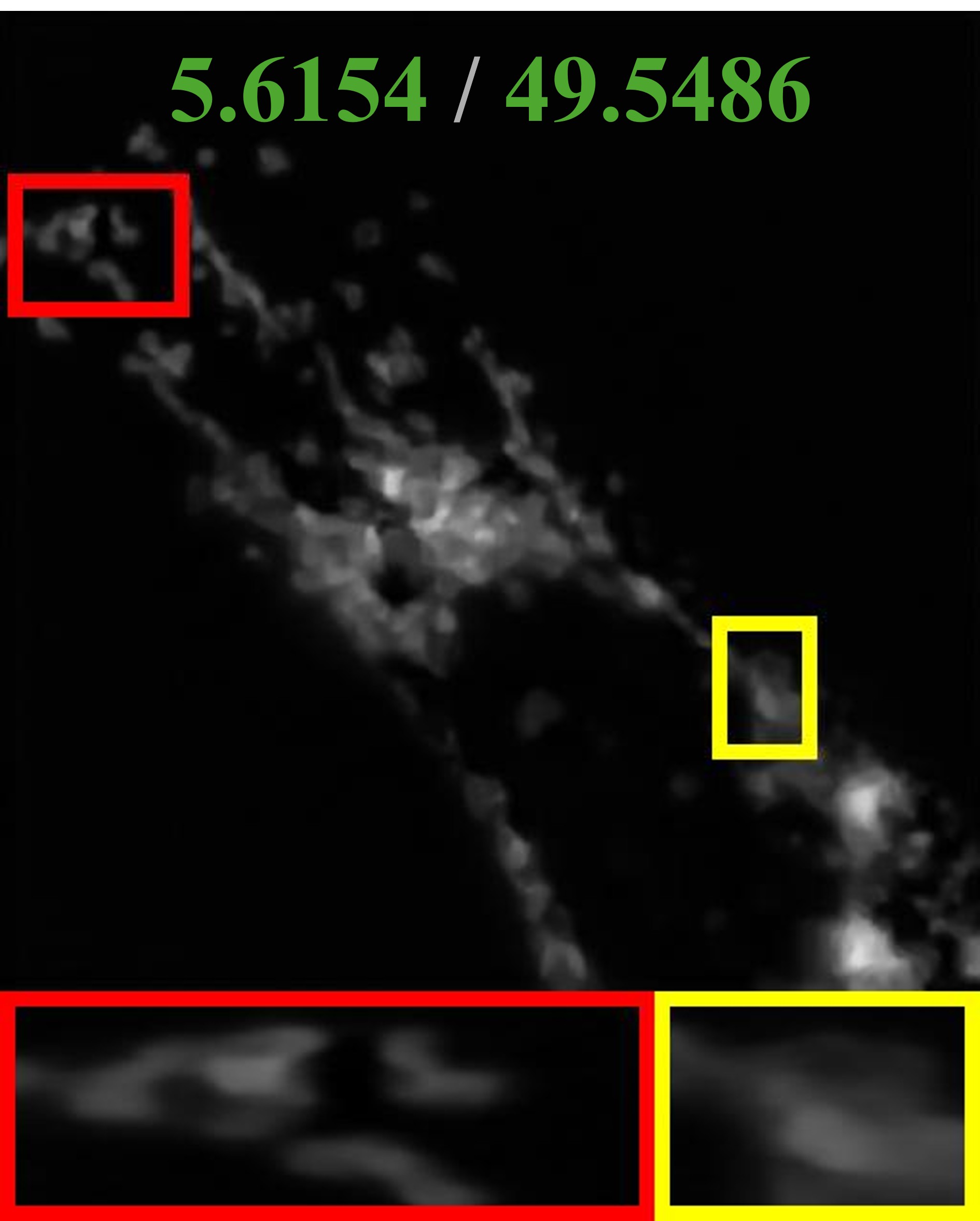}}
		\end{tabular}
	\end{adjustbox}
	\caption{Representative RK3588 deployment results under different W2S averaging levels. The RK3588 outputs remain visually close to the PyTorch outputs, indicating that RKNN conversion and INT8 quantization preserve the main structural details during edge-side inference.}
	\label{fig:rk3588_results}
\end{figure}

To further assess deployment reliability, we compare the desktop-side PyTorch output with the RK3588-side output. As shown in Fig.~\ref{fig:rk3588_results}, the RK3588 results remain visually close to the PyTorch results under different W2S averaging levels, preserving the main structural details and local tissue textures. This comparison suggests that RKNN conversion and INT8 quantization do not introduce obvious deployment-induced visual artifacts.

\subsection{Structure Visibility Analysis}
\label{subsec:structure_visibility}

In addition to the visual deployment comparison shown in Fig.~\ref{fig:rk3588_results}, we further quantify the structural visibility of the RK3588 outputs using a Gaussian-smoothed Tenengrad score. Since W2S biomedical images often contain sensor-induced noise, each image is first lightly smoothed using a $5\times5$ Gaussian filter with $\sigma=0.95$, and the Sobel gradient magnitude is then computed. The resulting Tenengrad score is used as a no-reference indicator of local structural sharpness.

Let $I$ denote the input image and $G_{\sigma}$ denote the $5\times5$ Gaussian filter with $\sigma=0.95$. The smoothed image is computed as
\begin{equation}
	\label{eq:gaussian_smoothing}
	I_{\sigma}=G_{\sigma} * I,
\end{equation}
where $*$ denotes convolution. The horizontal and vertical Sobel gradients are then obtained by
\begin{equation}
	\label{eq:sobel_gradient}
	G_x=S_x * I_{\sigma}, \qquad
	G_y=S_y * I_{\sigma},
\end{equation}
where $S_x$ and $S_y$ are the Sobel operators along the horizontal and vertical directions, respectively. The Gaussian-smoothed Tenengrad score is defined as
\begin{equation}
	\label{eq:tenengrad_score}
	T(I)=\frac{1}{|\Omega|}
	\sum_{(x,y)\in\Omega}
	\left(G_x^2(x,y)+G_y^2(x,y)\right),
\end{equation}
where $\Omega$ denotes the image domain. A higher $T(I)$ indicates stronger local structural sharpness after Gaussian smoothing.

\begin{table}[htbp]
	\centering
	\caption{Structure visibility analysis on W2S images using Gaussian-smoothed Tenengrad scores. Higher values indicate stronger structural sharpness, and all scores are multiplied by $10^4$ for readability.}
	\label{tab:structure_visibility}
	\scriptsize
	\setlength{\tabcolsep}{4pt}
	\renewcommand{\arraystretch}{1.10}
	\resizebox{\columnwidth}{!}{%
		\begin{tabular}{c|ccc}
			\hline
			\hline
			W2S Level & Input & RK3588 PP-Net & Improvement \\
			\hline
			avg1   & 184.30 & 191.19 & 3.74\%  \\
			avg16  & 186.73 & 208.66 & 11.74\% \\
			avg400 & 80.48  & 97.25  & 20.84\% \\
			\hline
			\hline
		\end{tabular}%
	}
\end{table}

As shown in Table~\ref{tab:structure_visibility}, RK3588 PP-Net obtains higher Gaussian-smoothed Tenengrad scores than the degraded inputs at different W2S averaging levels. This result quantitatively supports the visual observation in Fig.~\ref{fig:rk3588_results}, indicating that edge-side PP-Net inference improves structural visibility while maintaining practical deployment efficiency.

\section{Conclusion}
\label{sec:conclusion}

This paper presented PP-Net, an edge-oriented hybrid physical-prior neural network for biomedical scattered-light removal. PP-Net integrates DFN-Net, ASAP, and GF-Net into a progressive \emph{network-prior-network} pipeline, combining noise suppression, scattering-map estimation, prior-map recovery, and lightweight GF-Net refinement. To reduce the dependence on paired biomedical ground truth, we further developed a progressive synthetic training and cross-domain transfer strategy for real biomedical image inference.

Experiments on multiple benchmark datasets demonstrated the effectiveness of ASAP, the strong restoration performance of PP-Net$_{\mathrm{P}}$, and the robustness of PP-Net under joint noise-and-scattering degradation. Direct transfer to real biomedical images also showed promising visual enhancement and competitive no-reference image quality results. Furthermore, RK3588 deployment with RKNN conversion and INT8 quantization achieved an average inference latency of approximately 200 ms per image over 360 test images, confirming efficient edge-side biomedical enhancement.

Future work will focus on unsupervised biomedical domain adaptation, task-oriented clinical validation, and hardware-aware acceleration for real-time biomedical video enhancement on embedded devices.
\section*{Acknowledgments}
This work was supported in part by National Natural Science Foundation of China under Grant 12471502, Science and Technology Development Plan Project of Jilin Province, China under Grant 20260204053YY, and CAS Hundred Talents Program.

%
%
%
%
%
%
%
%

\bibliographystyle{IEEEtran}
\bibliography{references}

\vfill

\end{document}